\documentclass[runningheads]{llncs}
\usepackage[T1]{fontenc}
\usepackage{graphicx}
\usepackage{booktabs}
\usepackage[misc]{ifsym}
\newcommand{\corr}{(\Letter)}
\usepackage{mwe}
\usepackage{cite}
\usepackage{amsmath,amssymb,amsfonts}
\usepackage{algorithmic}
\usepackage{textcomp}
\usepackage{svg}
\usepackage{comment}
\usepackage{subcaption}
\usepackage{placeins}

\PassOptionsToPackage{table}{xcolor}
\usepackage{pgffor}
\usepackage[labelfont=bf]{caption}
\usepackage[table]{xcolor}
\usepackage{array}
\usepackage{colortbl}
\usepackage{geometry}
\usepackage{float} 
\usepackage{hyperref} 

\usepackage{multibib}

\newcites{app}{Appendix References} 

\newcommand{\rootdir}{figures/qualitative_results}

\newlength{\thumbwidth}
\newcommand{\CellImg}[3]{%
  \includegraphics[width=\thumbwidth]{\rootdir/#1/#2/vis_data/vis_image/#3}%
}
\newcommand{\ImgRow}[1]{%
  \CellImg{deeplab}{inference_pics_synth}{#1} &
  \CellImg{deeplab}{inference_pics_16_025_lora_gray}{#1} &
  \CellImg{segformer}{inference_pics_synth}{#1} &
  \CellImg{segformer}{inference_pics_16_025_lora_gray}{#1} &
  \CellImg{mask2former}{inference_pics_synth}{#1} &
  \CellImg{mask2former}{inference_pics_full_stylized}{#1} \tabularnewline
}
\newcommand{\CompareAcrossModels}{%
  \begin{figure}[h!]
    \centering
    \setlength{\tabcolsep}{1pt}%
    \resizebox{\textwidth}{!}{%
    \begin{tabular}{@{} cc @{\hspace{4pt}} cc @{\hspace{4pt}} cc @{}}
      \multicolumn{2}{c}{\bfseries Deeplabv3+} &
      \multicolumn{2}{c}{\bfseries SegFormer} &
      \multicolumn{2}{c}{\bfseries Mask2Former} \\
      \scriptsize Baseline & \scriptsize Best &
      \scriptsize Baseline & \scriptsize Best &
      \scriptsize Baseline & \scriptsize Best \\
      \midrule
      \ImgRow{img_val_00024.png_25.png}%
      \ImgRow{img_val_00049.png_50.png}
      \ImgRow{img_val_00099.png_100.png}
      \ImgRow{img_val_00124.png_125.png}
      \ImgRow{img_val_00149.png_150.png}
       \ImgRow{img_val_00199.png_200.png}
      \ImgRow{img_val_00224.png_225.png}
      
    \end{tabular}
    }
    \caption{More qualitative results per models with baseline vs best checkpoint. }
    \label{fig:models-per-column_qual}
  \end{figure}%
}

\definecolor{cls_background}{RGB}{0,0,0}
\definecolor{cls_seat_backrest}{RGB}{76,0,92}
\definecolor{cls_glasses}{RGB}{255,225,0}
\definecolor{cls_face_mask}{RGB}{66,102,0}
\definecolor{cls_seat_headrest}{RGB}{240,163,255}
\definecolor{cls_head_ware}{RGB}{255,168,187}
\definecolor{cls_face}{RGB}{255,204,153}
\definecolor{cls_person}{RGB}{0,0,128}
\definecolor{cls_seat_belt}{RGB}{157,204,0}
\definecolor{cls_seat_pad}{RGB}{194,0,136}
\definecolor{cls_child_seat}{RGB}{148,255,181}
\definecolor{cls_hair}{RGB}{43,206,72}
\definecolor{cls_clothing}{RGB}{153,0,0}
\definecolor{cls_other_object}{RGB}{255,0,16}

\begin{document}
\title{Texture-Shape Bias Balancing for Robust Synthetic-to-Real Semantic Segmentation in Automotive NIR Imagery}
\titlerunning{Bias Balancing for Robust Synth2Real Semantic Segmentation in Automotive NIR}
\author{Felix Stillger\inst{1,2}\thanks{These authors contributed equally.} \corr \and Ben Hamscher\inst{3}\textsuperscript{$\star$} \and Lukas Hahn\inst{2} \and Annika Mütze\inst{4} \and \\ Tobias Meisen\inst{1} \and Kira Maag\inst{3}} 
\authorrunning{F. Stillger et al.} 
\institute{University of Wuppertal, Germany  \email{\{felix.stillger,meisen\}@uni-wuppertal.de} \and Aptiv, Germany \email{lukas.hahn@aptiv.com} \and Heinrich Heine University Düsseldorf, Germany, \email{\{ben.hamscher,kira.maag\}@hhu.de} \and Osnabrück University, Germany, \email{annika.muetze@uni-osnabrueck.de}}
\toctitle{Bias Balancing for Robust Synth2Real Semantic Segmentation in Automotive NIR}
\tocauthor{Felix~Stillger, Ben~Hamscher, Lukas~Hahn, Annika~Mütze, Tobias~Meisen, Kira~Maag}
\maketitle              

\begin{abstract}
Semantic segmentation is a fundamental component of visual perception in modern automotive systems, enabling pixel-level scene understanding. %
Near-Infrared imaging (NIR) offers stable detection under difficult illumination conditions, but the development of domain-specific semantic segmentation models remains challenging due to the lack of high-quality annotated data from real-world scenarios. 
Synthetic datasets offer a scalable alternative, but models trained on synthetic images often suffer performance degradation when transferred to real domains. 
We present the first systematic study on synthetic to real domain adaptation for semantic segmentation in NIR images in the automotive domain. 
We propose a generative augmentation framework that transforms synthetic images into realistic NIR-style variants via our introduced target style adaptation (TSA). 
TSA fine-tunes a latent diffusion model via low-rank adaptation on a small curated set of real NIR images and applies it to synthetic training data using structure-preserving multi-signal conditioning.
To reduce texture bias and improve segmentation robustness, we further apply a Voronoi-based style diversification strategy (VSD) that modifies the original textures while preserving scene geometry. %
Experiments with multiple model architectures on NIR data from vehicle interiors and street scenes show that balancing inductive bias during training leads to noticeably more robust semantic segmentation and effectively reduces the domain gap in our real-world scenarios by up to 63.6\,\% on exterior and 28.4\,\% on interior data.
The code is available at \href{https://github.com/felixstillger/synth2real-nir-segmentation}{GitHub}\footnote{\url{https://github.com/felixstillger/synth2real-nir-segmentation}}.

\keywords{Semantic Segmentation \and Synthetic-to-Real \and Domain Adaptation \and Near-Infrared \and Bias Balancing \and Texture and Shape Bias \and Robustness}
\end{abstract}

\section{Introduction}
\begin{figure*}[t]
    \centering
    \newlength{\myimgwidth}
    \setlength{\myimgwidth}{0.185\linewidth} 
    \setlength{\tabcolsep}{0.3pt}
    \newcommand{\imgcell}[1]{%
        \parbox[c]{\myimgwidth}{%
            \includegraphics[width=\linewidth]{#1}%
        }%
    }
    \newcommand{\imgcellext}[1]{%
        \parbox[c]{\myimgwidth}{%
            \includegraphics[width=\linewidth, trim=0.0cm 2cm 0.0cm 0.5cm, clip]{#1}%
        }%
    }

    \begin{tabular}{ c @{\hspace{3pt}} c c c@{\hspace{5pt}} c c }
        \rotatebox[origin=c]{90}{\scriptsize \textbf{Interior}} &
        \imgcell{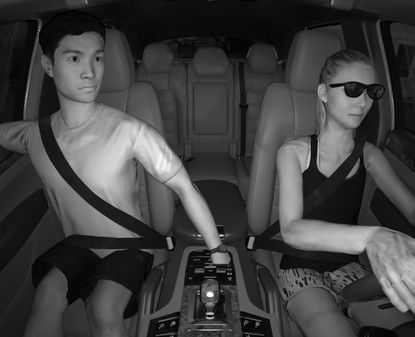} &
        \imgcell{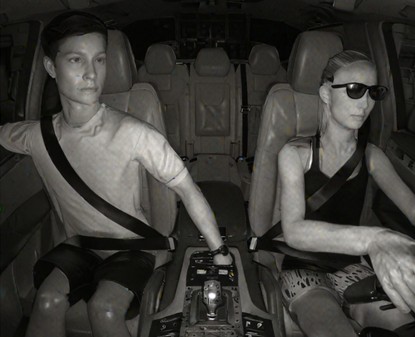} &
        \imgcell{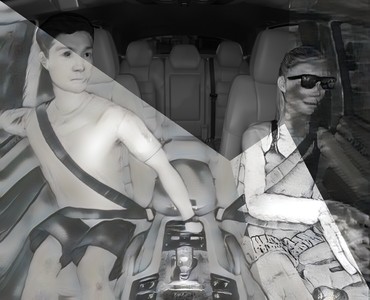} &
        \imgcell{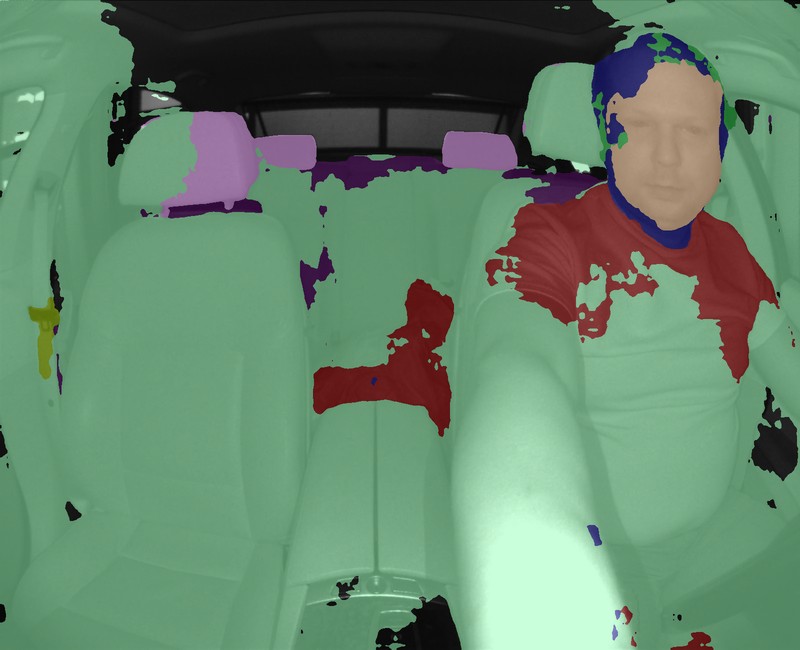} &
        \imgcell{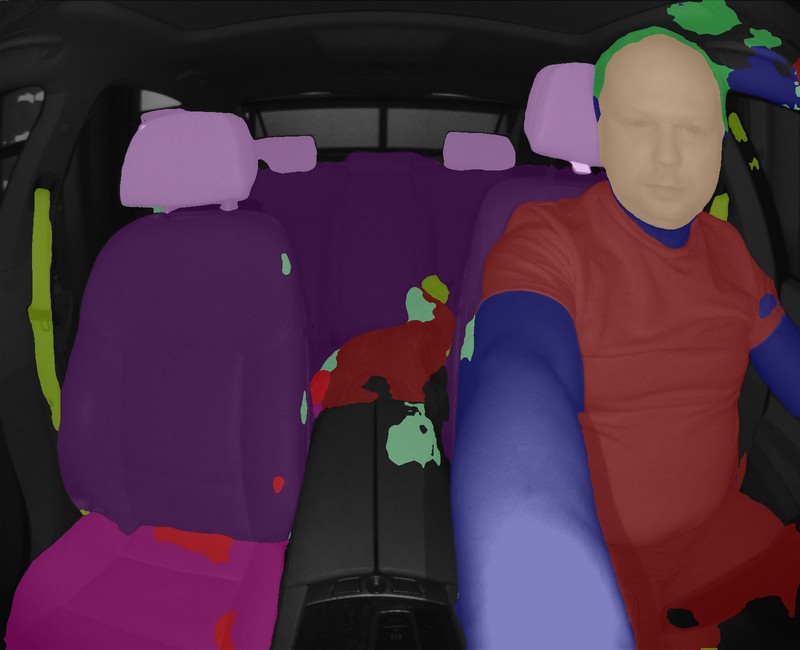} \\
        \noalign{\vskip 0.5pt}        
        \rotatebox[origin=c]{90}{\scriptsize \textbf{Exterior}} &
        \imgcellext{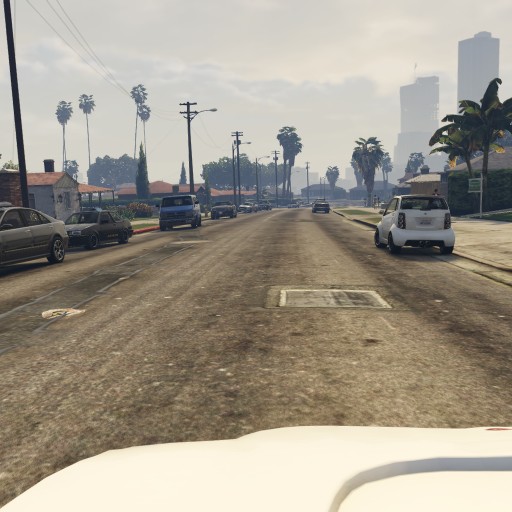} &
        \imgcellext{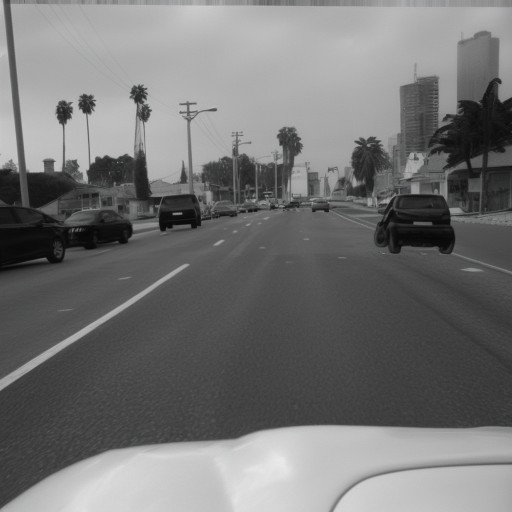} &
        \imgcellext{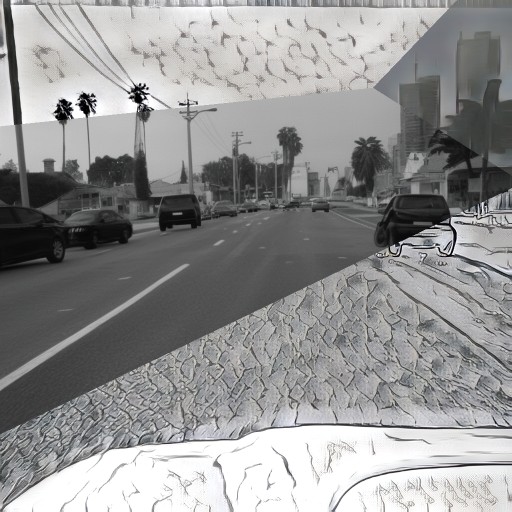} &
        \imgcellext{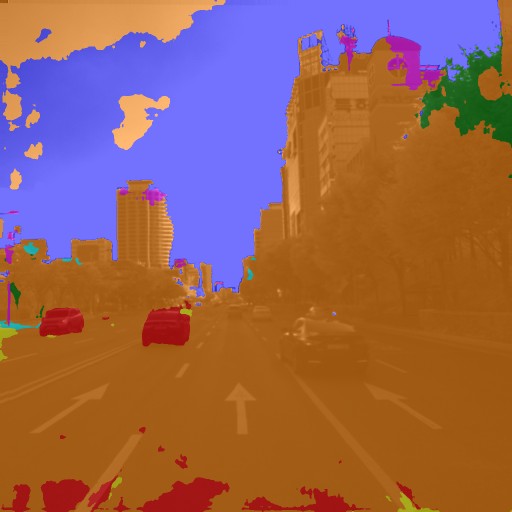} &
        \imgcellext{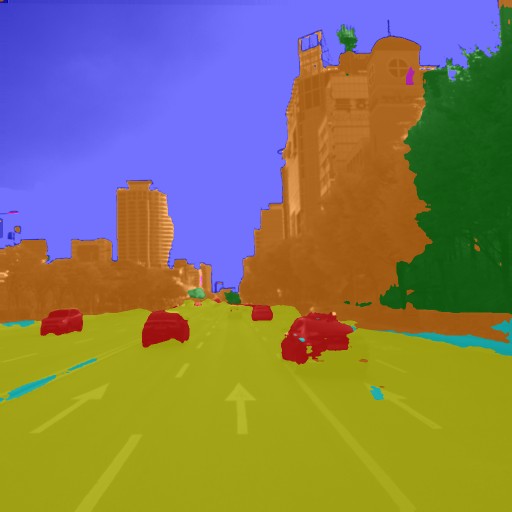} \\

        & \scriptsize (a) Synthetic Image & \scriptsize (b) + TSA & \scriptsize (c) + Style Transfer (VSD) & \scriptsize (d) Baseline & \scriptsize (e) Ours \\

    \end{tabular} 
    \caption{
    (a)-(c) show the progressive training data augmentation pipeline. 
    (d) \& (e) show a blend of original image and prediction for a qualitative comparison of the DeepLabV3+ inference when trained on baseline synthetic data from (a) compared to 
    augmented images given in (c).}
    \label{fig:teaser}
\end{figure*}
Modern vehicle systems increasingly rely on visual perception to ensure safe and reliable operation. 
Advanced driver assistance systems and automated driving functions require a comprehensive understanding of both the external environment and the condition of occupants inside the vehicle. 
While external perception facilitates tasks such as obstacle detection, interior monitoring is crucial for assessing the driver's state and preventing accidents caused by distraction or fatigue. 
EU Regulation~2019/2144 mandates driver drowsiness and distraction warning systems for all new vehicles from July~2024~\cite{EURegulation2019}, and ISO/PAS~8800~\cite{ISOPAS8800} sets safety requirements for automotive AI, reflecting a growing demand for robust 
scene understanding.
A core component of visual perception 
is semantic segmentation\cite{xie2021segformer},
which assigns a class label to every pixel in an image. 
This supports downstream tasks for interior sensing, such as
occupant detection and localization~\cite{katrolia2021ticam}, 
as well as general exterior scene parsing for automated driving~\cite{Cordts2016Cityscapes}. Near-infrared (NIR) imaging is particularly well suited in automotive scenarios, as NIR sensors operate at wavelengths invisible to the human eye, and deliver consistent image contrast across the full day-night cycle~\cite{Cheng2019Non-Local,choe2018ranus}. A more detailed description of the NIR-spectrum and its applications in automotive contexts is provided in App.~\ref{sec:nir}. 
The practical application of segmentation models is constrained by the availability of large amounts of accurately annotated training data. 
While extensive annotated datasets exist for RGB imagery, comparable resources for NIR data remain limited. In principle, similar datasets could be created for NIR as well,  but creating pixel-level annotations is especially labor-intensive and costly, due to the need for manual labeling by human experts and careful quality control. At the same time, NIR imaging is used in more niche application domains, which limits the incentives for building large-scale annotated datasets. 
Synthetic data generated from simulation environments offers a highly scalable alternative, enabling the creation of densely labeled training data at negligible marginal cost, making it feasible to expand dataset size without a substantial increase in expenses~\cite{richter2016playing}. 
Yet models trained on synthetic imagery routinely suffer substantial performance degradation when evaluated on real data, a phenomenon known as the \textit{domain gap}~\cite{torralba2011datasetbias}.
These discrepancies are amplified in the NIR range, where synthetic simulators are almost exclusively designed for the RGB spectrum, introducing an additional spectral discrepancy with no established adaptation baseline. Closing this \emph{synthetic-to-real} (synth2real) gap without access to large quantities of labeled real NIR data is, therefore, a prerequisite for practical deployment.
Beyond the domain gap between synthetic and real data, a key challenge stems from the inductive biases learned by deep networks. Rather than learning robust and transferable representations, models often exploit superficial statistical regularities in the training data \cite{geirhos2020Shortcutlearninga}. In particular, models trained on synthetic or stylistically homogeneous data tend to rely on local texture patterns instead of global shape cues, a phenomenon known as \emph{texture bias} \cite{geirhos2018imagenet}. This becomes especially problematic in synth2real settings, where texture statistics strongly depend on rendering style and sensor modality \cite{cohen2025TextureSAMTexture}, causing models trained on synthetic data to generalize poorly to real-world images.
In contrast, shape-based representations are generally more stable across domains, e.g. 
stylistic diversification during training improves the texture-shape balance
~\cite{hamscher2025TransferringStyles}. 
However, systematic investigations of texture and shape bias in dense prediction tasks remain limited~\cite{mutze2025InfluenceShape,heinert2025shapebiasrobustnessevaluation}. 
NIR images exhibit different texture characteristics than RGB while preserving 
edges and large-scale structures~\cite{Cheng2019Non-Local}, potentially 
increasing shape reliance. 
In this work, we present a systematic investigation of synth2real domain adaptation for semantic segmentation in the NIR spectrum. 
We introduce a generative augmentation framework designed to improve cross-domain generalization by explicitly compensating for inductive biases during training (see Fig.~\ref{fig:teaser}).  
This design enables realistic variations in appearance aligned with the original annotations by combining a multi-conditioned low-rank adaptation~\cite{hu2022lora} and Voronoi-based style diversification (VSD).
Evaluations are performed using 
representative semantic segmentation models, DeepLabV3+~\cite{chen2018encoder}, SegFormer~\cite{xie2021segformer}, and Mask2Former~\cite{cheng2022masked}, that capture convolutional, transformer-based, and hybrid design paradigms. 
Experiments on interior perception are conducted using a proprietary in-cabin NIR dataset. For road-scenes, we use the publicly available RANUS dataset~\cite{choe2018ranus} as the real target domain and GTA5~\cite{richter2016playing} as the synthetic source domain. 
\noindent Summarized, our contributions are as follows:
\begin{itemize}
    \item We provide, to the best of our knowledge, the first comprehensive investigation of synth2real transfer for semantic segmentation in automotive NIR imagery, covering both interior and exterior perception scenarios.
    \item We propose a structure-preserving generative augmentation pipeline that transfers synthetic images into realistic NIR-style variants, enabling effective training without annotated real data.
    \item We investigate how controlled style variations influence texture-shape bias and show that balancing biases improves robustness and cross-domain generalization in semantic NIR segmentation.
    \item 
    We validate the proposed approach using multiple segmentation architectures and datasets, including proprietary interior data and public benchmarks for street scenes. 
\end{itemize}

\section{Related Work}
\subsection{Texture and Shape Bias in Vision Models.}
Previous investigations show that NIR imaging may alter the representation of fine texture and color relative to RGB, but preserves edges and large-scale structure \cite{Cheng2019Non-Local}, potentially encouraging greater shape reliance. 
In the specific domain of infrared small‑target detection, characterized by dim texture, complex backgrounds, and variable target shapes, state‑of‑the‑art detectors emphasize shape cues, by employing large‑kernel convolutions and edge/contour supervision, to improve detection under low‑texture conditions \cite{Lin2024LearningContrastEnhanced,Lin2024LearningShapeBiased}.
Beyond the NIR domain, modern deep learning architectures often bypass high-level semantic understanding in favor of shortcut learning, as shown by Geirhos et al.~\cite{geirhos2020Shortcutlearninga}. This over-reliance on spurious correlations, such as local texture cues, renders models brittle when subjected to domain shifts, as observed by Cohen et al.~\cite{cohen2025TextureSAMTexture}. 
While convolutional neural networks trained on ImageNet tend to be more texture biased \cite{geirhos2018imagenet}, Vision transformer models show greater shape sensitivity when evaluated with cue-conflicting images \cite{naseer2021intriguing}, though results vary with architectures, training data, augmentations and application domains  \cite{naseer2021intriguing, burgert2026imagenettrained} 
and differ slightly across evaluation protocols \cite{ geirhos2018imagenet, burgert2026imagenettrained, heinert2025shapebiasrobustnessevaluation}. 
Deliberately manipulating inductive biases via data stylization or auxiliary supervision can improve accuracy or robustness to common corruptions and align closer with target domains \cite{ kwak2019ImpactTexture,hamscher2025TransferringStyles}. 
While texture bias has been extensively studied for image classification \cite{geirhos2018imagenet,hermann2020origins}, 
systematic investigations of dense prediction tasks such as semantic segmentation remain underexplored.
\subsection{Domain Adaptation and Generalization.} 
Domain shifts occur when training (source) and test (target) data follows different distributions.
In semantic segmentation, domain gaps are commonly caused by
appearance shift, sensor/pipeline shift, geometry/viewpoint shift, context shift, as well as dataset bias stemming from benchmark-specific collection and labeling conventions \cite{torralba2011datasetbias}.
Unsupervised domain adaptation (UDA) aims to bridge the gap between source and target domains through three common strategies. First, \emph{appearance alignment} transfers low‑frequency appearance statistics from target to source \cite{yang2020fda}. Second, \emph{representation alignment} leverages output‑space adversarial training and entropy minimization \cite{Rahul2025}. Third, \emph{self‑training} relies on pseudo labels to iteratively refine model predictions \cite{hoyer2023mic}. 
As Hoyer et al. have shown, for example, transformer backbones outperform CNN baselines for semantic segmentation with UDA (e.g.\ \cite{hoyer2023mic}). While UDA assumes unlabeled target data during training, domain generalization (DG) aims to learn from source domains only and generalize to unseen targets, posing a strictly harder problem.
Prior work by Liu et al. has focused on feature alignment and invariant representation learning, in which networks are trained to encode features independently of the domain \cite{LIU2023117}.
Concurrently, data augmentation strategies emerged, including style transfer, to increase the diversity of data and thus promote generalization \cite{zhao2022shade}. 
Recently, Jia et al. used generative diffusion models to synthesize diverse, semantically controlled images to train domain generalizable segmentation models, thereby improving their robustness across multiple real-world datasets \cite{jia2024dginstyle}. 

Although UDA and DG have been extensively studied for RGB semantic segmentation, neither of these methods takes into account the specific characteristics of NIR images. 
While UDA for NIR remains relatively underexplored, some works have begun to address this gap. Chen et al. used contour features to transfer segmentation knowledge from visible to thermal infrared urban scenes \cite{chen2022light}. Furthermore, Pandey et al. proposed a generative latent search approach to find source domain proxies to guide segmentation of unlabeled NIR images \cite{pandey2020skin}.
Moreover, to the best of our knowledge, no work benchmarks shape versus texture bias in NIR analogous to Stylized‑ImageNet for RGB~\cite{geirhos2018imagenet}. In contrast, our work systematically investigates synth-to-real domain adaptation specifically for automotive NIR semantic segmentation, combining generative augmentation, structure-preserving conditioning, and bias-balancing strategies. Unlike the approaches by Chen~\cite{chen2022light} and Pandey ~\cite{pandey2020skin}, we do not rely on visible-spectrum correspondence and nearest-clone optimization. Instead, we explicitly address the role of balancing inductive biases in improving cross-domain robustness.

\section{Method}
\begin{figure}[t]
\centering
\includegraphics[width=\linewidth]{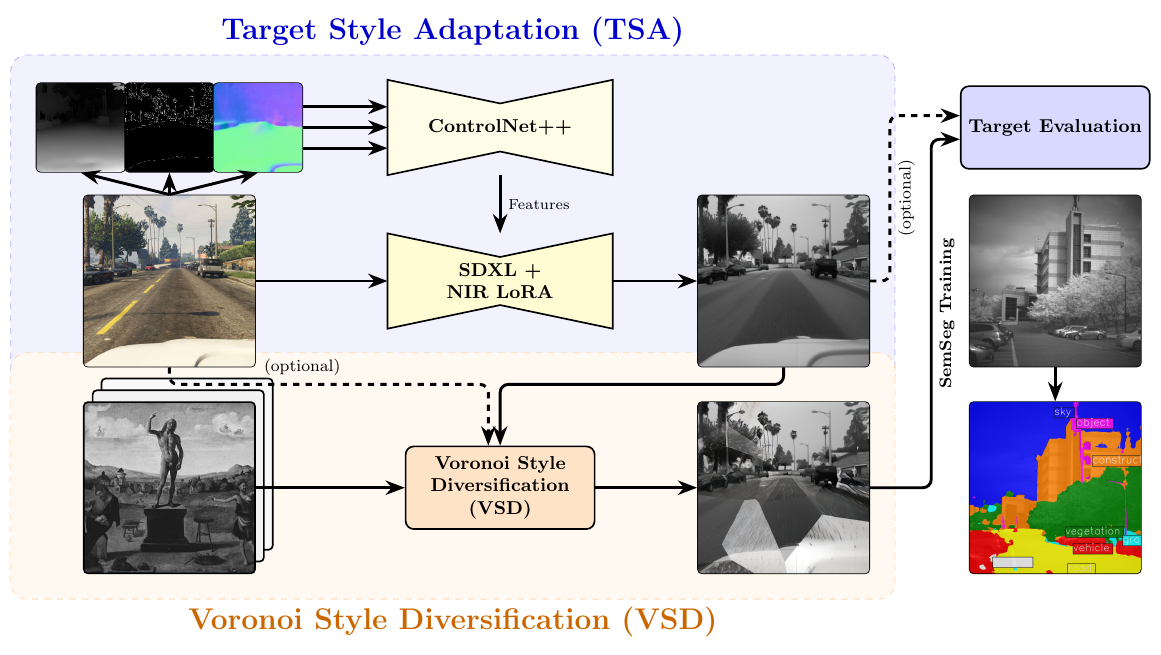}

\caption{Our method augments data from a source domain, applies a target style adaptation via an LDM, and corrupts the texture by Voronoi style diversification. The resulting augmented dataset is utilized to train a semantic segmentation model which generalizes to real (target) data. Dashed arrows indicate optionally applying only one modular component.}
\label{fig:method}
\end{figure}

We describe our modular generative augmentation framework that transforms synthetic images via target style adaptation (TSA) into realistic NIR-style variants, as well as the Voronoi-based style diversification strategy (VSD) to achieve more robust semantic segmentation performance. A schematic overview of our method is shown in Fig.~\ref{fig:method}.

\paragraph{Target Domain (NIR) Style Adaptation (TSA).}
Bridging the appearance gap between RGB source images and real NIR imagery requires a generative component capable of realistic modality translation. We therefore employ a fine-tuned Latent Diffusion Model (LDM)~\cite{rombach2022high} as the first module of our pipeline. 
LDMs formulate image synthesis as a sequential denoising process while reducing computational cost by operating in the latent space of an autoencoder. Conditioning signals such as images, or semantic maps can be incorporated through cross-attention layers in the UNet. Stable Diffusion~\cite{podell2023sdxl}, a widely used LDM, leverages this mechanism to align textual and visual representations. Fine-tuning a full diffusion model requires prohibitive amounts of labeled target data. We therefore train our own LoRA~\cite{ryu2022lora}, which injects low-rank matrices into the attention layers, enabling custom adaptation with only a small number of unlabeled target images.
Note, that applying this adaptation technique results in UDA, while omitting TSA corresponds to DG.
To mimic the domain-specific appearance characteristics of the target domain, we train a LoRA on a curated subset of a few dozen real target-domain images.
This enables efficient adaptation to the NIR modality. To support the LoRA training, image captions are automatically generated using a vision–language model (see App. \ref{subsec:loratrain}).
A key challenge during diffusion-based image translation is the preservation of small and thin semantic structures, which can easily be degraded or removed during the denoising process, especially critical for safety-relevant classes such as glasses or seat belts.
To address this issue, we employ a multi-signal conditioning strategy using ControlNet~\cite{xinsir6controlnetplus,zhao2023uni,zhang2023adding} to constrain the generation process. Fig.~\ref{fig:lora} and Fig.~\ref{fig:lora_exterior} in App.~\ref{sec:traindetails} illustrate examples of conditioning signals and the resulting TSA. The segmentation mask itself is provided as conditioning to enforce consistency between the generated appearance and the semantic layout. Canny edges extracted from the input image provide structural contours of the scene, while additional edges derived from the segmentation masks directly prevent inter-class bleeding and preserve class boundaries during the diffusion process. The style adaptation process mimics the target domain while keeping the content consistent with the source domain and its segmentation mask. 
All conditioning signals are derived exclusively from the source domain, ensuring that no target-domain supervision is incorporated. 
\paragraph{Style Diversification Method.}
While TSA reduces the global appearance gap between synthetic and real data, models trained on visually homogeneous datasets may still overfit to specific texture statistics and fail to generalize under domain shift. To mitigate this, we introduce a style diversification step that deliberately perturbs local texture patterns to encourage models to rely more on structural cues rather than dataset-specific textures, thereby improving robustness and cross-domain generalization.
For parsing the style and texture of a given style image to those of a content image while adhering to the original
content's structure, we adopt the lightweight Adaptive Instance Normalization (AdaIN,\cite{huang_arbitrary_2017}) approach. 
AdaIN enables efficient style transfer without retraining and reliably preserves geometric scene content, making it well suited for pixel-level tasks such as semantic segmentation. Compared to more recent stylization approaches, it provides a favorable trade-off between computational efficiency, controllability, and structural fidelity~\cite{cho_one-shot_2024, kwon2023diffusionbasedimagetranslationusing}.
AdaIN uses the initial four layers from a VGG-19 network~\cite{simonyan2015deepconvolutionalnetworkslargescale} pretrained on ImageNet as a fixed feature extractor $f$. Given content input $c\in \mathbb R^{C\times H \times W}$ and style input $s\in \mathbb R^{C\times H_\text{style}\times W_\text{style}}$, the AdaIN operation normalizes the content features to match the statistical properties of the style features:
\begin{align}
    z =\sigma (f(s))\left( \frac{f(c)-\mu(f(c))}{\sigma(f(c))}\right)+\mu(f(s)) \enspace .
\end{align}
As the employed style transfer operates on RGB images, we further apply a grayscale conversion to suppress color cues and obtain an appearance closer to NIR imagery. 
Motivated by Hamscher et al. \cite{hamscher2025TransferringStyles}, we decompose each image into locally coherent but irregular regions via Voronoi partitioning, applying independent stylization to each cell. This partitioning further supports stochastic yet spatially coherent texture modifications, as both the number of Voronoi cells and the probability of stylizing each cell can be adjusted to control the density and distribution of style perturbations across the image.
In the case of stylization, we sample a style from the Painter by Numbers dataset~\cite{painterbynumbers2016} and apply it via AdaIN to the corresponding cell. 
This setup allows us to impose spatially varying texture changes while preserving the global structure of the original scene. 
\section{Experiments}
We evaluate our generative augmentation framework for synth2real NIR segmentation across interior and exterior automotive scenarios. A core premise of our approach is that synthetic-only training yields texture-dominated models that transfer poorly to real NIR imagery and that controlled generative augmentation can rebalance this inductive bias toward shape, thereby narrowing the synth2real gap without any annotated real data. The following experiments are designed to test this and to verify that the observed improvements stem from genuine texture-shape rebalancing rather than from overfitting to synthetic artifacts. After detailing the datasets, architectures, and augmentation hyperparameters, we quantify the reduction of the synth2real domain gap on real-world data and analyze the induced texture-shape bias and corruption robustness to probe the underlying mechanism.
\subsection{Experimental Setup}
\paragraph{Datasets.}
Our interior sensing dataset 
contains coarse- and fine-grained semantic classes covering vehicle components and human-centric elements (see App. \ref{app:interior} for details), enabling reasoning about occupant presence, pose, and interactions. Safety-critical elements such as seat belts and potentially left-behind children or objects are included. The dataset is challenging due to high visual diversity, frequent occlusions, and small or thin classes such as glasses or seat belts. It comprises 4,341 synthetic images (train/val: 3,501/840) and 1,151 real-world images (train/val: 924/227).

Our exterior sensing datasets consist of synthetic GTA5~\cite{richter2016playing} and real-world RANUS~\cite{choe2018ranus} images. They contain semantic classes covering urban infrastructure and traffic participants, enabling reasoning about the vehicle's surroundings, autonomous navigation, and obstacle avoidance. These datasets are challenging due to highly dynamic scenes, complex illumination, and significant domain gaps. GTA5 provides RGB frames from a virtual city, capturing diverse scenes from different perspectives. RANUS contains co-registered RGB–NIR frames captured in Seoul, South Korea, introducing both synth2real and geographic domain shifts. Both datasets are mapped to eight shared semantic classes and filtered for quality (see App. \ref{app:exterior} for details), resulting in 6,766 synthetic 
images (train/val: 2,960/3,806) and 3,772 real-world pairs (train/val: 2,796/976).

\begin{figure}[t]
  \newcommand{\w}{0.194\linewidth} 

  \centering
  \begin{subfigure}[t]{\w}
    \centering
    \includegraphics[width=\linewidth]{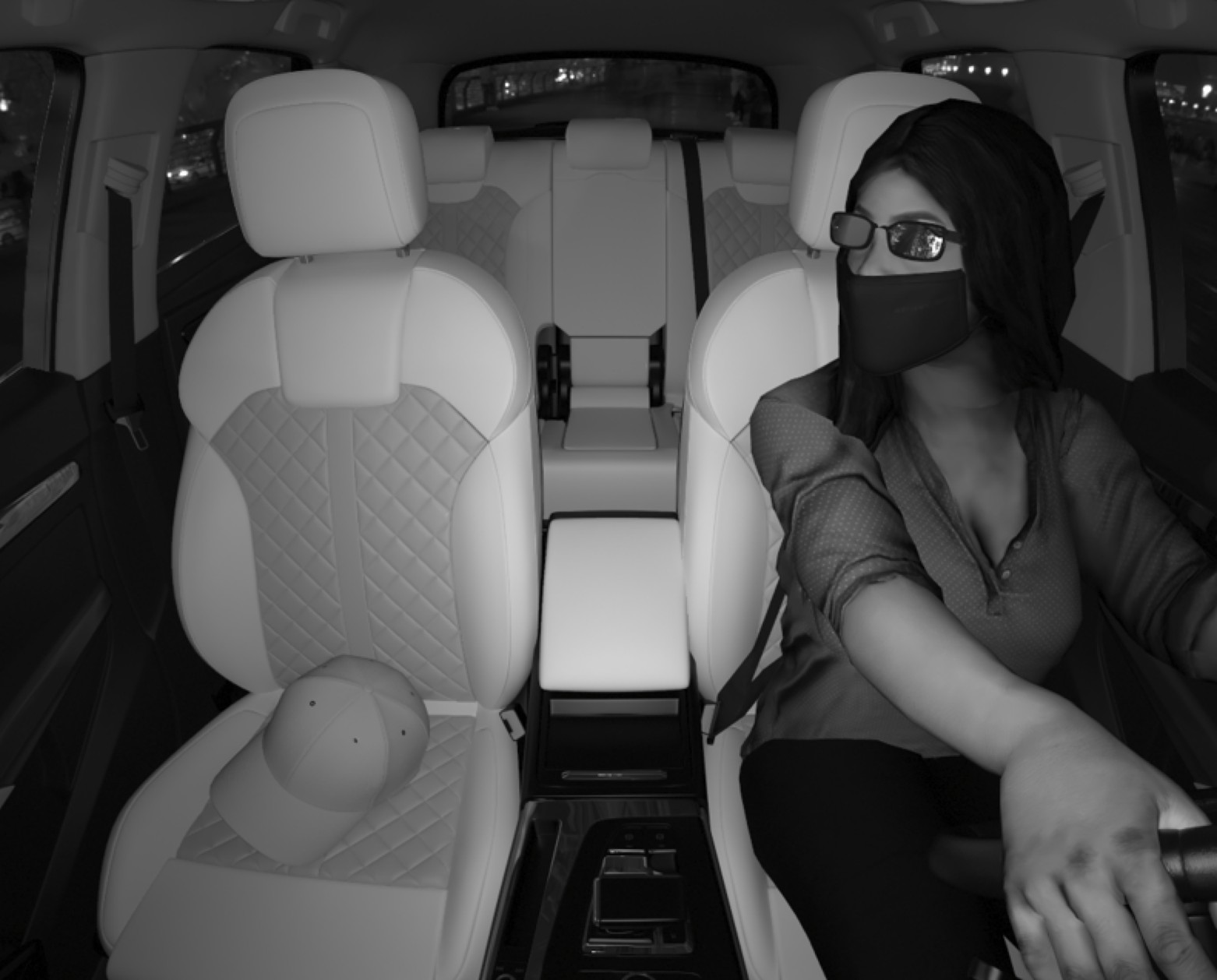}
    \caption{Original Synthetic Image}
  \end{subfigure}\hfill
  \begin{subfigure}[t]{\w}
    \centering
    \includegraphics[width=\linewidth]{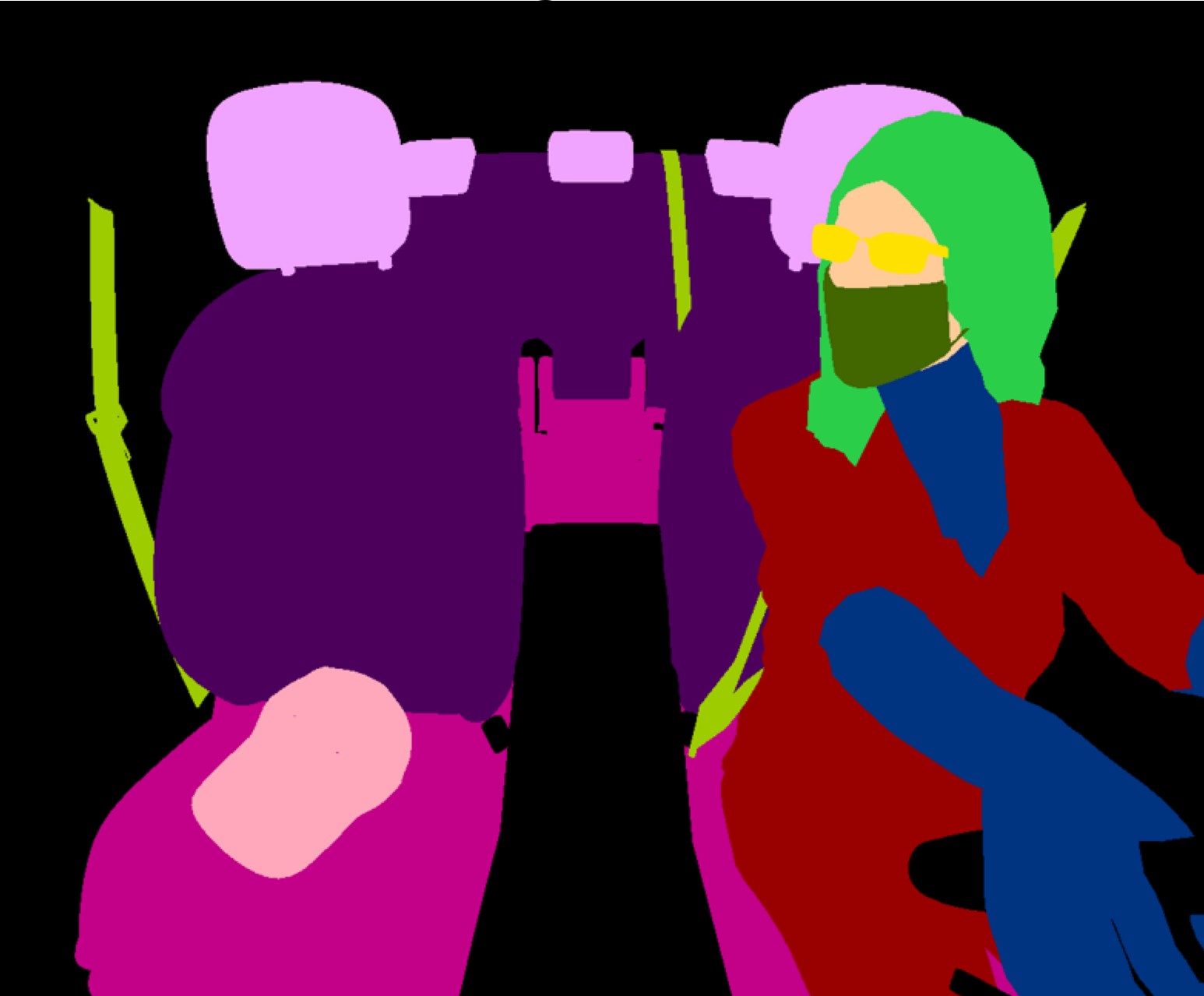}
    \caption{Semantic Segmentation}
  \end{subfigure}\hfill
  \begin{subfigure}[t]{\w}
    \centering
    \includegraphics[width=\linewidth]{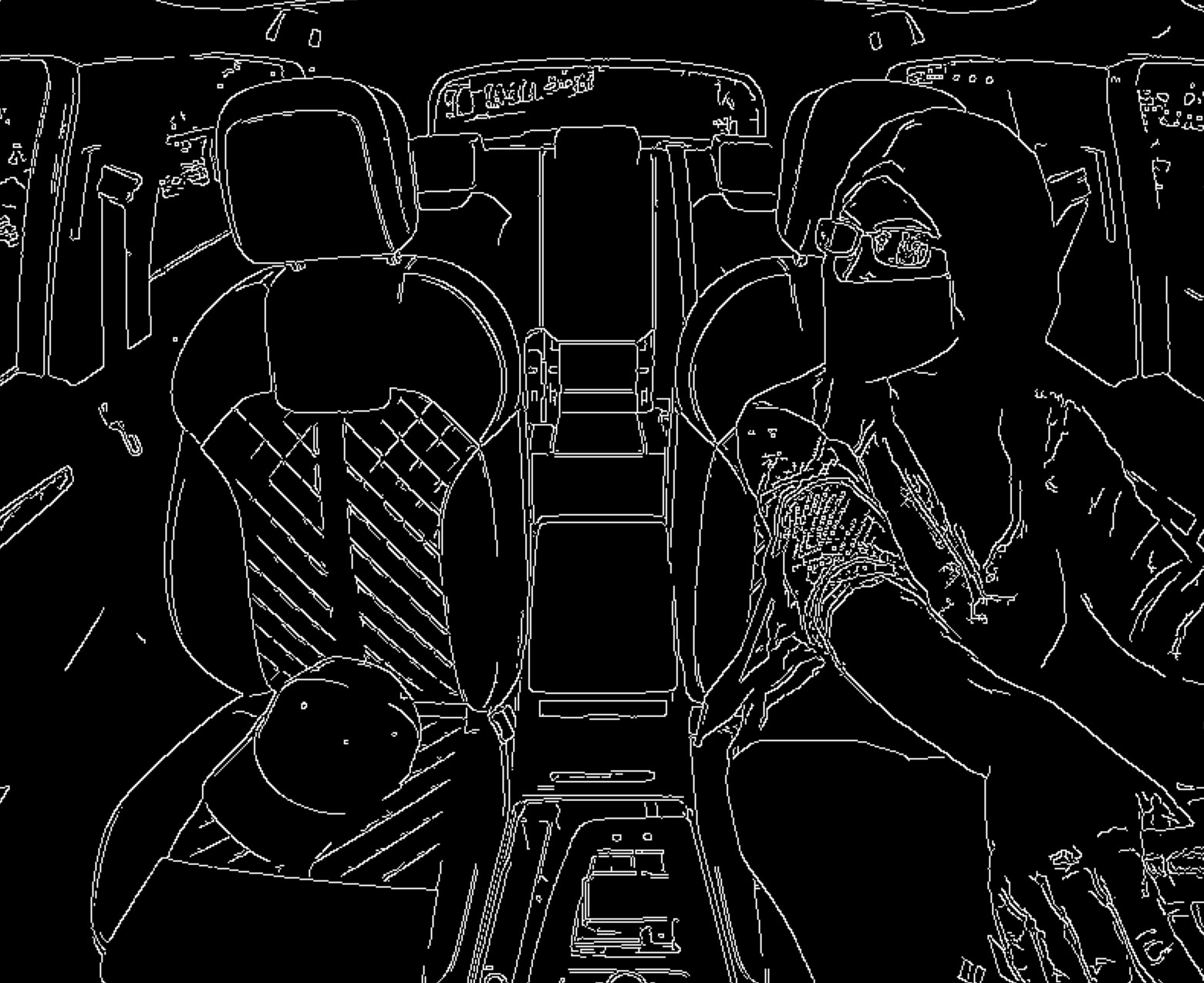}
    \caption{Canny (Image)}
  \end{subfigure}\hfill
  \begin{subfigure}[t]{\w}
    \centering
    \includegraphics[width=\linewidth]{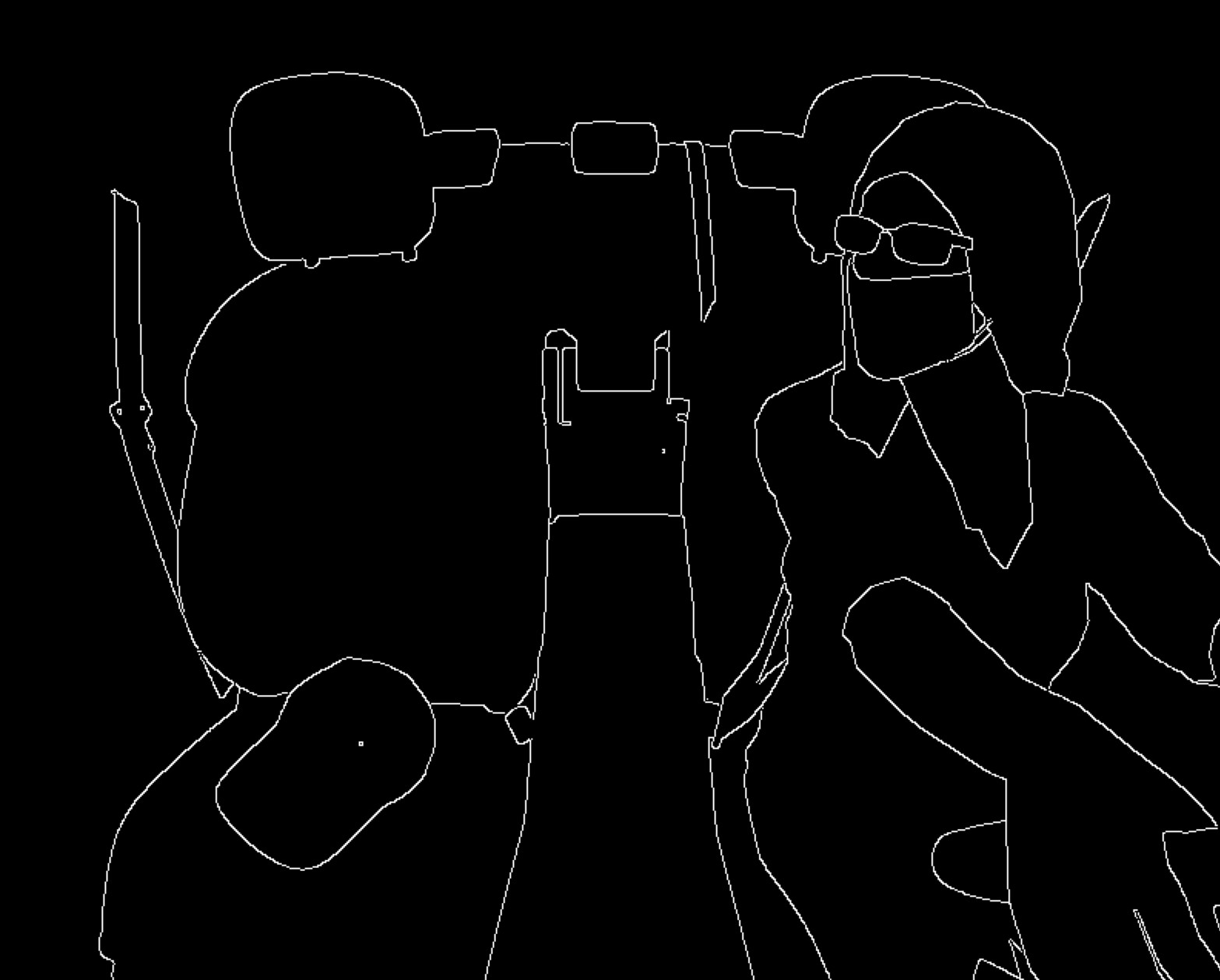}
    \caption{Canny (Segmentation)}
  \end{subfigure}\hfill
  \begin{subfigure}[t]{\w}
    \centering
    \includegraphics[width=\linewidth]{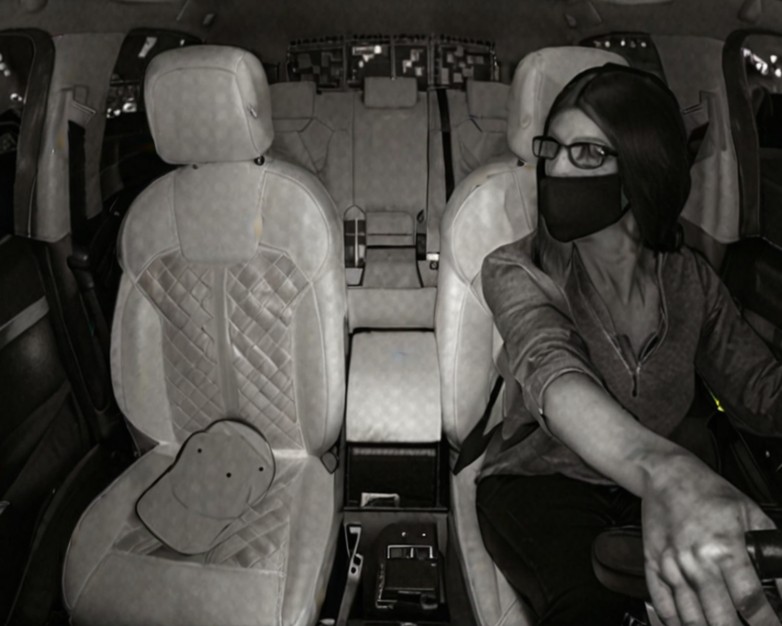}
    \caption{TSA}
  \end{subfigure}
  \caption{TSA with multiple conditioning signals. 
  }
  \label{fig:lora}
\end{figure}
\paragraph{Hyperparameters for NIR Target Style Adaptation.}
As LDM, we employ Stable Diffusion XL (SDXL~\cite{podell2023sdxl}), which enables higher‑quality image generation. For the interior setup, the LoRA hyperparameter study resulted in an inference configuration with 30 denoising steps, denoising strength 0.5, conditioning weights of 1.0 for Canny masks, 0.5 for segmentation, and 0.5 for Canny images (see Fig.~\ref{fig:lora}). For the exterior, qualitative evaluation led to 50 steps, strength 0.99, and conditioning weights of 0.9 (Canny mask), 0.2 (depth), and 0.3 (surface normals), see App.~\ref{sec:traindetails}, Fig.~\ref{fig:lora_exterior}. The differences arise from the RGB to NIR translation task and higher variability in exterior scenes. Canny edges on segmentation masks proved critical for preserving structure and preventing instance merging, while depth and normal conditioning mainly benefited exterior scenes.
\paragraph{Hyperparameters for Stylization Method.}
\begin{figure*}[t]
  \newcommand{\w}{0.194\linewidth} 

    \centering
    \begin{subfigure}[t]{\w}
        \centering
        \includegraphics[width=\textwidth]{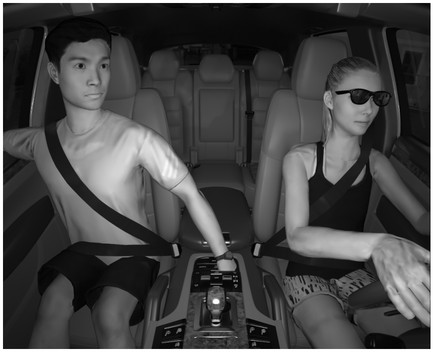}
        \caption{Synthetic Baseline}
        \label{fig:av_synthetic_baseline}
    \end{subfigure}
    \hfill
    \begin{subfigure}[t]{\w}
        \centering
        \includegraphics[width=\textwidth]{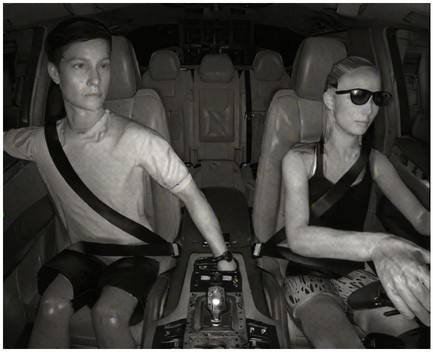}
        \caption{TSA}
        \label{fig:av_lora}
    \end{subfigure}
    \hfill
    \begin{subfigure}[t]{\w}
        \centering
        \includegraphics[width=\textwidth]{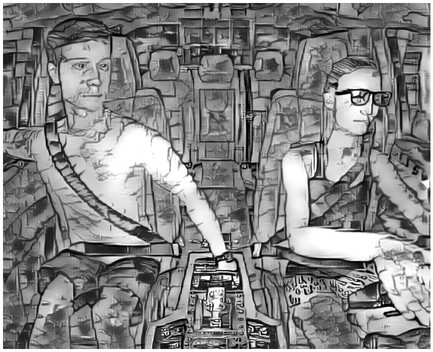}
        \caption{VSD(1,1.0)}
        \label{fig:av_Full_Stylized_p=1.0}
    \end{subfigure}
    \hfill
    \begin{subfigure}[t]{\w}
        \centering
        \includegraphics[width=\textwidth]{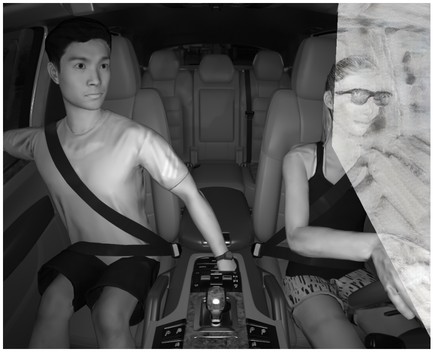}
        \caption{VSD(4,0.25)}
        \label{fig:av_Voronoi4_p=0.25}
    \end{subfigure}
    \hfill
    \begin{subfigure}[t]{\w}
        \centering
        \includegraphics[width=\textwidth]{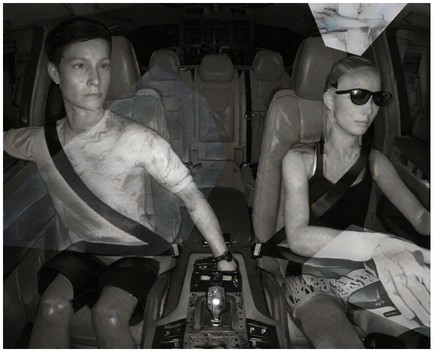}
        \caption{VSD(16,0.25) + TSA}
        \label{fig:av_Voronoi4_p=0.25_plusLoRA}
    \end{subfigure}
    \caption{
    Selection of augmentation variants generated with our method (all in App. Fig. \ref{fig:overview_datasets_full}).
    }
    \label{fig:overview_datasets}
\end{figure*}
Fig.~\ref{fig:overview_datasets} depicts the generated training datasets, which vary in transformation intensity and spatial structure to systematically analyze the impact of stylization on domain generalization. These include Voronoi-based style diversification VSD(n,p) with $n \in \{4, 8, 16\}$ cells combined with stylization probabilities $p\in \{0.25,0.75\}$, a fully stylized dataset VSD(1,1.0) with randomly sampled styles, and sequential variants applying TSA followed by stylization. Two baselines are included: the original synthetic source dataset and the real target dataset, resulting in 28 datasets in total (see App. \ref{subsec:hyperparamter} for full overview). To obtain preliminary insights, we evaluate stylization effects using pixel-based (e.g. PSNR, SSIM), perceptual (LPIPS, DISTS), and distributional (FID) metrics. While TSA-only transformations remain close to the synthetic baseline, stronger stylization variants introduce larger appearance shifts.
\paragraph{Semantic Segmentation Models.}
We evaluate three architectures representing different design families: DeepLabV3+ (convolutional neural network (CNN), ResNet-50 backbone, \cite{chen2018encoder}), SegFormer (Transformer, MiT-B5 backbone, \cite{xie2021segformer}), and Mask2Former (high-capacity transformer, Swin-Large backbone, \cite{cheng2022masked}). This selection allows us to study how architectural choices influence model biases and robustness while leveraging pretrained weights for competitive performance. 
Training details are given in App. \ref{subsec:training}.
\subsection{Experimental Results: Semantic Segmentation Performance}
\begin{table}[t]
\centering
\caption{mIoU on real-world validation set (median 
over 3 runs, $\pm$ shows distance to min/max). VSD($n,p$) with $n$: Voronoi cells and $p$: stylization probability. Best \textbf{bold} (full in App. Table~\ref{tab:full_main_results_3runs}).
}
\label{tab:main_results_3runs}
\setlength{\tabcolsep}{2pt}
\begin{tabular}{l ccc ccc}
\toprule
& \multicolumn{3}{c}{\textbf{Interior}}
& \multicolumn{3}{c}{\textbf{Exterior}} \\
\cmidrule(lr){2-4}\cmidrule(lr){5-7}
\textbf{Method} & DeepLabV3+ & SegFormer & Mask2Former & DeepLabV3+ & SegFormer & Mask2Former \\
\midrule
Real Baseline                  & $75.12^{+0.40}_{-2.15}$ & $81.71^{+0.57}_{-0.67}$ & $82.87^{+0.05}_{-0.14}$ & $66.99^{+0.78}_{-0.10}$ & $62.55^{+0.83}_{-1.76}$ & $73.48^{+0.02}_{-0.32}$ \\
Synthetic Baseline                     & $32.00^{+1.11}_{-2.31}$ & $49.18^{+2.79}_{-0.13}$ & $56.79^{+1.78}_{-0.67}$ & $21.57^{+1.39}_{-0.11}$ & $28.14^{+1.44}_{-2.34}$ & $58.95^{+0.46}_{-0.06}$ \\
\midrule
TSA Only                      & $35.18^{+3.17}_{-0.14}$ & $53.97^{+1.69}_{-0.02}$ & $58.15^{+0.08}_{-0.56}$ & $44.95^{+1.30}_{-0.14}$ & $41.84^{+0.22}_{-2.86}$ & $55.26^{+1.12}_{-1.02}$ \\
VSD(1,1.0)                  & $29.72^{+1.86}_{-1.13}$ & $54.25^{+0.60}_{-0.01}$ & $\textbf{62.04}^{+0.37}_{-0.44}$ & $41.59^{+2.95}_{-2.79}$ & $40.53^{+0.20}_{-1.14}$ & $\textbf{61.91}^{+0.39}_{-0.04}$ \\
VSD(8, 0.25) + TSA   & $43.92^{+0.30}_{-0.44}$ & $56.75^{+0.77}_{-0.78}$ & $60.13^{+0.54}_{-0.26}$ & $\textbf{50.44}^{+0.87}_{-0.69}$ & $42.02^{+1.38}_{-0.10}$ & $57.16^{+1.06}_{-0.97}$ \\
VSD(16, 0.25) + TSA             & $\textbf{44.26}^{+0.93}_{-0.41}$ & $\textbf{57.30}^{+0.08}_{-0.32}$ & $60.55^{+0.77}_{-0.74}$ & $48.31^{+0.03}_{-1.76}$ & $42.44^{+2.99}_{-0.61}$ & $56.29^{+0.26}_{-1.84}$ \\
VSD(16,0.75) + TSA             & $43.62^{+0.70}_{-2.17}$ & $54.32^{+1.09}_{-0.94}$ & $60.62^{+0.99}_{-0.84}$ & $46.36^{+0.51}_{-0.53}$ & $\textbf{43.69}^{+0.32}_{-1.66}$ & $55.35^{+0.30}_{-1.58}$ \\
\bottomrule
\end{tabular}
\end{table}

\paragraph{Overall Domain Gap.}
We evaluate our approach against two natural baselines, real-domain (upper bound) and synthetic only training. Our VSD stylization is applied with varying cell counts and stylization probabilities, 
with and without additional TSA fine-tuning. Moreover, we provide TSA only and VSD(1,1.0) variants. 
Table \ref{tab:main_results_3runs} reports results showing that incorporating 
VSD (together with TSA) 
improves real-world generalization. 
For the interior setting, our VSD + TSA variants lead to substantial gains in comparison to the synthetic baseline. The best overall performance is obtained with 16 cells and 25\% probability.
In the exterior evaluations on the RANUS NIR data, 
we do not identify a single VSD variant that generally performs best. However, we observe that our method outperforms the synthetic baseline, in particular improving the median mIoU for the DeepLabV3+ from 21.57 to 50.44. 
As with the interior dataset, Mask2Former achieves the highest mIoU values with the fully stylized variant. 
An interesting observation is that the Mask2Former model demonstrates strong robustness to spectral variations. When evaluated on real RANUS RGB images, its mIoU exceeds the NIR performance by only 2.27 points, indicating a comparatively low reliance on spectral cues. In contrast, the performance gap is notably larger for DeepLabV3+ and SegFormer, suggesting that architectural design plays a significant role in cross-spectral generalization (see App. Table~\ref{tab:spectral_dependence}).
To investigate whether these improvements in performance are due to overfitting of the domain, we also present results from our mid-training evaluation (see App. Table~\ref{tab:30k_results_3runs}). We observe that
the gains appear early and remain stable, indicating that the method consistently improves domain robustness rather than exploiting transient artifacts. Qualitative comparisons in Fig. \ref{fig:models-per-column} further demonstrate the real-world benefit. Learned class features appear more robust and predictions show fewer artifacts compared to the baseline, which results in a general improvement of mIoU.
\begin{table}[t]
\centering
\caption{Domain gap and gap closed (median mIoU) for interior and exterior data.}
\label{tab:domain_gap_interior}
\setlength{\tabcolsep}{2pt}
\renewcommand{\arraystretch}{1.05}
\begin{tabular}{l c c c l c c c c}
\toprule
\textbf{Arch.} & \textbf{Real} & \textbf{Synth.} & \textbf{Best} & \textbf{Best Method}
& \textbf{Gap} & \textbf{Rem. Gap} & \textbf{Gap Closed} & \textbf{Closed (\%)} \\
\midrule
\multicolumn{9}{l}{\textit{Interior}} \\
DeepLabV3+   & 75.12 & 32.00 & 44.26 & VSD(16,0.25) + TSA & 43.12 & 30.86 & 12.26 & 28.4 \\
SegFormer    & 81.71 & 49.18 & 57.30 & VSD(16,0.25) + TSA & 32.53 & 24.41 & 8.12  & 25.0 \\
Mask2Former  & 82.87 & 56.79 & 62.04 & VSD(1,1.0)         & 26.08 & 20.83 & 5.25  & 20.1 \\
\midrule
\multicolumn{9}{l}{\textit{Exterior}} \\
DeepLabV3+   & 66.99 & 21.57 & 50.44 & VSD(8,0.25) + TSA  & 45.42 & 16.55 & 28.87 & 63.6 \\
SegFormer    & 62.55 & 28.14 & 43.69 & VSD(16,0.75) + TSA & 34.41 & 18.86 & 15.55 & 45.2 \\
Mask2Former  & 73.48 & 58.95 & 61.91 & VSD(1,1.0)         & 14.53 & 11.57 & 2.96  & 20.4 \\
\bottomrule
\end{tabular}
\end{table}
Table \ref{tab:domain_gap_interior} summarizes how the different augmentation strategies reduce the synth2real domain gap for interior NIR segmentation. The synthetic baseline exhibits a substantial drop in performance relative to the real-world upper bound. All proposed augmentations consistently narrow this gap, with model-specific differences. For DeepLabV3+, the VSD(16,0.25) + TSA and VSD(8,0.25) + TSA variants close the largest portion of the gap (28.4\% \& 63.6\%), indicating that mild, spatially localized stylization combined with target style adaptation effectively compensates for the texture sensitivity of CNNs. SegFormer shows a similar pattern, with the same variant closing 25.0\% of the gap on interior data and VSD(16,0.75) + TSA closing 45.2\% on exterior data, demonstrating that structured stylization particularly benefits mid-level Transformer encoders, too. In contrast to that, Mask2Former benefits most from the fully stylized dataset, closing 20.1\% of the gap, consistent with its strong robustness to global appearance changes. Across all architectures, the synthetic baseline yields the weakest real-world performance, and every stylization strategy improves generalization, confirming that the proposed bias-balancing augmentations reduce the synth2real mismatch rather than overfitting to synthetic artifacts.
\paragraph{Class Specific Domain Gap.}
Examining the class‑wise domain improvements reveals that the largest gains occur for classes with stable geometry and clear structural boundaries, most notably the seat components in the interior setting (see detailed results in App. Table~\ref{tab:per_class_by_model_interior}). \emph{Seat backrest} and \emph{seat pad} show improvements of up to 52 IoU points for DeepLabV3+ and 15-26 IoU points for SegFormer and Mask2Former, indicating that these categories profit from less texture bias. Mid-sized classes such as \emph{child seat} and \emph{seat belt} also show notable gains (between 5 and 25 points depending on the model). In contrast, small or visually subtle classes yield mixed results, which is a common effect in domain generalization. For \emph{clothing}, \emph{hair}, and \emph{face}, improvements are modest or slightly negative, highlighting their sensitivity to stylization. The \emph{glasses} class exhibits inconsistent behavior, dropping for DeepLabV3+ but improving for Mask2Former. This indicates that small, high-frequency details remain challenging or might be corrupted by the TSA. The catch-all rest class \emph{other object} shows no improvement. 
We observe similar effects on the exterior data (see full results in App. Table~\ref{tab:per_class_by_model_exterior}), where larger and more prominent classes, such as \emph{vegetation}, \emph{sky}, and \emph{road}, exhibit the most notable gains of up to $60$ IoU points for DeepLabV3+. The remaining classes demonstrate moderate improvement, except for \emph{human} and \emph{object}, which are the least prevalent in both the synthetic training set and the real test set and typically depict smaller objects within scenes. Overall, the proposed augmentations most reliably enhance classes characterized by recurring geometry. Fine-grained, small-scale categories improve in an architecture-dependent manner. 
\subsection{Experimental Results: Bias and Robustness}

\begin{figure*}[t]
  \newcommand{\w}{0.194\linewidth} 

  \centering
  \newcommand{\imgw}{\w}

  \begin{subfigure}[t]{\imgw}
    \centering
    \includegraphics[width=\linewidth]{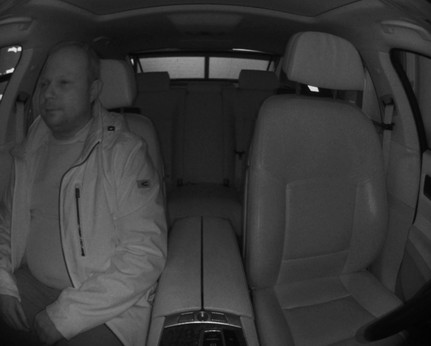}
    \caption{Original}
  \end{subfigure}\hfill
  \begin{subfigure}[t]{\imgw}
    \centering
    \includegraphics[width=\linewidth]{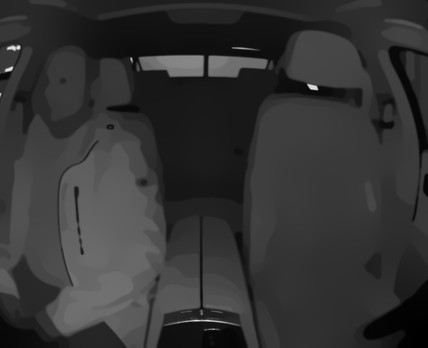}
    \caption{EED }
  \end{subfigure}\hfill
  \begin{subfigure}[t]{\imgw}
    \centering
    \includegraphics[width=\linewidth]{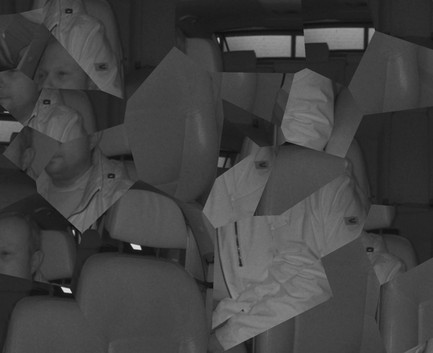}
    \caption{VPS (32)}
  \end{subfigure}\hfill
  \begin{subfigure}[t]{\imgw}
    \centering
    \includegraphics[width=\linewidth]{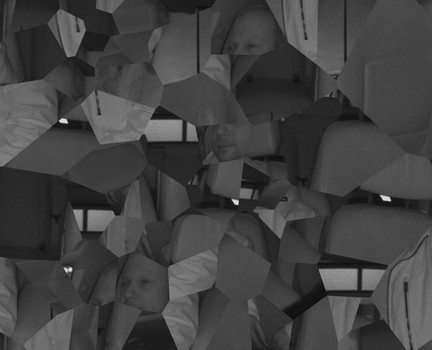}
    \caption{VPS (64)}
  \end{subfigure}\hfill
  \begin{subfigure}[t]{\imgw}
    \centering
    \includegraphics[width=\linewidth]{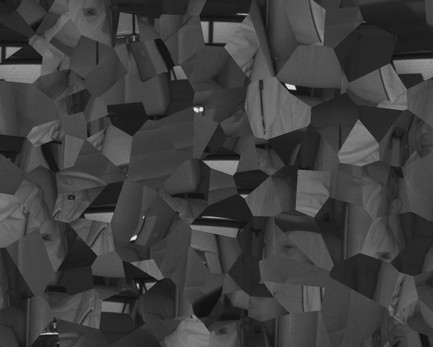}
    \caption{VPS (128)}
  \end{subfigure}
  \caption{Comparison of distortions for shape and texture bias analysis. Edge‑enhancing diffusion (EED) 
  suppresses fine texture and Voronoi patch shuffling (VPS) disrupts global spatial structure.}
  \label{fig:eed_voronoi}
\end{figure*}
\paragraph{Shape‑Texture Bias.}
We aim to quantify whether a model relies more on global shape cues or local texture patterns by evaluating it on cue decomposed data. 
Firstly, to obtain a version of our data with reduced texture content, we apply Edge Enhancing Diffusion (EED) with orientation-smoothing~\cite{heinert2024reducingtexturebiasdeep}, see 
Fig. \ref{fig:eed_voronoi}. 
Using this method, we generate a new, texture‑reduced dataset that complements our stylized variants and enables direct analysis of models under conditions with minimal texture information. 
%
Secondly, to obtain 
shape-reduced variants of our data and intentionally remove global object shape, we apply Voronoi patch shuffling (VPS), following 
\cite{heinert2025shapebiasrobustnessevaluation}. In this process, the image is partitioned into Voronoi cells, and each cell is filled with a randomly sampled patch from the original image. One example on our target domain is visualized in Fig. \ref{fig:eed_voronoi} for 32, 64 and 128 segments. This preserves local texture statistics while destroying spatial layout and object boundaries, resulting in images with dominating texture cues and minimal shape information. 
Concretely, stronger performance on \emph{shape-only} inputs (EED) indicates greater shape reliance, whereas stronger performance on \emph{texture-only} inputs (VPS) indicates greater texture reliance. These performances are combined in the Cue Decomposition Shape Bias (CDSB, \cite{heinert2025shapebiasrobustnessevaluation}) score 
\begin{equation}
S_{cd} = \frac{\tfrac{1}{s}\,Q_S}{\tfrac{1}{s}\,Q_S + \tfrac{1}{t}\,Q_T} \ \in[0,1] \enspace,
\end{equation}
where $Q_S$ and $Q_T$ are the model mIoUs 
on EED and VPS inputs, and $s$ and $t$ are their respective means over all evaluation models. 
Values closer to 1 indicate stronger shape bias, whereas closer to 0 indicate texture bias.
\begin{table}[t]
\centering
\caption{Cue-Decomposition values (EED, VPS (128), $S_{\mathrm{cd}} \in [0,1]$) on real-world validation sets (mean over 3 runs). Full results in App. Tables \ref{tab:cdsb_voronoi_128_full} and \ref{tab:scd_voronoi_128}.}
\label{tab:shape_bias_main_results_3runs}
\setlength{\tabcolsep}{2pt}
\begin{tabular}{l ccc ccc}
\toprule
& \multicolumn{3}{c}{\textbf{Interior}}
& \multicolumn{3}{c}{\textbf{Exterior}} \\
\cmidrule(lr){2-4}\cmidrule(lr){5-7}
\textbf{Method} & DeepLabV3+ & SegFormer & Mask2Former & DeepLabV3+ & SegFormer & Mask2Former \\
\midrule
Real Baseline   & $0.620$ & $0.489$ & $0.410$ & $0.371$ & $0.431$ & $0.537$ \\

Synthetic Baseline  & $0.296$ & $0.394$ & $0.399$ & $0.430$ & $0.301$ & $0.601$ \\
\midrule
TSA Only  & $0.490$ & $0.562$ & $0.549$ & $0.390$ & $0.470$ & $0.668$ \\
VSD(1,1.0)                  & $0.513$ & $0.637$ & $0.595$ & $0.437$ & $0.516$ & $0.684$ \\
VSD(4,0.75) + TSA             & $0.583$ &  $0.595$ & $0.553$ & $0.477$ & $0.577$ & $0.529$ \\
VSD(8,0.25) + TSA             & $0.508$ & $0.531$ & $0.517$ & $0.388$ & $0.487$ & $0.564$ \\
VSD(16,0.25) + TSA             & $0.497$ & $0.524$ & $0.492$ & $0.411$ & $0.486$ & $0.542$ \\
VSD(16,0.75) + TSA             & $0.558$ & $0.580$ & $0.526$ & $0.452$ & $0.517$ & $0.557$ \\
\bottomrule
\end{tabular}
\end{table}
The cue‑decomposition results, presented in Table~\ref{tab:shape_bias_main_results_3runs} quantify how strongly each model relies on shape cues or texture cues in the resulting shape‑bias score $S_{cd}$. 
Across all architectures, the real baseline exhibits a noticeably higher shape‑bias than the synthetic baseline on the interior data, indicating that synthetic‑only training produces models rely more heavily on texture and less on shape. This gap is particularly visible for DeepLabV3+, where synthetic training shows a low $S_{cd}$=0.296 compared to 0.62 for the real baseline, but the effect is also present for SegFormer and Mask2Former. In the exterior experiments, this pattern emerges for the SegFormer architecture, whereas the opposite behavior can be observed in DeepLabV3+ and Mask2Former models. The full results in Table~\ref{tab:scd_voronoi_128} reveal that the DeepLabV3+ and Mask2Former synthetic GTA5 RGB baseline models achieve the lowest measured mIoU on VPS NIR texture images, indicating that the learned synthetic RGB textures do not transfer to the NIR spectrum images.
The augmentation strategies shift these biases in different directions depending on the architecture.
TSA consistently increases mIoU on EED shape samples, which leads to a higher measured shape bias in most evaluations. The exception of the DeepLabV3+ model on exterior data arises from a simultaneously increased mIoU on Voronoi-shuffled images, which likely arises from a better transferability of learned texture features to real NIR images.
Training on fully stylized images VSD(1,1.0) increases shape‑bias even further for SegFormer and Mask2Former in both evaluation settings, producing some of the highest $S_{cd}$ values in their columns, but has a milder effect for DeepLabV3+, where texture responses remain relatively high.
The Voronoi + TSA variants generally balance the cue responses, i.e., they increase shape sensitivity relative to the synthetic baseline while maintaining moderate texture responses, resulting in intermediate $S_{cd}$ values. This indicates that combined stylization and NIR adaptation provides a more controlled balance of texture and shape. A higher proportion of stylized segments consistently results in an increased shape bias measurement, with configuration VSD(4,0.75) + TSA leading to the highest observed shape bias for the DeepLabV3+ model in both evaluations and for the SegFormer model in exterior evaluations.
The results indicate that synthetic‑only training leads to texture‑dominated models, while stylization and generative augmentation systematically shift the models toward a more shape‑balance, with the magnitude and direction of this shift depending on model architecture and augmentation type.

\begin{figure}[t]
  \newcommand{\w}{0.194\linewidth} 
    \centering
    \begin{subfigure}[t]{\w}
        \centering
        \includegraphics[width=\textwidth]{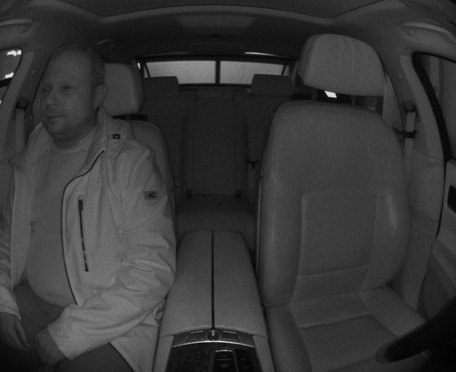}
        \caption{Original}
    \end{subfigure}
    \hfill
    \begin{subfigure}[t]{\w}
        \centering
        \includegraphics[width=\textwidth]{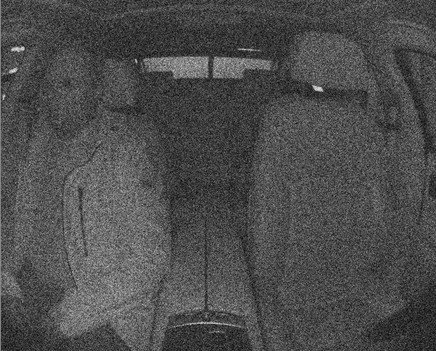}
        \caption{Uniform}
    \end{subfigure}
    \hfill
    \begin{subfigure}[t]{\w}
        \centering
        \includegraphics[width=\textwidth]{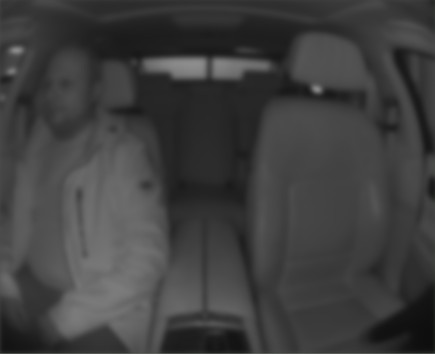}
        \caption{Low-pass}
    \end{subfigure}
    \hfill
    \begin{subfigure}[t]{\w}
        \centering
        \includegraphics[width=\textwidth]{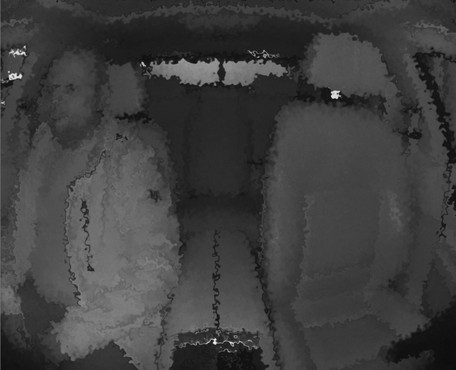}
        \caption{Elastic}
    \end{subfigure}
    \hfill
    \begin{subfigure}[t]{\w}
        \centering
        \includegraphics[width=\textwidth]{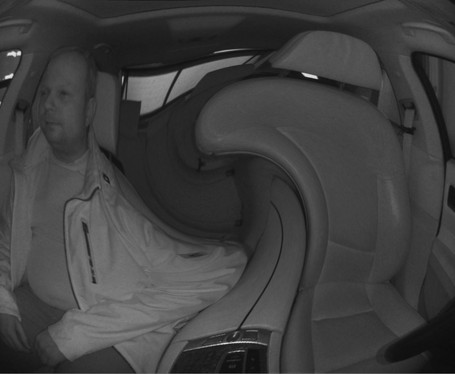}
        \caption{Swirl}
    \end{subfigure}
    \caption{Distortions: Uniform ($\sigma=0.6$), Low-pass ($\sigma=10$), Elastic ($\alpha=150$) and Swirl ($\sigma=2$). }
    \label{fig:single_noise_comparison}
\end{figure}

\paragraph{Distortion Results.}
To systematically investigate model robustness, we apply four types of distortions that mimic failure modes, separating texture-dependent behavior from shape sensitivity (see Fig.~\ref{fig:single_noise_comparison}). For each type, severity is gradually increased to measure the incremental performance drop. 
We apply two forms of texture‑targeting distortions. Firstly, low‑pass filtering removes high‑frequency texture by using a Gaussian filter with standard deviation $\sigma \in \{1, 5, 10, 15\}$. This operation keeps the coarse structure of the image intact. 
Secondly, uniform noise introduces pixel‑level perturbations with noise magnitude $w \in \{0.03, \allowbreak 0.05, \allowbreak 0.1, \allowbreak 0.2, \allowbreak 0.35, \allowbreak 0.6, \allowbreak 0.9\}$.
 The added noise disrupts fine texture but leaves the overall shape unchanged.
We also apply two forms of shape‑targeting distortions. Firstly, elastic deformation creates spatially coherent geometric warping based on displacement magnitude $\alpha \in \{10, 20, 50, 100, 150\}$ with fixed standard deviation $\sigma=4$. This distortion changes global shape while preserving texture. Secondly, swirl transformation introduces a rotation field that gradually weakens with distance from the center. It uses strength $s \in \{0.05, 0.1, 0.5, 1.0, 2.0\}$ and with fixed radius of 225 pixels. This transformation alters global shape yet keeps local texture.
These distortions allow us to separate texture robustness from shape robustness under controlled and interpretable perturbations.
We summarize corruption robustness using the area-normalized mean performance under corruption ($mPC_{AUC}$). For each corruption type $c$, we compute the 
severity-normalized area under the mIoU-vs-severity curve via the trapezoidal rule
and average across corruption types, 
\[
\mathrm{AUC}_{m}(c)
    = \frac{1}{s_{\max}-s_{\min}}
      \int_{s_{\min}}^{s_{\max}} \mathrm{mIoU}_{m}(c, s)\,\mathrm{d}s \enspace, \enspace \enspace
\mathrm{mPC_{AUC}}
    = \frac{1}{|\mathcal{C}|}
      \sum_{c\in\mathcal{C}} \mathrm{AUC}_{m}(c) \enspace.
\]
For interpretability, we \emph{normalize} $mPC_{AUC}$ by dividing each 
model's score by that of the real Baseline (100 = parity), inspired by mCE by Hendrycks~\cite{hendrycks2019benchmarking}.
\begin{table}[t]
\centering
\caption{Area-normalized mIoU under corruptions. Normalized with real baseline = 100. Bold marks the highest score in each shown model group (see full App. Tables  \ref{tab:robustness_norm_combined}, \ref{tab:robustness_summary_exterior_all}, \ref{tab:robustness_norm_all_models} and \ref{tab:per_distortion_ranus_nir}). }
\label{tab:robustness_merged_logical}
\setlength{\tabcolsep}{2pt}

\begin{tabular}{l ccccc | ccccc}
\toprule
& \multicolumn{5}{c|}{\textbf{Interior (Normalized mIoU)}} & \multicolumn{5}{c}{\textbf{Exterior (Normalized mIoU)}} \\
\textbf{Method} & Elastic & Low-pass & Swirl & Uniform & Mean & Elastic & Low-pass & Swirl & Uniform & Mean \\
\midrule
\multicolumn{11}{l}{\textbf{DeepLabV3+}} \\
Synth. Baseline & 19.1 & 23.0 & 38.1 & 21.5 & 25.4 & 23.5 & 49.8 & 27.6 & 22.9 & 31.0 \\
TSA Only        & 34.0 & 31.9 & 40.9 & 18.9 & 31.4 & 38.2 & 72.0 & 60.9 & 39.1 & 52.5 \\
VSD(1,1.0)      & 41.7 & 18.1 & 31.7 & 24.7 & 29.0 & 45.5 & 54.5 & 39.4 & 35.4 & 43.7 \\
VSD(8,0.25) + TSA & 60.2 & 41.2 & 46.1 & 64.9 & 53.1 & \textbf{51.1} & \textbf{78.3} & \textbf{70.6} & \textbf{72.4} & \textbf{68.1} \\
VSD(16,0.75) + TSA & \textbf{60.8} & \textbf{57.3} & \textbf{46.5} & \textbf{70.8} & \textbf{58.8} & 42.6 & 45.3 & 62.0 & 58.7 & 52.1 \\
\midrule
\multicolumn{11}{l}{\textbf{SegFormer}} \\
Synth. Baseline & 37.6 & 34.0 & 53.6 & 53.0 & 44.5 & 41.7 & 27.7 & 39.9 & 30.8 & 35.0 \\
VSD(1,1.0)      & \textbf{74.5} & \textbf{66.7} & 58.7 & \textbf{76.8} & \textbf{69.2} & 44.7 & 55.2 & 45.4 & 102.2 & 61.9 \\
VSD(4,0.75) + TSA & 72.3 & 54.9 & 55.2 & 73.8 & 64.1 & \textbf{60.4} & \textbf{61.8} & 56.2 & \textbf{113.3} & \textbf{72.9} \\
VSD(16,0.25) + TSA & 72.0 & 51.9 & \textbf{59.8} & 71.4 & 63.8 & 58.3 & 55.3 & \textbf{59.5} & 89.9 & 65.8 \\
\midrule
\multicolumn{11}{l}{\textbf{Mask2Former}} \\
Synth. Baseline & 47.9 & 53.4 & 60.0 & 68.3 & 57.4 & 69.4 & 58.9 & 68.2 & 75.7 & 68.0 \\
TSA Only        & 61.9 & 53.2 & 59.3 & 52.3 & 56.7 & 59.0 & 57.7 & 68.3 & 72.0 & 64.2 \\
VSD(1,1.0)      & \textbf{83.8} & \textbf{73.9} & \textbf{67.7} & \textbf{78.7} & \textbf{76.0} & \textbf{93.8} & \textbf{78.2} & \textbf{77.6} & \textbf{86.6} & \textbf{84.0} \\
\bottomrule
\end{tabular}
\end{table}
Table~\ref{tab:robustness_merged_logical} depicts selected results for the performance under corruptions (see distortion versus robustness curves in App.~\ref{sec:detailed_robust}). There are fundamental differences between the models. DeepLabV3+ trained on the VSD(16,0.75) + TSA has the best performance over all corruptions on interior data. On exterior data the DeepLabV3+ model shows the highest robustness when trained with VSD(8,0.25) + TSA.
The SegFormer evaluations yield the highest robustness for the fully stylized VSD(1,1.0) on interior data, whereas VSD(4,0.75) shows the highest exterior performance. 
For Mask2Former, the VSD(1,1.0) variant achieves by far the best mean performance under corruptions for both interior and exterior. In most cases, higher style diversification increased robustness in both Transformer models, while the synthetic baseline remained weak under corruption and improved with every stylization method.
\subsection{Discussion}
\paragraph{Domain Gap Reduction.}All proposed augmentation methods improve upon the synthetic baseline. This confirms that the domain gap between synthetic and real NIR imagery is substantial and that targeted style adaptation meaningfully reduces it. The best-performing configurations close between 20.1\% and 28.4\% of the domain gap on interior data, depending on the architecture. On exterior images, the gap is reduced between 20.4\% and 63.6\%, with the additional RGB$\rightarrow$NIR domain gap heavily influencing the lightweight models. The CNN-based DeepLabV3+ benefits most, 
while the 
Mask2Former benefits least, 
consistent with its stronger inductive priors from 
pretraining.

The benefits of augmentation are unevenly distributed across semantic classes. Categories with stable, recurring geometry such as seat parts show the largest gains, in some cases exceeding 50 IoU points for DeepLabV3+ (see App. Tables \ref{tab:per_class_by_model_interior},\ref{tab:per_class_by_model_exterior}). This is consistent with the hypothesis that texture-biased models struggle most with structurally predictable classes whose synthetic and real appearances differ primarily in texture. Small or visually irregular classes 
show modest or inconsistent improvements. Fine-grained appearance details remain challenging under domain shift. Notably, the glasses class decreases for DeepLabV3+ while improving for Mask2Former, pointing to architecture-specific sensitivity to the TSA for high-frequency, small-scale features.
\paragraph{Effect of TSA.}
TSA-only training improves mIoU over the synthetic baseline across all three architectures. This demonstrates that generative NIR style transfer is a practically effective adaptation strategy even without additional stylization. The cue-decomposition results support this finding. The training with TSA consistently shifts models toward higher shape-bias scores compared to the synthetic baseline. This suggests that exposure to more realistic NIR appearance introduces a kind of noise to the data, that encourages models to rely on structural rather than textural cues. We also investigated the effect of VSD without TSA in App. Tables \ref{tab:ablation_all_main_results_3runs} and \ref{tab:ablation_all_main_results_3runs_ext}, which show the same pattern but with lower overall performance, demonstrating the benefit of combining TSA with VSD.
\paragraph{Effect of VSD.}
We conducted an ablation on Voronoi stylization without TSA (see App. Tables \ref{tab:ablation_all_main_results_3runs},\ref{tab:ablation_all_main_results_3runs_ext})) which already improves over the pure synthetic baseline for DeepLabV3+ , while gains for SegFormer and Mask2Former remain marginal. Adding RGB instead of grayscale stylization yields no consistent further benefit, suggesting that color diversity matters little without TSA. This confirms that TSA is the primary driver of improvement, with style diversification acting as a complementary regularizer.
Combining VSD with TSA yields the best segmentation performance for DeepLabV3+ and SegFormer. Mask2Former, however, benefits more from the fully stylized variant. Additionally, training on data with VSD augmentation only, yields moderate improvements in the target domain compared to the synthetic baseline. This architecture-dependent pattern is informative. Convolutional and more lightweight (regarding whole pre-training and model size) transformer architectures appear to benefit from the controlled, localized texture variation that Voronoi stylization introduces. Mask2Former's shifted window attention mechanism and larger capacity allow it to leverage global stylization without loss of structural fidelity. Higher stylization probability tends to increase shape-bias scores but does not always translate to higher segmentation performance. This points to a trade-off between texture reduction and the preservation of domain-relevant appearance features.
\paragraph{Robustness to Distortions.}
The distortion experiments show that the synthetic baseline is consistently weak under corruption, and almost every augmentation strategy improves upon it. However, the optimal augmentation strategy for distortion robustness does not always coincide with the best strategy for best mIoU. For DeepLabV3+, mild Voronoi stylization achieves the best mean robustness. For SegFormer and Mask2Former, the fully stylized variant result in the highest robustness. This reveals that the relationship between appearance diversity during training and robustness to corruptions is architecture-dependent. A single augmentation strategy may not be able to simultaneously optimize both performance and corruption robustness.
\paragraph{Bias Balancing and Generalization.}
The cue-decomposition and segmentation results support the hypothesis, that synthetic-only training produces texture-dominated models that generalize poorly to real NIR data, while controlled style augmentation systematically shifts the bias toward shape, improving cross-domain performance and robustness. The magnitude of this shift depends on augmentation intensity, model architecture, and data, confirming that bias balancing is not a one-size-fits-all process and relies on network's specific inductive properties.
\begin{figure}[t]
    \centering
    \setlength{\tabcolsep}{1pt}%
    \resizebox{\textwidth}{!}{%
    \begin{tabular}{@{} cc @{\hspace{4pt}} cc @{\hspace{4pt}} cc @{}}
\multicolumn{2}{c}{\scriptsize \textbf{Deeplabv3+}} &
\multicolumn{2}{c}{\scriptsize \textbf{SegFormer}} &
\multicolumn{2}{c}{\scriptsize \textbf{Mask2Former}} \\
      \scriptsize Baseline & \scriptsize Best &
      \scriptsize Baseline & \scriptsize Best &
      \scriptsize Baseline & \scriptsize Best \\
      \midrule
      \multicolumn{2}{c}{\scriptsize \textcolor{green!60!black}{$\uparrow$ \textbf{+44.1} mIoU}} &
      \multicolumn{2}{c}{\scriptsize \textcolor{green!60!black}{$\uparrow$ \textbf{+17.5} mIoU}} &
      \multicolumn{2}{c}{\scriptsize \textcolor{green!60!black}{$\uparrow$ \textbf{+15.3} mIoU}} \\
        \includegraphics[width=\thumbwidth]{figures/new_header/worst_deeplab_img_val_00174.jpg}%
        &
        \includegraphics[width=\thumbwidth]{figures/new_header/best_deeplab_img_val_00174.jpg}
        &
        \includegraphics[width=\thumbwidth]{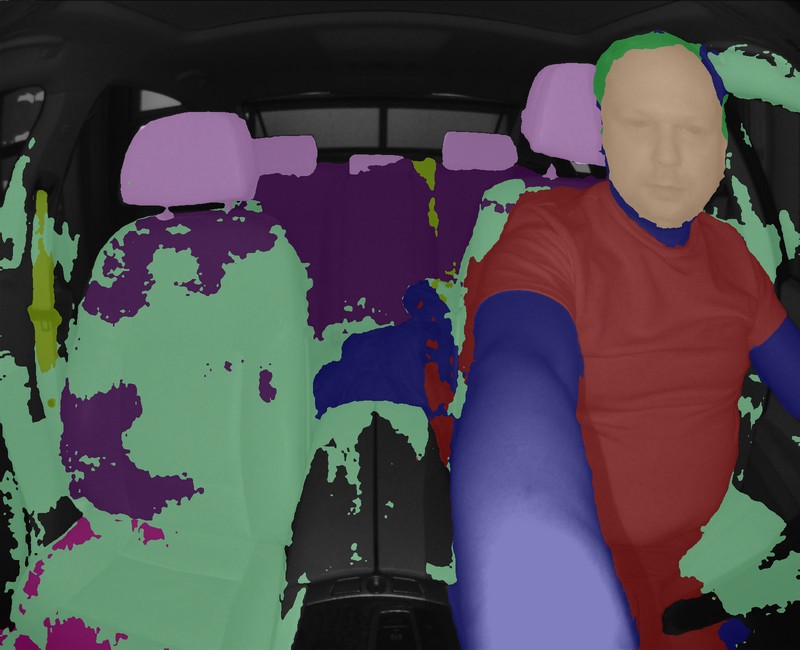}%
        &
        \includegraphics[width=\thumbwidth]{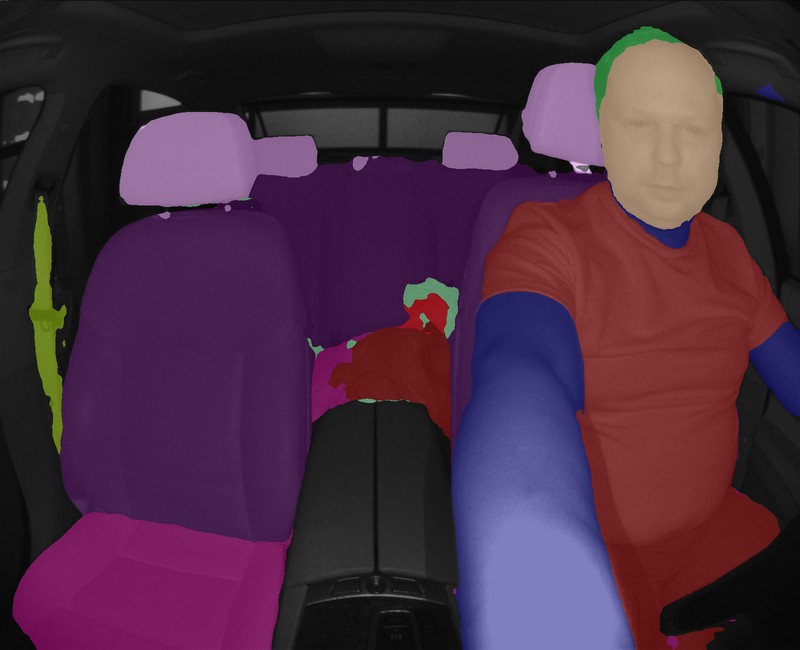}
        &
        \includegraphics[width=\thumbwidth]{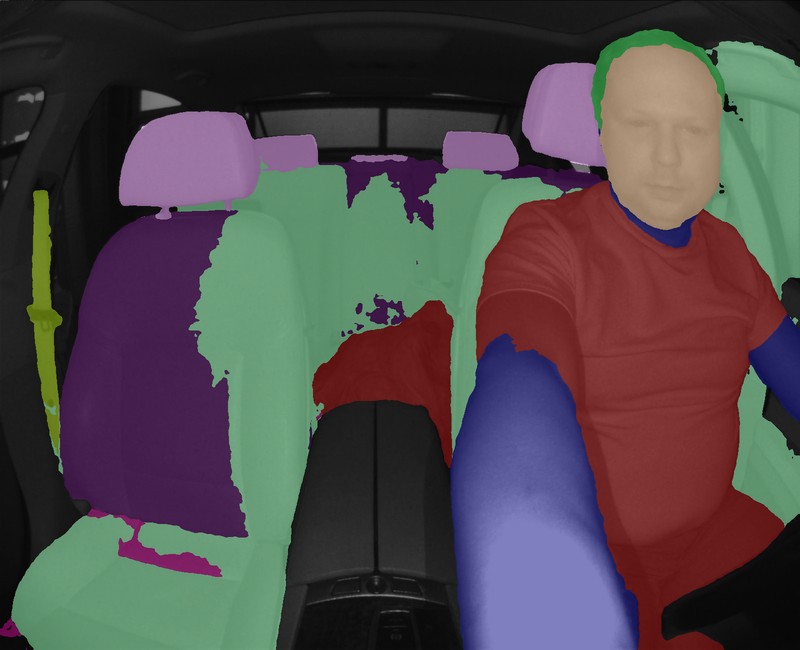}%
         &
        \includegraphics[width=\thumbwidth]{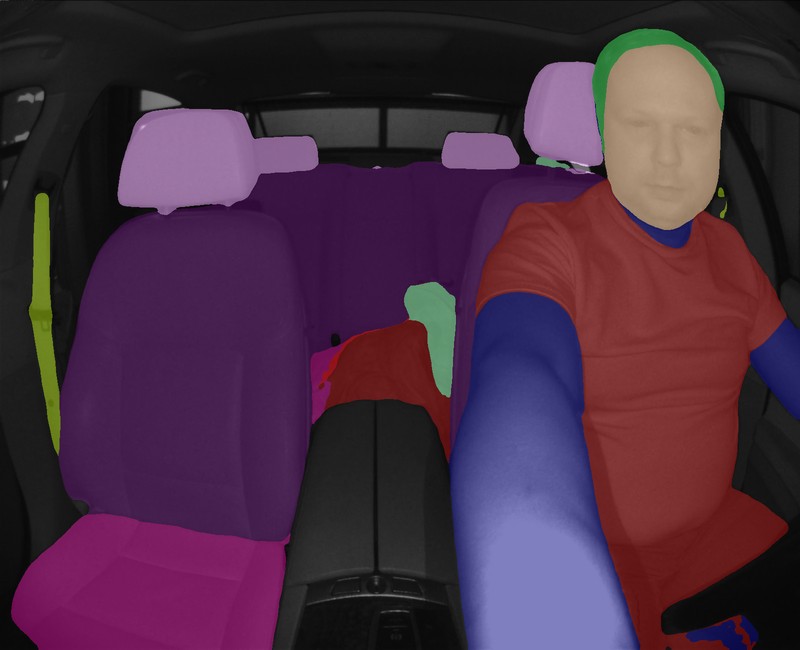}\\
      
      \includegraphics[width=\thumbwidth, trim=0.0cm 0.5cm 0.0cm 5cm, clip]{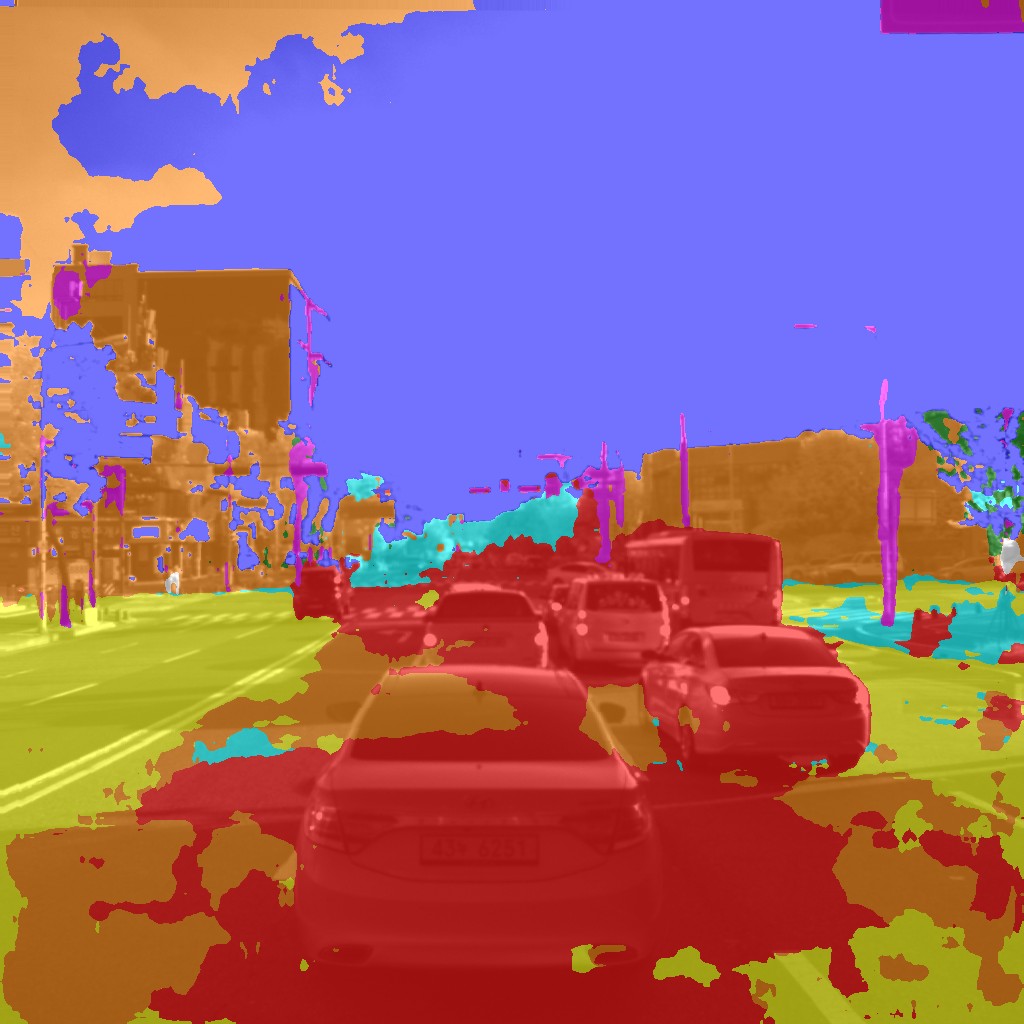}%
      &
      \includegraphics[width=\thumbwidth, trim=0.0cm 0.5cm 0.0cm 5cm, clip]{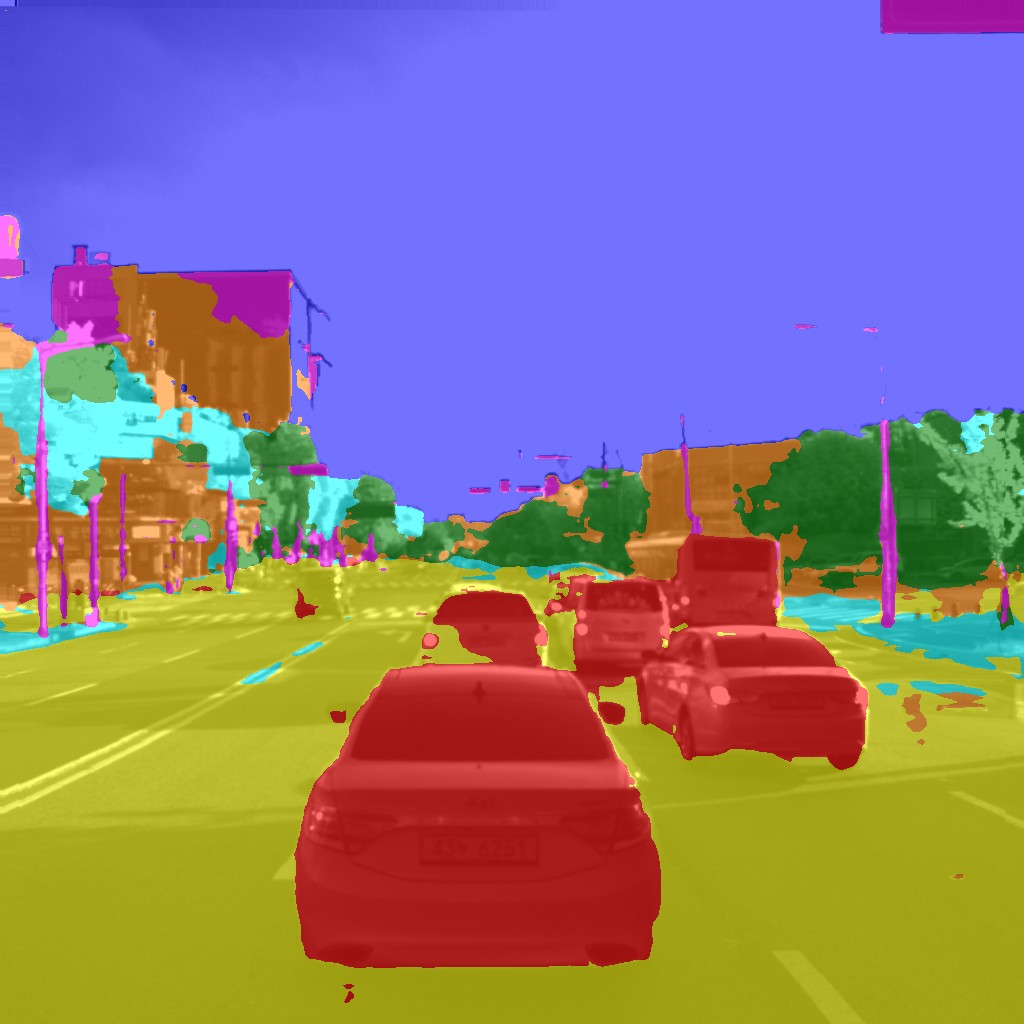}%
      &
      \includegraphics[width=\thumbwidth, trim=0.0cm 0.5cm 0.0cm 5cm, clip]{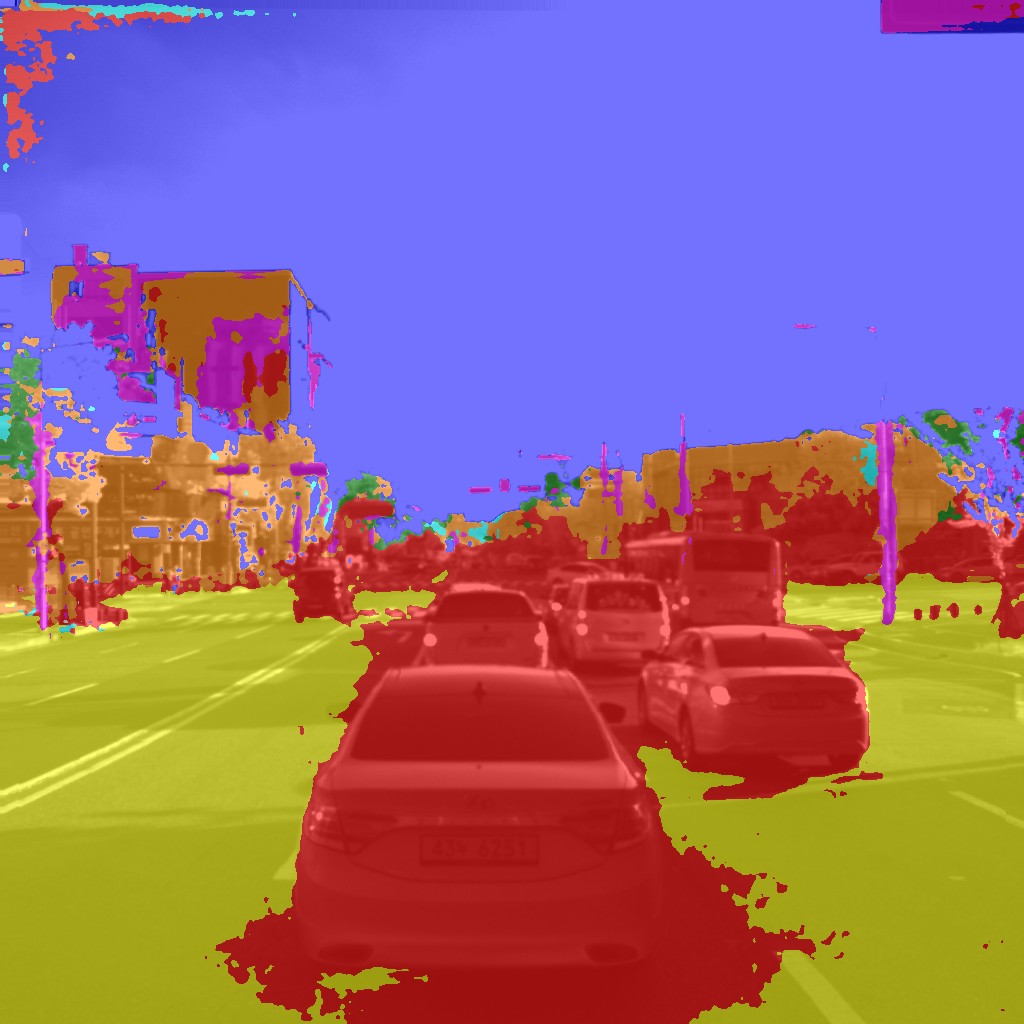}%
      &
      \includegraphics[width=\thumbwidth, trim=0.0cm 0.5cm 0.0cm 5cm, clip]{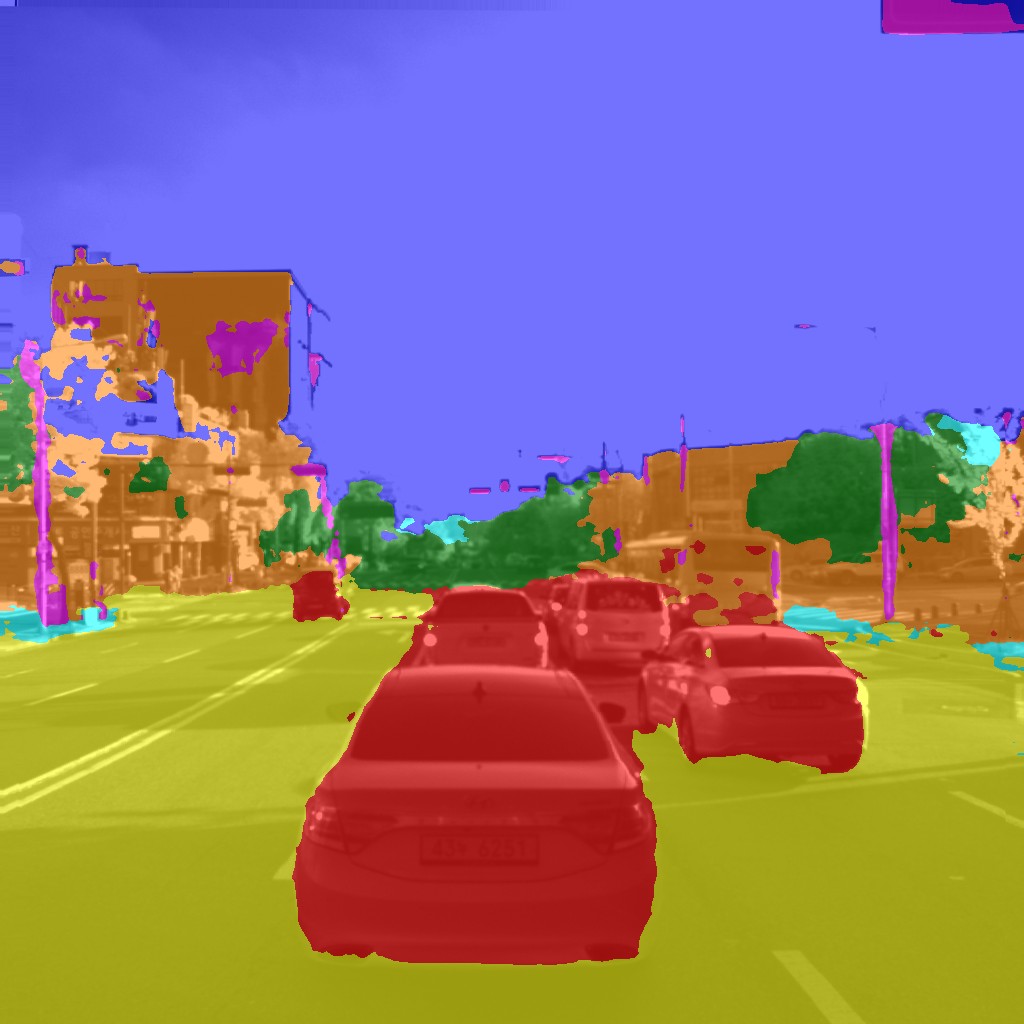}%
      &
      \includegraphics[width=\thumbwidth, trim=0.0cm 0.5cm 0.0cm 5cm, clip]{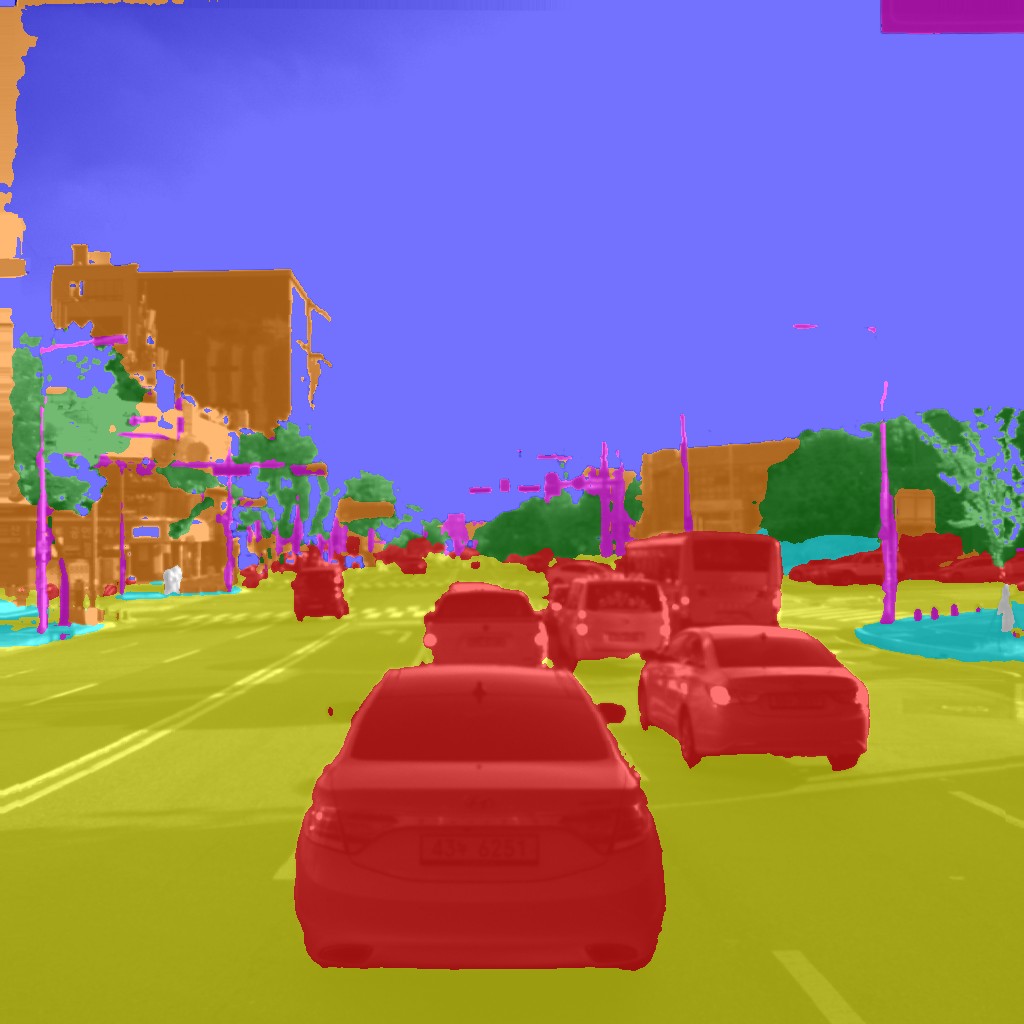}%
      &
      \includegraphics[width=\thumbwidth, trim=0.0cm 0.5cm 0.0cm 5cm, clip]{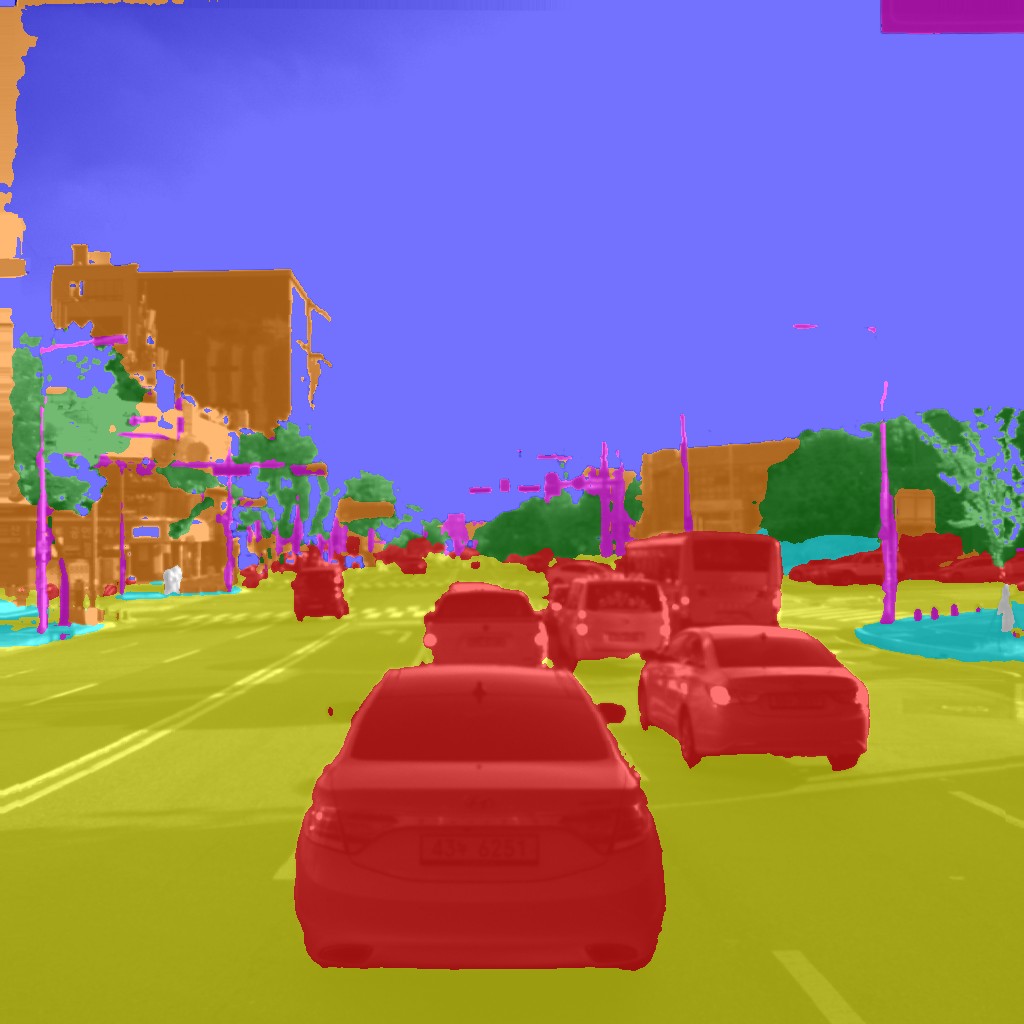}\\[-1ex] 
      
      \multicolumn{2}{c}{\scriptsize \textcolor{green!60!black}{$\uparrow$ \textbf{+27.8} mIoU}} &
      \multicolumn{2}{c}{\scriptsize \textcolor{green!60!black}{$\uparrow$ \textbf{+18.7} mIoU}} &
      \multicolumn{2}{c}{\scriptsize \textcolor{green!60!black}{$\uparrow$ \textbf{+4.8} mIoU}} \\
    \end{tabular}
    }
    \caption{Qualitative inference with baseline versus best configuration. mIoU improvements in green.}

    \label{fig:models-per-column}
  \end{figure}%
\paragraph{Domain Impact.}
For automotive NIR perception, our method closes 63.6\% on exterior (and 28.4\% on interior) of the synth2real gap without additional real annotations, reducing reliance on costly pixel‑level labels for driver- and occupant-monitoring 
and exterior scene parsing (see Fig. \ref{fig:models-per-column} for real-world effect and more in App. Table \ref{fig:models-per-column_qual}). By rebalancing texture–shape inductive bias, it yields models that generalize better across domains and improves robustness to common corruptions. Gains are strongest for geometrically stable, safety‑relevant classes (e.g., seat components), while exterior results bolster scene‑layout understanding for road/vegetation/sky. The pipeline is model‑agnostic and relies only on unlabeled target images for generative 
target style adaptation, easing integration into existing training stacks.

\section{Conclusion}
We presented a novel method that enables a systematic study of synth2real domain adaptation for semantic segmentation in automotive near-infrared imagery, covering both in-cabin monitoring and exterior road scene perception. Nevertheless, the approach is model agnostic and can be applied to other domains and settings. Our generative augmentation framework combines a LoRA-based target style adaptation with structure-preserving multi-signal conditioning. It is further paired with a Voronoi-partitioned style diversification strategy to explicitly influence texture and shape 
biases during training.
Experiments across three 
segmentation architectures demonstrate that the proposed approach consistently reduces the synth2real domain gap by up to 28.4\,\% on interior and 63.6\% on exterior data.
Cue-decomposition analysis confirms that synthetic-only training leads to texture-dominated models. The proposed augmentations 
shift the bias toward shape-based representations, improving both segmentation accuracy and robustness to image corruptions. Class-level analysis shows that gains are largest for geometrically stable categories, while fine-grained small classes remain challenging across architectures.
These findings suggest that inductive bias balancing through generative data augmentation is effective for bridging the synth2real gap in semantic segmentation. Our method enables us to identify augmentation settings to balance biases. 
Across our data a mild stylization for DeepLabV3+ and SegFormer is the best performing augmentation configuration and for Mask2Former a full stylization with only one cell yields the best performance.
Future work includes embedding the proposed augmentation into self-training or knowledge-distillation pipelines to generate reliable pseudo-labels for domain-sensitive lightweight models.

\begin{credits}
\subsubsection{\discintname}The authors have no competing interests to declare.

\subsubsection{Disclosure of LLM Usage.} Gemini is used for image captioning of RANUS data (App. \ref{subsec:loratrain}), Gemini and Claude were used for language polishing. We manually checked all LLM-polished text.

\end{credits}
%
%
%
\bibliographystyle{splncs04}
\bibliography{arxiv_long_bib}

\begin{thebibliography}{10}
\providecommand{\url}[1]{\texttt{#1}}
\providecommand{\urlprefix}{URL }
\providecommand{\doi}[1]{https://doi.org/#1}

\bibitem{bae2021estimating}
Bae, G., Budvytis, I., Cipolla, R.: Estimating and exploiting the aleatoric
  uncertainty in surface normal estimation. In: ICCV. pp. 13137--13146 (2021)

\bibitem{burgert2026imagenettrained}
Burgert, T., Stoll, O., Rota, P., Demir, B.: Imagenet-trained cnns are not
  biased towards texture: Revisiting feature reliance through controlled
  suppression (2025)

\bibitem{chen2022light}
Chen, J., Liu, Z., Jin, D., Wang, Y., Yang, F., Bai, X.: Light transport
  induced domain adaptation for semantic segmentation in thermal infrared urban
  scenes. IEEE Transactions on Intelligent Transportation Systems
  \textbf{23}(12),  23194--23211 (2022)

\bibitem{chen2018encoder}
Chen, L.C., Zhu, Y., Papandreou, G., Schroff, F., Adam, H.: Encoder-decoder
  with atrous separable convolution for semantic image segmentation. In:
  Proceedings of the European conference on computer vision (ECCV). pp.
  801--818 (2018)

\bibitem{cheng2022masked}
Cheng, B., Misra, I., Schwing, A.G., Kirillov, A., Girdhar, R.:
  Masked-attention mask transformer for universal image segmentation. In:
  Proceedings of the IEEE/CVF conference on computer vision and pattern
  recognition. pp. 1290--1299 (2022)

\bibitem{Cheng2019Non-Local}
Cheng, Z., Zheng, Y., You, S., Sato, I.: Non-local intrinsic decomposition with
  near-infrared priors. In: Proceedings of the IEEE/CVF international
  conference on computer vision. pp. 2521--2530 (2019)

\bibitem{cho_one-shot_2024}
Cho, H., Lee, J., Chang, S., Jeong, Y.: One-shot structure-aware stylized image
  synthesis. In: Proceedings of the IEEE/CVF Conference on Computer Vision and
  Pattern Recognition. pp. 8302--8311 (2024)

\bibitem{choe2018ranus}
Choe, G., Kim, S.H., Im, S., Lee, J.Y., Narasimhan, S.G., Kweon, I.S.: Ranus:
  Rgb and nir urban scene dataset for deep scene parsing. IEEE Robotics and
  Automation Letters  \textbf{3}(3),  1808--1815 (2018)

\bibitem{cohen2025TextureSAMTexture}
Cohen, I., Meivar, B., Tu, P., Avidan, S., Oren, G.: Texturesam: Towards a
  texture aware foundation model for segmentation. arXiv preprint
  arXiv:2505.16540  (2025)

\bibitem{comanici2025gemini}
Comanici, G., Bieber, E., Schaekermann, M., Pasupat, I., Sachdeva, N., Dhillon,
  I., Blistein, M., Ram, O., Zhang, D., Rosen, E., et~al.: Gemini 2.5: Pushing
  the frontier with advanced reasoning, multimodality, long context, and next
  generation agentic capabilities. arXiv:2507.06261  (2025)

\bibitem{Cordts2016Cityscapes}
Cordts, M., Omran, M., Ramos, S., Rehfeld, T., Enzweiler, M., Benenson, R.,
  Franke, U., Roth, S., Schiele, B.: The cityscapes dataset for semantic urban
  scene understanding. In: Proceedings of the IEEE conference on computer
  vision and pattern recognition. pp. 3213--3223 (2016)

\bibitem{cruz2020sviro}
Cruz, S.D.D., Wasenmuller, O., Beise, H.P., Stifter, T., Stricker, D.: Sviro:
  Synthetic vehicle interior rear seat occupancy dataset and benchmark. In:
  WACV. pp. 973--982 (2020)

\bibitem{painterbynumbers2016}
small~yellow duck, Kan, W.: Painter by numbers (2016), kaggle Dataset

\bibitem{EURegulation2019}
{European Parliament}, {Council of the European Union}: Regulation ({EU})
  2019/2144 on type-approval requirements for motor vehicles and their
  trailers, and systems, components and separate technical units intended for
  such vehicles, as regards their general safety and the protection of vehicle
  occupants and vulnerable road users. Regulation 2019/2144, European
  Parliament and Council of the European Union (12 2019)

\bibitem{geirhos2020Shortcutlearninga}
Geirhos, R., Jacobsen, J.H., Michaelis, C., Zemel, R., Brendel, W., Bethge, M.,
  Wichmann, F.A.: Shortcut learning in deep neural networks. Nature Machine
  Intelligence  \textbf{2}(11),  665--673 (2020)

\bibitem{geirhos2018imagenet}
Geirhos, R., Rubisch, P., Michaelis, C., Bethge, M., Wichmann, F.A., Brendel,
  W.: Imagenet-trained cnns are biased towards texture; increasing shape bias
  improves accuracy and robustness. In: ICLR (2018)

\bibitem{hamscher2025TransferringStyles}
Hamscher, B., Heinert, E., M{\"u}tze, A., Maag, K., Rottmann, M.: Transferring
  styles for reduced texture bias and improved robustness in semantic
  segmentation networks (2025)

\bibitem{heinert2025shapebiasrobustnessevaluation}
Heinert, E., Gottwald, T., M{\"u}tze, A., Rottmann, M.: Shape bias and
  robustness evaluation via cue decomposition for image classification and
  segmentation. arXiv preprint arXiv:2503.12453  (2025)

\bibitem{heinert2024reducingtexturebiasdeep}
Heinert, E., Rottmann, M., Maag, K., Kahl, K.: Reducing texture bias of deep
  neural networks via edge enhancing diffusion. In: ECAI 2024. pp. 609--617.
  IOS Press (2024)

\bibitem{hendrycks2019benchmarking}
Hendrycks, D., Dietterich, T.: Benchmarking neural network robustness to common
  corruptions and perturbations (2019)

\bibitem{hermann2020origins}
Hermann, K., Chen, T., Kornblith, S.: The origins and prevalence of texture
  bias in convolutional neural networks. NeurIPS  \textbf{33},  19000--19015
  (2020)

\bibitem{hoyer2023mic}
Hoyer, L., Dai, D., Wang, H., Van~Gool, L.: Mic: Masked image consistency for
  context-enhanced domain adaptation. In: Proceedings of the IEEE/CVF
  conference on computer vision and pattern recognition. pp. 11721--11732
  (2023)

\bibitem{hu2022lora}
Hu, E.J., Shen, Y., Wallis, P., Allen-Zhu, Z., Li, Y., Wang, S., Wang, L.,
  Chen, W., et~al.: Lora: Low-rank adaptation of large language models. Iclr
  \textbf{1}(2), ~3 (2022)

\bibitem{huang_arbitrary_2017}
Huang, X., Belongie, S.: Arbitrary style transfer in real-time with adaptive
  instance normalization. In: Proceedings of the IEEE international conference
  on computer vision. pp. 1501--1510 (2017)

\bibitem{ISOPAS8800}
{International Organization for Standardization}: Road vehicles --- safety and
  artificial intelligence. Tech. Rep. ISO/PAS 8800:2024, International
  Organization for Standardization, Geneva, Switzerland (Dec 2024),
  \url{https://www.iso.org/standard/83303.html}

\bibitem{jia2024dginstyle}
Jia, Y., Hoyer, L., Huang, S., Wang, T., Van~Gool, L., Schindler, K., Obukhov,
  A.: Dginstyle: Domain-generalizable semantic segmentation with image
  diffusion models and stylized semantic control. In: European Conference on
  Computer Vision. pp. 91--109. Springer (2024)

\bibitem{katrolia2021ticam}
Katrolia, J.S., El{-}Sherif, A., Feld, H., Mirbach, B., Rambach, J.R.,
  Stricker, D.: Ticam: A time-of-flight in-car cabin monitoring dataset. In:
  BMVC. p.~277. {BMVA} Press (2021)

\bibitem{kwak2019ImpactTexture}
Kwak, G.H., Park, N.W.: Impact of texture information on crop classification
  with machine learning and uav images. Applied Sciences  \textbf{9}(4), ~643
  (2019)

\bibitem{kwon2023diffusionbasedimagetranslationusing}
Kwon, G., Ye, J.C.: Diffusion-based image translation using disentangled style
  and content representation (2023)

\bibitem{LaserFocusWorld2013}
{Laser Focus World Editorial Staff}: High-power vcsels rule ir illumination.
  Laser Focus World  (2013),
  \url{https://www.laserfocusworld.com/lasers-sources/article/16556827/vcsel-illumination-high-power-vcsels-rule-ir-illumination}

\bibitem{Lin2024LearningContrastEnhanced}
Lin, F., Bao, K., Li, Y., Zeng, D., Ge, S.: Learning contrast-enhanced
  shape-biased representations for infrared small target detection. IEEE
  Transactions on Image Processing  \textbf{33},  3047--3058 (2024)

\bibitem{Lin2024LearningShapeBiased}
Lin, F., Ge, S., Bao, K., Yan, C., Zeng, D.: Learning shape-biased
  representations for infrared small target detection. IEEE Transactions on
  Multimedia  \textbf{26},  4681--4692 (2023)

\bibitem{LIU2023117}
Liu, Q., Dou, Q., Chen, C., Heng, P.A.: Domain generalization of deep networks
  for medical image segmentation via meta learning. In: Meta Learning with
  Medical Imaging and Health Informatics Applications, pp. 117--139. Elsevier
  (2023)

\bibitem{mortimer2022tas}
Mortimer, P., Wuensche, H.J.: Tas-nir: A vis+ nir dataset for fine-grained
  semantic segmentation in unstructured outdoor environments. arXiv preprint
  arXiv:2212.09368  (2022)

\bibitem{mutze2025InfluenceShape}
Mütze, A., Grabowsky, N., Heinert, E., Rottmann, M., Gottschalk, H.: On the
  Influence of Shape, Texture and Color for Learning Semantic Segmentation, pp.
  1727--1735. IOS Press (10 2025). \doi{10.3233/FAIA251001}

\bibitem{naseer2021intriguing}
Naseer, M.M., Ranasinghe, K., Khan, S.H., Hayat, M., Shahbaz~Khan, F., Yang,
  M.H.: Intriguing properties of vision transformers. Advances in Neural
  Information Processing Systems  \textbf{34},  23296--23308 (2021)

\bibitem{PalaoEtAl2023}
Palao, A., Fredriksson, R., Lenn{\'e}, M.: Euro ncap’s current and future
  in-cabin monitoring systems assessment. In: Proceedings of the 27th
  International Technical Conference on the Enhanced Safety of Vehicles (ESV)
  National Highway Traffic Safety Administration. No. 23-0286 (2023)

\bibitem{pandey2020skin}
Pandey, P., Tyagi, A.K., Ambekar, S., Prathosh, A.: Unsupervised domain
  adaptation for semantic segmentation of nir images through generative latent
  search. In: European Conference on Computer Vision. pp. 413--429. Springer
  (2020)

\bibitem{podell2023sdxl}
Podell, D., English, Z., Lacey, K., Blattmann, A., Dockhorn, T., M{\"u}ller,
  J., Penna, J., Rombach, R.: Sdxl: Improving latent diffusion models for
  high-resolution image synthesis. arXiv preprint arXiv:2307.01952  (2023)

\bibitem{Pyo2025NIRBlocking}
Pyo, J.Y., Kim, M.C., Oh, S.J., Mah, K.C., Jang, J.Y.: Evaluation of optical
  and thermal properties of nir-blocking ophthalmic lenses under controlled
  conditions. Sensors  \textbf{25}(11), ~3556 (2025)

\bibitem{Rahul2025}
Rahul, D., Sreelekshmi, P., Amritha, P., Krishna, A.: Application of gans in
  high-resolution image synthesis domain adaptation and image-to-image
  translation. In: Revolution with Generative AI: Trends and Techniques, pp.
  21--41. Springer (2025)

\bibitem{ranftl2021vision}
Ranftl, R., Bochkovskiy, A., Koltun, V.: Vision transformers for dense
  prediction. In: Proceedings of the IEEE/CVF international conference on
  computer vision. pp. 12179--12188 (2021)

\bibitem{rangesh2020driver}
Rangesh, A., Zhang, B., Trivedi, M.M.: Driver gaze estimation in the real
  world: Overcoming the eyeglass challenge. In: 2020 IEEE Intelligent vehicles
  symposium (IV). pp. 1054--1059. IEEE (2020)

\bibitem{richter2016playing}
Richter, S.R., Vineet, V., Roth, S., Koltun, V.: Playing for data: Ground truth
  from computer games. In: European conference on computer vision. pp.
  102--118. Springer (2016)

\bibitem{rombach2022high}
Rombach, R., Blattmann, A., Lorenz, D., Esser, P., Ommer, B.: High-resolution
  image synthesis with latent diffusion models. In: Proceedings of the IEEE/CVF
  conference on computer vision and pattern recognition. pp. 10684--10695
  (2022)

\bibitem{ryu2022lora}
Ryu, S.: Low-rank adaptation for fast text-to-image diffusion fine-tuning.
  Software repository (2022)

\bibitem{shaik2024idd}
Shaik, F.A., Reddy, A., Billa, N.R., Chaudhary, K., Manchanda, S., Varma, G.:
  Idd-aw: A benchmark for safe and robust segmentation of drive scenes in
  unstructured traffic and adverse weather. In: Proceedings of the IEEE/CVF
  Winter Conference on Applications of Computer Vision. pp. 4614--4623 (2024)

\bibitem{simonyan2015deepconvolutionalnetworkslargescale}
Simonyan, K., Zisserman, A.: Very deep convolutional networks for large-scale
  image recognition. In: 3rd International Conference on Learning
  Representations, {ICLR} 2015, San Diego, CA, USA, May 7-9, 2015, Conference
  Track Proceedings (2015)

\bibitem{SonyIMX7752025}
{Sony Semiconductor Solutions Corporation}: Sony releases rgb-ir image sensor
  imx775 for in-cabin monitoring cameras (2025),
  \url{https://www.sony-semicon.com/en/news/2025/2025100201.html}

\bibitem{torralba2011datasetbias}
Torralba, A., Efros, A.A.: Unbiased look at dataset bias. In: CVPR 2011. pp.
  1521--1528. IEEE (2011)

\bibitem{xie2021segformer}
Xie, E., Wang, W., Yu, Z., Anandkumar, A., Alvarez, J.M., Luo, P.: Segformer:
  Simple and efficient design for semantic segmentation with transformers.
  Advances in neural information processing systems  \textbf{34},  12077--12090
  (2021)

\bibitem{xinsir6controlnetplus}
xinsir6: {ControlNetPlus}: All-in-one {ControlNet} for image generation and
  editing (2024), \url{https://github.com/xinsir6/ControlNetPlus}, model
  weights available on HuggingFace. Accessed: Oct 2025

\bibitem{yang2024comprehensive}
Yang, G., Ridgeway, C., Miller, A., Sarkar, A.: Comprehensive assessment of
  artificial intelligence tools for driver monitoring and analyzing safety
  critical events in vehicles. Sensors  \textbf{24}(8), ~2478 (2024)

\bibitem{yang2020fda}
Yang, Y., Soatto, S.: Fda: Fourier domain adaptation for semantic segmentation.
  In: Proceedings of the IEEE/CVF conference on computer vision and pattern
  recognition. pp. 4085--4095 (2020)

\bibitem{zhang2023adding}
Zhang, L., Rao, A., Agrawala, M.: Adding conditional control to text-to-image
  diffusion models. In: Proceedings of the IEEE/CVF international conference on
  computer vision. pp. 3836--3847 (2023)

\bibitem{zhao2023uni}
Zhao, S., Chen, D., Chen, Y.C., Bao, J., Hao, S., Yuan, L., Wong, K.Y.K.:
  Uni-controlnet: All-in-one control to text-to-image diffusion models.
  Advances in neural information processing systems  \textbf{36},  11127--11150
  (2023)

\bibitem{zhao2022shade}
Zhao, Y., Zhong, Z., Zhao, N., Sebe, N., Lee, G.H.: Style-hallucinated dual
  consistency learning for domain generalized semantic segmentation. In:
  European conference on computer vision. pp. 535--552. Springer (2022)

\end{thebibliography}

\clearpage
\newpage
\section*{Appendix}
\appendix

\section{Camera Systems - Near Infrared}
\label{sec:nir}
Near-infrared (NIR) spectrum cameras have become the de facto choice for in-cabin sensing, providing consistent image quality from pitch-dark cabins to sunlit conditions and supporting both driver-monitoring (DMS) and occupant-monitoring (OMS) functions~\cite{yang2024comprehensive}. For exterior NIR-cameras are researched but limited by active illumination and not covering colors. To this end, they are typically paired with active 850-940\,nm LED or Vertical-Cavity Surface-Emitting Laser (VCSEL) illumination, driven in a pulsed regime of very short, high-intensity flashes synchronized with the camera shutter~\cite{LaserFocusWorld2013}. 
Assessment roadmaps further reinforce this shift toward direct, camera-based monitoring, with Euro\,NCAP's 2026 protocols placing greater emphasis on robust DMS/OMS performance and EU Regulation~2019/2144 mandating direct driver attention monitoring for all new vehicle types from July~2024~\cite{PalaoEtAl2023,EURegulation2019}. Compared with visible imaging, NIR illumination is effectively invisible to occupants, reducing distraction while maintaining high machine-vision signal quality. At 850\,nm, sensors benefit from higher responsivity but the emitter produces a faint residual red glow, whereas 940\,nm is genuinely covert at the source~\cite{LaserFocusWorld2013}. 
The same imaging stack extends to cabin-wide OMS features such as occupant presence and classification and child-presence detection~\cite{yang2024comprehensive}. 
While prior in-cabin datasets exist, SVIRO targets the rear-seat only and provides a limited set of occupant/object classes (e.g., empty seat, child seat, passenger), which constrains generalization to broader OMS scenarios~\cite{cruz2020sviro}. 
Outdoor navigation is tackled by exterior NIR datasets such as RANUS, which provides a large spatially-aligned RGB–NIR urban-scene corpus for road-scene parsing~\cite{choe2018ranus}.
Modern modules frequently employ RGB-IR sensors so that a single camera can deliver high-contrast IR for machine-vision algorithms and, when required, color for user-facing features~\cite{SonyIMX7752025}. 
Finally, real-world validation must account for eye-wear variability, as standard lenses are largely transparent at NIR wavelengths whereas lenses with explicit NIR-blocking coatings occlude ocular cues and can cause silent failure of pupil-based pipelines, necessitating test coverage across lens types during development~\cite{Pyo2025NIRBlocking,rangesh2020driver}.

\FloatBarrier
\section{Detailed Dataset Description}
In this section, we provide more details on used datasets within this work.

\subsection{Interior Sensing}
\label{app:interior}

\begin{figure}[t] 
  \centering
  \includegraphics[width=0.9\linewidth]{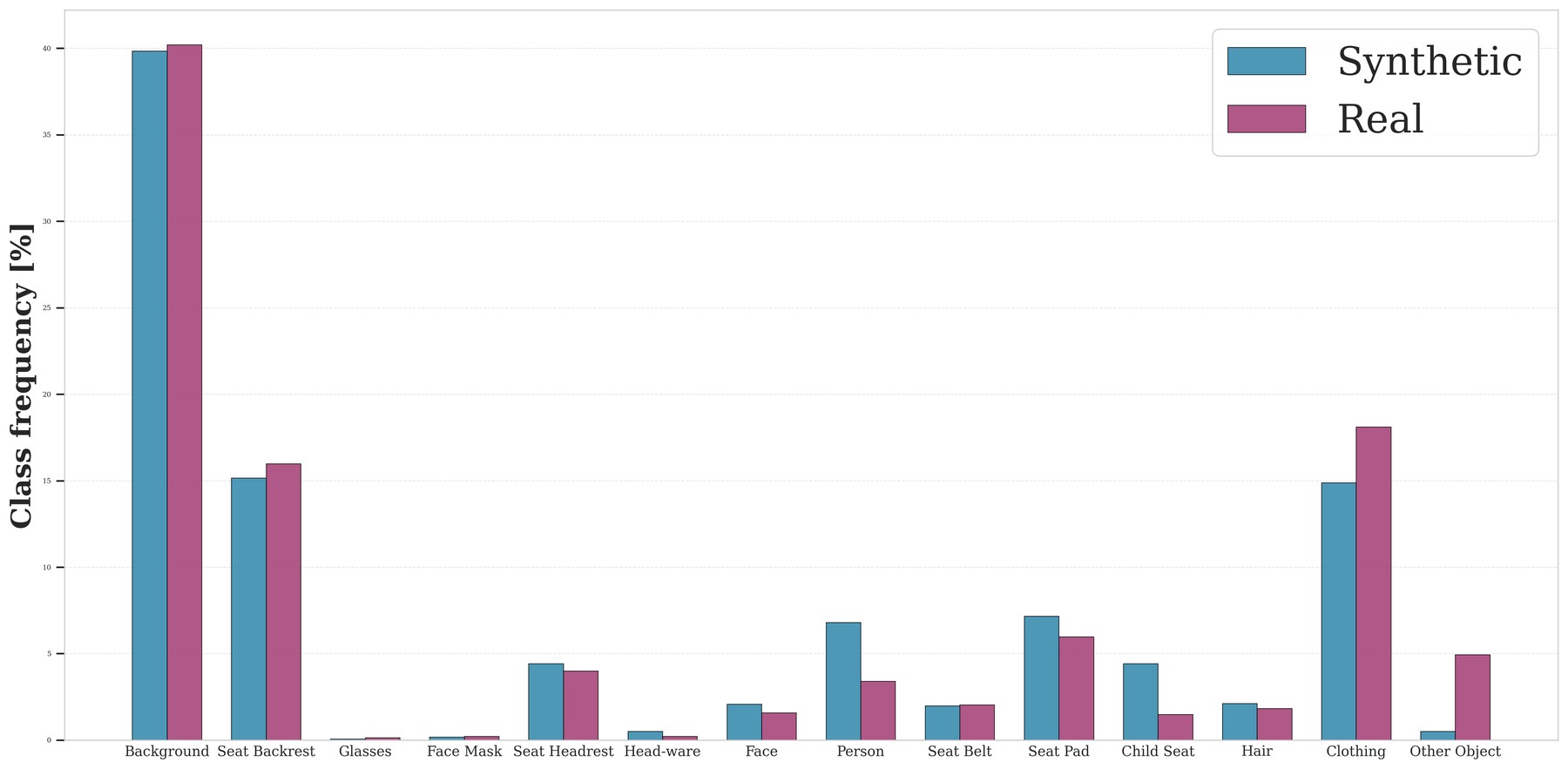}
  \caption{Label distributions interior dataset.}
  \label{fig:app_label_dist_interior}
\end{figure}

Our interior sensing dataset provides a coarse and fine-grained classes for semantic segmentation in vehicle's interiors. The taxonomy covers both vehicle components as well as human-centric classes (see Fig. \ref{fig:app_label_dist_interior} for class distribution). The combination of these classes enables reasoning about occupant presence, their pose, and potential interactions with surrounding objects. Especially important are safety-relevant elements such as seat belts, which can be checked via semantic segmentation to determine whether they are properly fastened, whether a person is distracted by food or other occupants, or whether the person is safely seated in the seat in case of a crash. Also it is possible to detect left behind children or objects inside the vehicle (see \ref{des:intclasses}).  
Dataset challenges arise from the high visual diversity and complex interactions between the annotated classes. Vehicle interiors vary widely in materials, geometry, and seat designs, while human‑centric classes exhibit large appearance variations in clothing, hairstyles, and accessories. Many classes are small, thin, or frequently occluded (e.g., seat belts, glasses, hair), and strong visual similarity between seat components further complicates segmentation. In addition, objects such as child seats, drinks, or bags introduce further variability and create complex spatial relationships between occupants, seats, and personal items.
In the following, we provide a description of the classes of our in-cabin dataset:

\begin{description}
\label{des:intclasses}
  \item[\textbf{background}]
  All pixels that do not belong to any annotated foreground class. This includes unassigned interior surfaces such as doors or dashboard areas without specific labels.

  \item[\textbf{seat-backrest}]
  Upright rear section of a seat that supports the torso. It includes seams and upholstery but it does not include the headrest or the seat pad.

  \item[\textbf{seat-headrest}]
  Upper padded structure that supports the occupant’s head.

  \item[\textbf{seat-pad}]
  Horizontal seating surface on which the occupant sits.

  \item[\textbf{seat-belt}]
  All visible parts of the safety belt system.

  \item[\textbf{child-seat}]
  Child seats for all age children. This includes the full shell the padding and any attached harness structures.

  \item[\textbf{person}]
  Coarse full body person mask that serves as a base region. More specific classes such as face hair and clothing are placed on top where relevant. In single label per pixel annotation this class represents visible skin that does not belong to any of the more specific person classes.

  \item[\textbf{clothing}]
  All visible garments such as jackets shirts trousers skirts shoes scarves and ties.

  \item[\textbf{hair}]
  All natural hair on the head excluding facial hair.

  \item[\textbf{face}]
  Frontal facial region that covers the forehead up to the hairline the cheeks the nose the lips and the chin.

  \item[\textbf{glasses}]
  All types of eyewear such as prescription glasses sunglasses and protective goggles. Both frames and lenses are included.

  \item[\textbf{face-mask}]
  Any mouth and nose covering such as medical masks fabric masks or respirators.

  \item[\textbf{head-ware}]
  Any accessory worn on the head such as caps beanies hats hoods or helmets.

  \item[\textbf{other object}]
  Any object inside the vehicle cabin that does not belong to any defined foreground class and is not structurally part of the vehicle.

\end{description}
\FloatBarrier
\subsection{Exterior Sensing}
\label{app:exterior}

\begin{figure}[t]
    \centering
    \includegraphics[width=\linewidth]{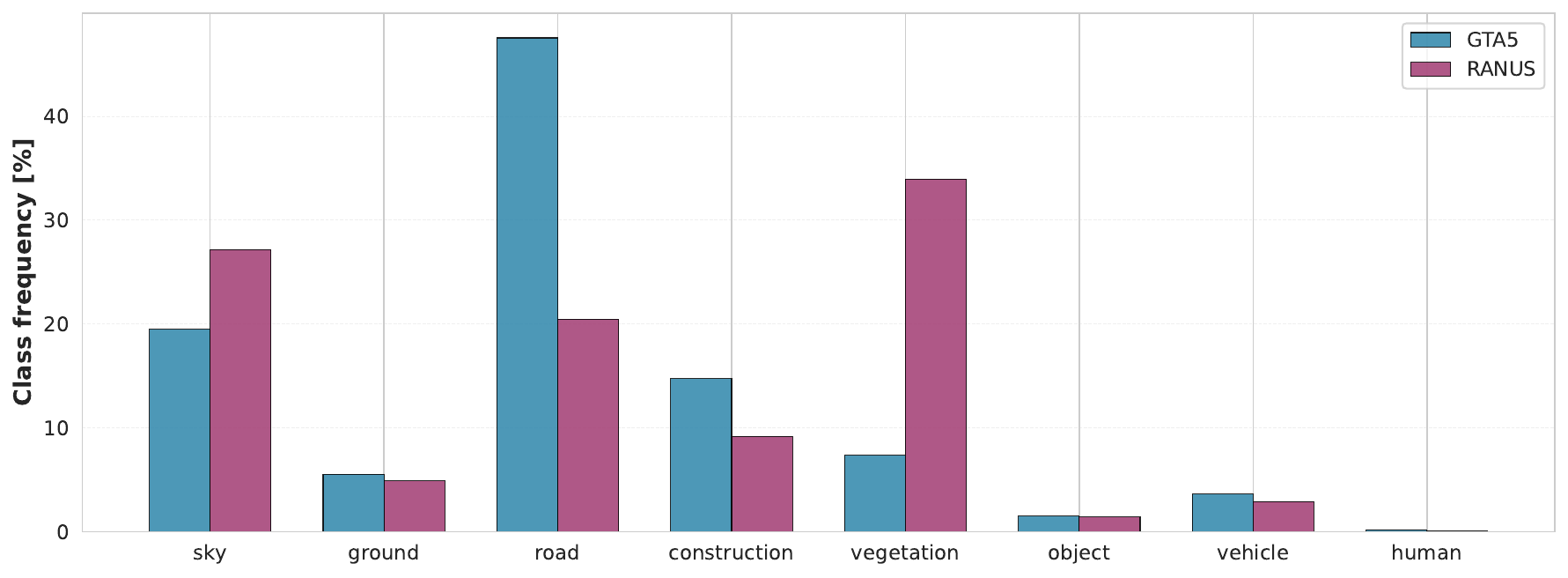}
    \caption{Distributions of class labels in the selected subset of GTA 5 and RANUS images after applied class mapping to common classes.}
    \label{fig:label_dists_ranus_gta}
\end{figure}

Our exterior sensing datasets cover two public available datasets, i.e., synthetic
GTA5~\cite{richter2016playing} and real-world
RANUS~\cite{choe2018ranus}. 
Note, alternative NIR datasets, such as TAS-NIR~\cite{mortimer2022tas} and IDD-AW~\cite{shaik2024idd}, were excluded as they target off-road or adverse-weather scenarios, while aerial and agricultural NIR datasets lack the semantic scope required for urban driving evaluation.
GTA5 is rendered from a video game environment depicting a large American
city and originally provides only RGB frames with a 19-class Cityscapes
taxonomy. All GTA5 images are resized and center-cropped to $1024\times1024$ pixels which is the native SDXL resolution and corresponds to the RANUS image size.
To obtain paired NIR imagery, we generate synthetic NIR images using our TSA method trained on a subset of curated RANUS images.
From an initial pool of 8,500 generated image pairs, we discard those
where the NIR output has a per-channel standard deviation above 0.5, leaving 6,766 valid pairs.
RANUS provides paired RGB and NIR frames captured with co-registered Point-Grey Grasshopper cameras (RGB: GS3-U3-41C6C-C; NIR: GS3-U3-41C6NIR-C) in and around Seoul, South Korea.
The geographic setting differs substantially from GTA5, which
introduces an additional appearance gap on top of the synth2real domain shift. 
RANUS originally uses a 10-class taxonomy, so we map both datasets onto the same 8 shared classes: \textit{sky},
\textit{ground}, \textit{road}, \textit{construction}, \textit{vegetation},
\textit{object}, \textit{vehicle}, and \textit{human}. The label distributions are visualized in \autoref{fig:label_dists_ranus_gta}, which reveals a
notable class distribution mismatch between the two domains, as road pixels
dominate in GTA5 (${\sim}47\,\%$) while vegetation is the largest class in
RANUS (${\sim}34\,\%$).

The original RANUS annotations contained labeling inconsistencies and compression-induced color deviations, which were corrected via fuzzy label mapping using an L1 color distance threshold of 10. To further improve annotation quality, pseudo ground truth was generated using a Mask2Former model~\cite{cheng2022masked} pretrained on Cityscapes~\cite{Cordts2016Cityscapes}, mapped to the 8 semantic classes shared between both datasets. Images were retained only where agreement between original and pseudo labels exceeded 80\%. 
Pseudo labels were used exclusively for this filtering step and not during training or evaluation, ensuring that all supervision remains grounded in human-annotated data.

\FloatBarrier
\section{Training Details}
\label{sec:traindetails}

\subsection{NIR LoRA Training}
\label{subsec:loratrain}
We train text-to-image LoRA (rank 8) for the SDXL base model~\cite{podell2023sdxl} for 2000 iterations using a constant learning rate of $10^{-4}$, batch size 2, and AdamW optimization. Text encoder fine-tuning was omitted following Stability AI recommendations for SDXL's dual encoder architecture. Training captions were derived from 65 diversity-sampled RANUS images, captioned via \texttt{gemini-2.5-flash-001}~\cite{comanici2025gemini} using structured JSON output. The caption schema decomposed each scene into setting, object lists, and spatial relations; style and mood fields were stripped and replaced with a dedicated \texttt{<nirstyle>} token to encode NIR appearance implicitly without leaking spectral characteristics through text. All captions were manually reviewed and corrected, with only a small number of complex scenes requiring substantial edits. An example caption takes the form: \texttt{<nirstyle> urban street, car, building, person, road, crosswalk, street sign, street light}.

\subsection{Synthetic NIR Generation}
\label{sec:nir_generation}
\begin{figure}[t]
  \newcommand{\w}{0.3\linewidth} 

  \centering
  \begin{subfigure}[t]{\w}
    \centering
    \includegraphics[width=\linewidth]{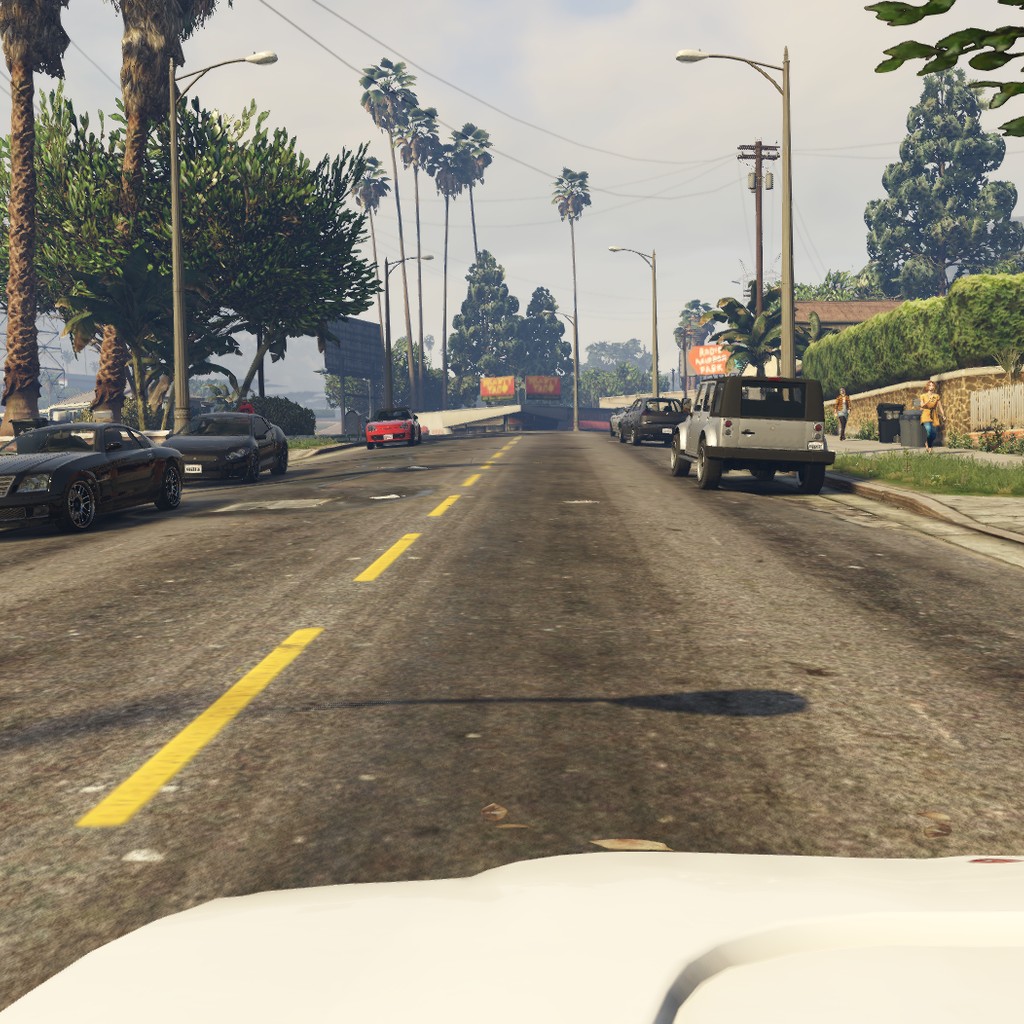}
    \caption{Original Synthetic Image}
  \end{subfigure}\hfill
  \begin{subfigure}[t]{\w}
    \centering
    \includegraphics[width=\linewidth]{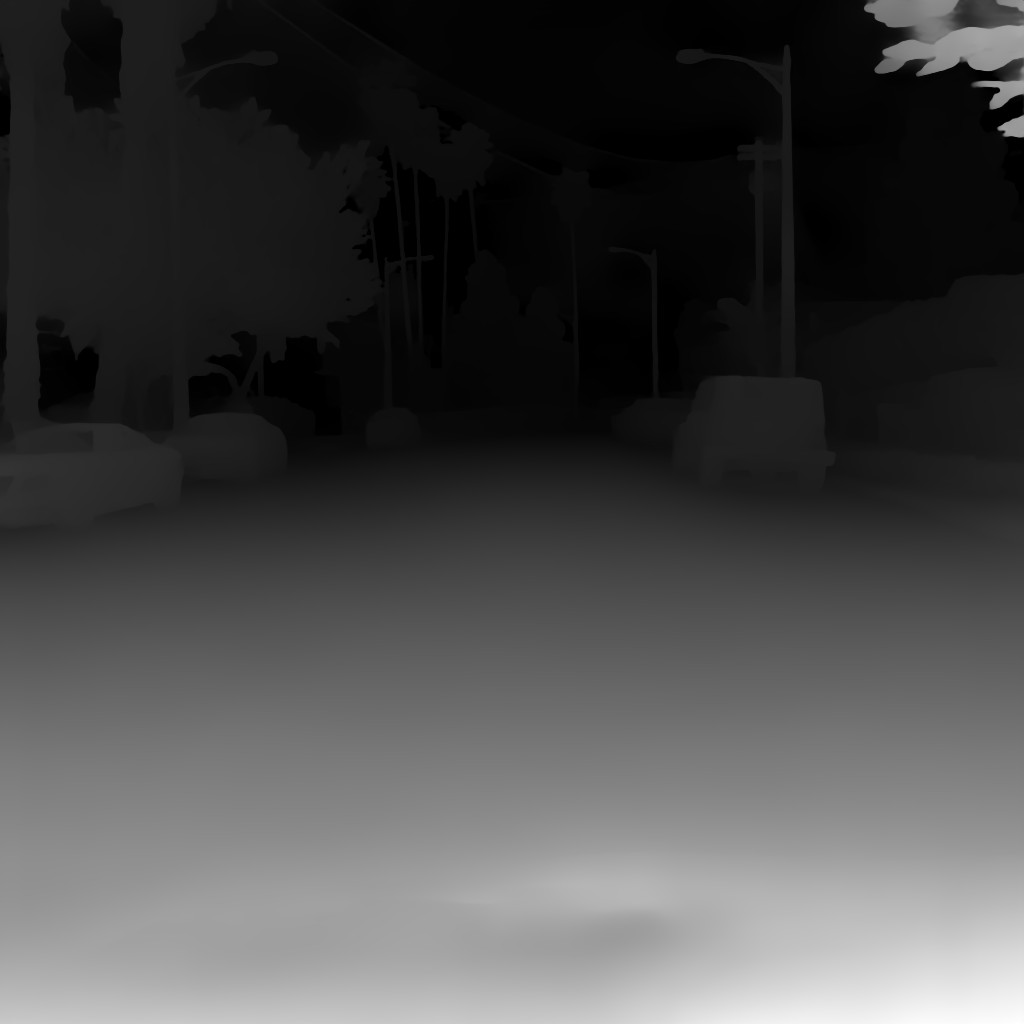}
    \caption{Depth (Image)}
  \end{subfigure}\hfill
  \begin{subfigure}[t]{\w}
    \centering
    \includegraphics[width=\linewidth]{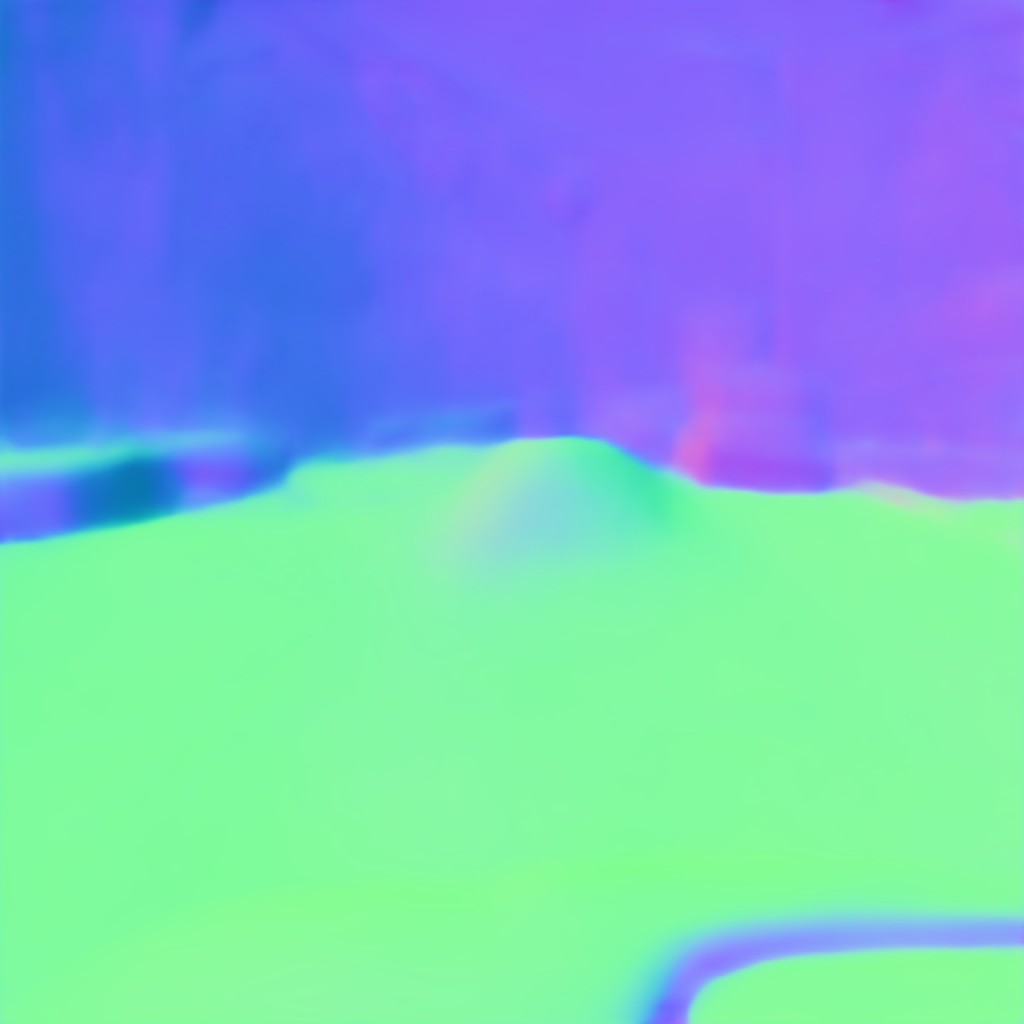}
    \caption{Normal Map (Image)}
  \end{subfigure} \\
  \begin{subfigure}[t]{\w}
    \centering
    \includegraphics[width=\linewidth]{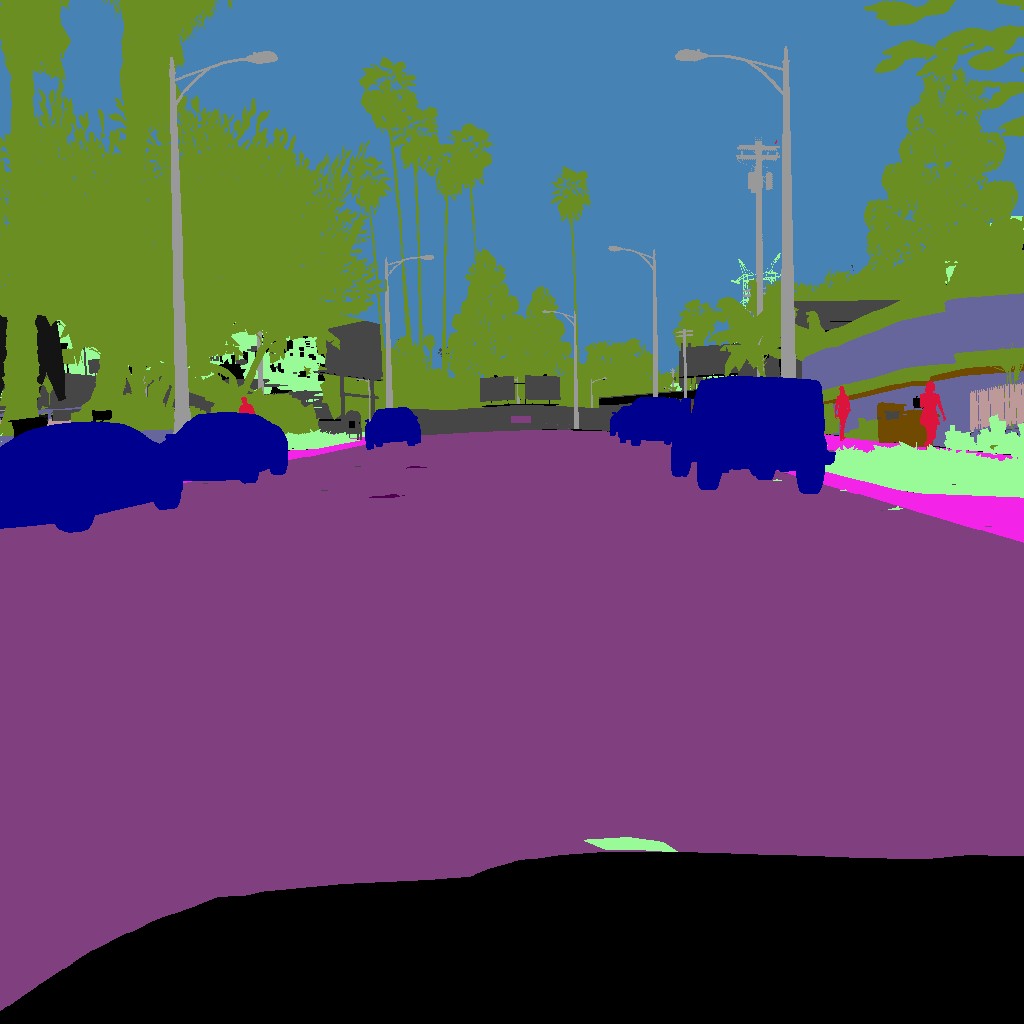}
    \caption{Semantic Segmentation}
  \end{subfigure}\hfill
  \begin{subfigure}[t]{\w}
    \centering
    \includegraphics[width=\linewidth]{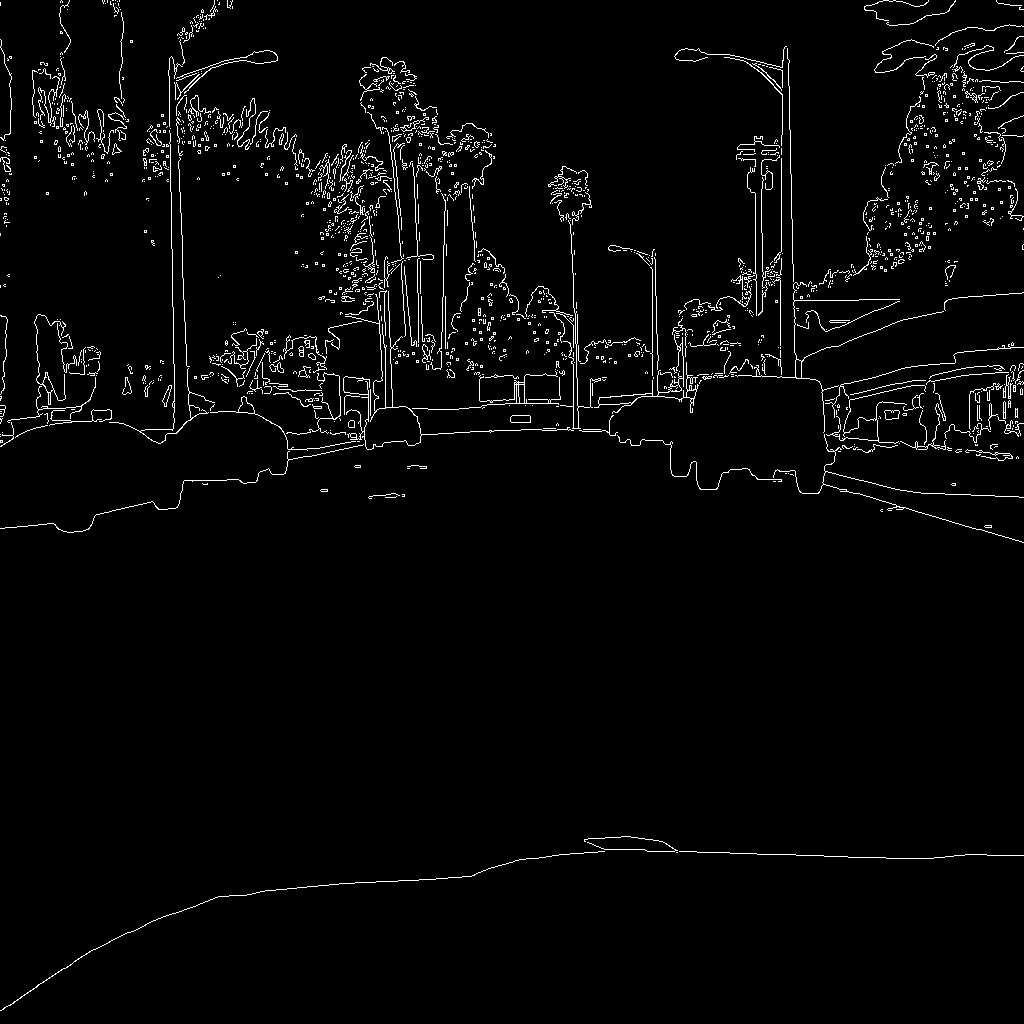}
    \caption{Canny (Segmentation)}
  \end{subfigure}\hfill
  \begin{subfigure}[t]{\w}
    \centering
    \includegraphics[width=\linewidth]{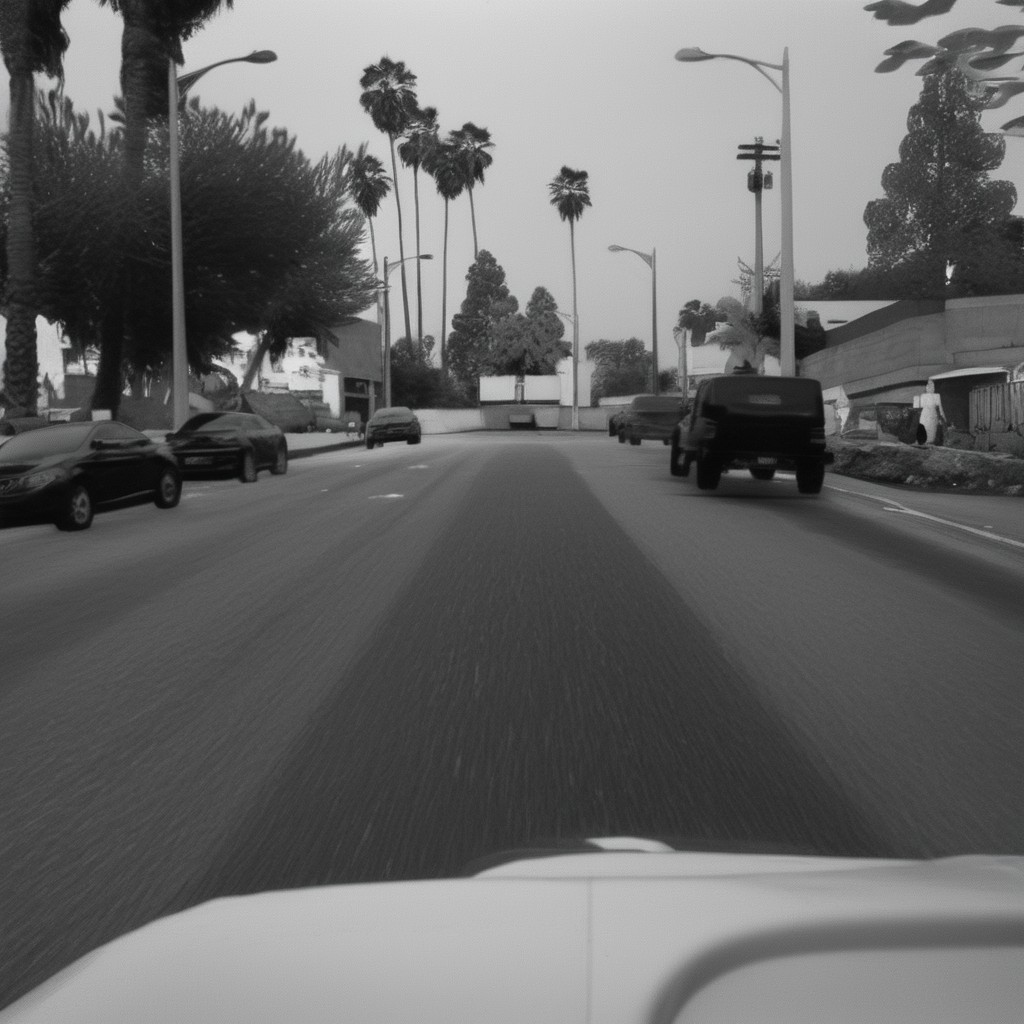}
    \caption{TSA}
  \end{subfigure}
  \caption{TSA with multiple conditioning signals for exterior data.
  }
  \label{fig:lora_exterior}
\end{figure}

All GTA5 images were resized to $1024 \times 1024$ via shortest-edge scaling and center crop, matching both the RANUS image resolution and the native resolution of SDXL. NIR images were generated using ControlNet-Union (ProMax)~\cite{xinsir6controlnetplus} conditioned on three simultaneous signals: Canny edges derived from segmentation masks, MiDaS depth maps~\cite{ranftl2021vision} and Normal-Bae surface normals~\cite{bae2021estimating}, at control strengths of 0.9, 0.2, and 0.3 respectively, visualized in Figure \ref{fig:lora_exterior}. The full inference parameters are reported in Table~\ref{tab:nir_inference_params}.

Inference parameters were selected based on visual inspection and two proxy metrics suited to NIR imagery: channel standard deviation (real NIR images are monochromatic and thus exhibit zero channel variance) and edge correlation between the input RGB and generated output. Standard metrics such as Fréchet Inception Distance  
were avoided as they rely on RGB-pretrained feature networks and no established reference distribution exists for synthetic NIR data. Of 8,500 generated images, 6,766 (79.6\%) passed a channel standard deviation threshold of 0.5 and were retained. The accepted images were split to mirror the filtered RANUS proportions, ensuring comparable data scale across domains.

\begin{table}[t]
\centering
\caption{SDXL LoRA inference parameters used for synthetic NIR generation.}
\label{tab:nir_inference_params}
\begin{tabular}{ll}
\hline
\textbf{Parameter} & \textbf{Value} \\
\hline
Base prompt & \texttt{nir, urban street, driver-view} \\
Negative prompt & (octane render, render, drawing, anime, bad photo, bad photography, graffiti, \\
& painting), (worst quality, low quality, blurry),  (daylight colors, natural colors, \\
 & warm tones, cool tones, color temperature, hue shift, color balance) \\
Scheduler type & Scheduled \\
LoRA strength & 0.99 \\
Inference steps & 50 \\
Scheduler start & 1.2 \\
Scheduler end & 0.3 \\
Scheduler decay steps & 30 \\
Control modes & \{1, 3, 4\} \\
Canny control strength & 0.9 \\
Depth control strength & 0.2 \\
Normal control strength & 0.3 \\
ControlNet version & ProMax \\
\hline
\end{tabular}
\end{table}

\subsection{TSA Inference Hyperparameters}
\label{subsec:hyperparamter}
ControlNet~\cite{zhang2023adding} augments diffusion models with trainable spatial-control branches attached to a frozen backbone, enabling conditioning on structured signals such as edges, depth, or segmentation maps. Uni-ControlNet~\cite{zhao2023uni} extends this idea by introducing shared adapters that support multiple control modalities without increasing model size per added condition. 

In the interior setting, we determined through the TSA-hyperparameter investigation an inference configuration of 30 denoising steps with a denoising strength of 0.5, a Canny mask conditioning weight of 1.0, a segmentation conditioning weight of 0.5, and a Canny image conditioning weight of 0.5. We observed that, depth and normal conditioning had no significant effect on the final output. Higher denoising strength resulted in a more pronounced sepia tone, while too few inference steps caused misaligned structures and RGB artifacts. Conversely, too little denoising strength produced only slight restructuring and did not meaningfully alter the overall texture. Direct segmentation conditioning also had no major impact. The most critical factor was the application of Canny edges to the segmentation masks. Incorporating real Canny edges further improved texture fidelity and prevented instances of the same class from collapsing into a single object when their masks overlapped, thereby avoiding incorrect instance merging by the LDM. However, if the canny‑over‑mask strength was too weak, certain small segments, such as glasses or face masks were partially destroyed.  

For the exterior use-case, a qualitative analysis on the test images resulted in an inference configuration of 50 denoising steps with a strength of 0.99, canny mask conditioning weight of 0.9, estimated depth conditioning weight of 0.2 and surface normal conditioning weight of 0.3. 
Depth and surface normal conditioning showed to be beneficial on the exterior images to improve the scene composition and object shapes for more distant objects.
The differences in inference parameters largely result from the distinct case of RGB to NIR translation and more variability in object sizes and distances. We note that alternative configurations may yield comparable results, and the chosen values represent a practical optimum under our experimental constraints identified by visual inspection of generated samples with varying inference parameters. 
\FloatBarrier
\subsection{Details on Stylization Method.} 
\begin{figure*}[t]
    \centering
    \captionsetup[subfigure]{labelformat=empty} 
    \begin{subfigure}[t]{0.124\textwidth}
        \centering
        \includegraphics[width=\textwidth]{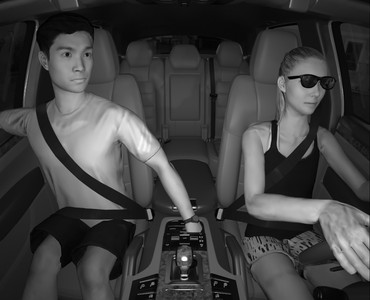}
        \caption{Synthetic Baseline}
    \end{subfigure}
    \hfill
    \begin{subfigure}[t]{0.124\textwidth}
        \centering
        \includegraphics[width=\textwidth]{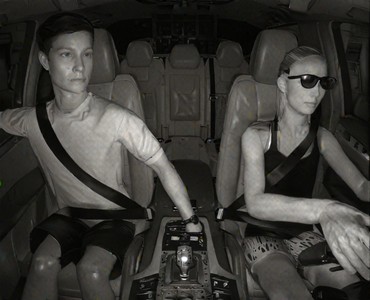}
        \caption{TSA Only}
    \end{subfigure}
    \hfill
    \begin{subfigure}[t]{0.124\textwidth}
        \centering
        \includegraphics[width=\textwidth]{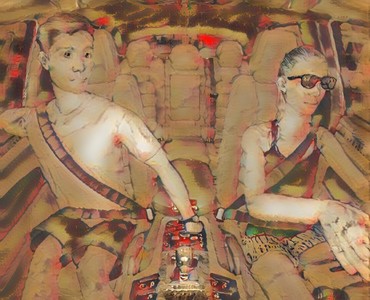}
        \caption{VSD(1,1.0) (RGB)}
    \end{subfigure}
    \hfill
    \begin{subfigure}[t]{0.124\textwidth}
        \centering
        \includegraphics[width=\textwidth]{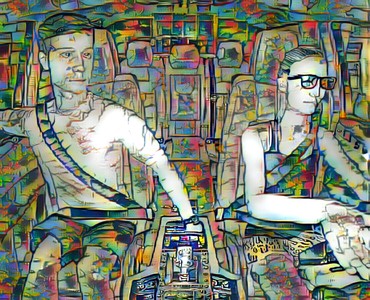}
        \caption{VSD(1,1.0) (Gray)}
    \end{subfigure}
    \hfill
    \begin{subfigure}[t]{0.124\textwidth}
        \centering
        \includegraphics[width=\textwidth]{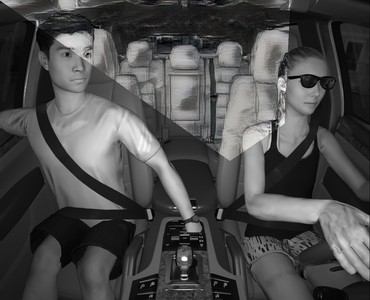}
        \caption{VSD(4,0.25) (RGB)}
    \end{subfigure}
    \hfill
    \begin{subfigure}[t]{0.124\textwidth}
        \centering
        \includegraphics[width=\textwidth]{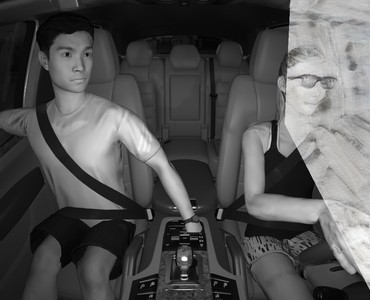}
        \caption{VSD(4,0.25) (Gray)}
    \end{subfigure}
    \hfill
    \begin{subfigure}[t]{0.124\textwidth}
        \centering
        \includegraphics[width=\textwidth]{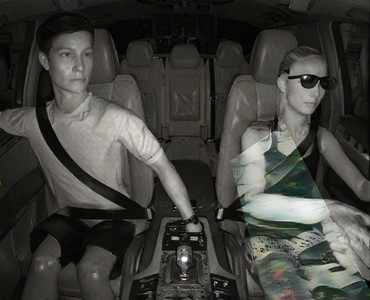}
        \caption{VSD(4,0.25)+ TSA (RGB)}
    \end{subfigure}
    \\[6pt]
    \begin{subfigure}[t]{0.124\textwidth}
        \centering
        \includegraphics[width=\textwidth]{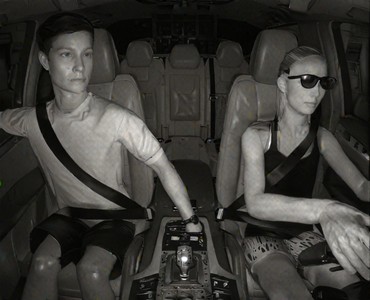}
        \caption{VSD(4,0.25)+ TSA (Gray)}
    \end{subfigure}
    \hfill
    \begin{subfigure}[t]{0.124\textwidth}
        \centering
        \includegraphics[width=\textwidth]{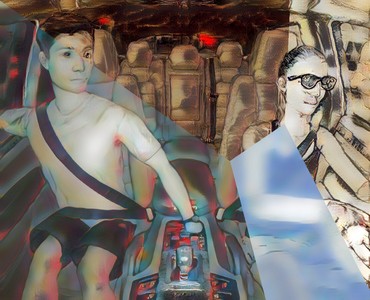}
        \caption{VSD(4,0.75) (RGB)}
    \end{subfigure}
    \hfill
    \begin{subfigure}[t]{0.124\textwidth}
        \centering
        \includegraphics[width=\textwidth]{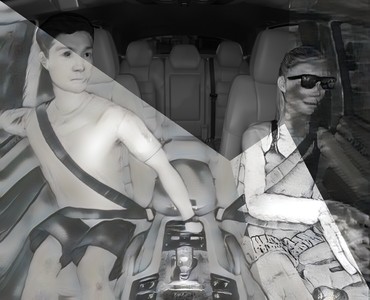}
        \caption{VSD(4,0.75) (Gray)}
        \label{fig:av_V4_p=0.75_gray}
    \end{subfigure}
    \hfill
    \begin{subfigure}[t]{0.124\textwidth}
        \centering
        \includegraphics[width=\textwidth]{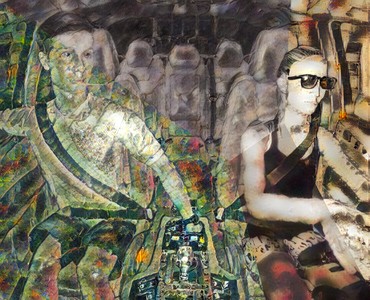}
        \caption{VSD(4,0.75)+ TSA (RGB)}
        \label{fig:av_V4_p=0.75_plusLoRA_color}
    \end{subfigure}
    \hfill
    \begin{subfigure}[t]{0.124\textwidth}
        \centering
        \includegraphics[width=\textwidth]{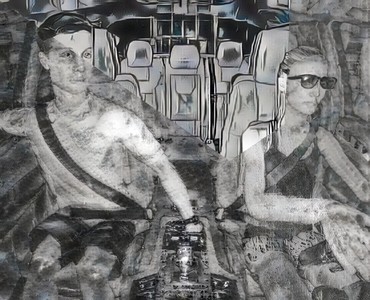}
        \caption{VSD(4,0.75)+ TSA (Gray)}
        \label{fig:av_V4_p=0.75_plusLoRA_gray}
    \end{subfigure}
    \hfill
    \begin{subfigure}[t]{0.124\textwidth}
        \centering
        \includegraphics[width=\textwidth]{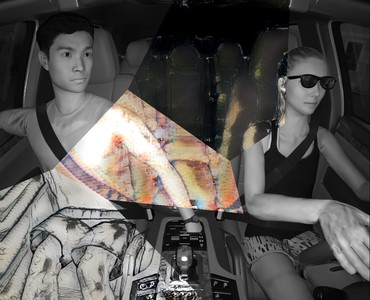}
        \caption{VSD(8,0.25) (RGB)}
        \label{fig:av_V8_p=0.25_color}
    \end{subfigure}
    \hfill
    \begin{subfigure}[t]{0.124\textwidth}
        \centering
        \includegraphics[width=\textwidth]{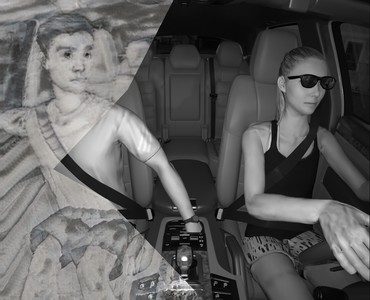}
        \caption{VSD(8,0.25) (Gray)}
        \label{fig:av_V8_p=0.25_gray}
    \end{subfigure}
    \\[6pt]
    \begin{subfigure}[t]{0.124\textwidth}
        \centering
        \includegraphics[width=\textwidth]{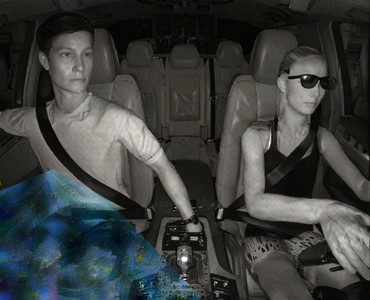}
        \caption{VSD(8,0.25)+ TSA (RGB)}
        \label{fig:av_V8_p=0.25_plusLoRA_color}
    \end{subfigure}
    \hfill
    \begin{subfigure}[t]{0.124\textwidth}
        \centering
        \includegraphics[width=\textwidth]{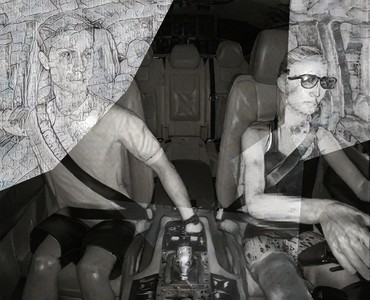}
        \caption{VSD(8,0.25)+ TSA (Gray)}
        \label{fig:av_V8_p=0.25_plusLoRA_gray}
    \end{subfigure}
    \hfill
    \begin{subfigure}[t]{0.124\textwidth}
        \centering
        \includegraphics[width=\textwidth]{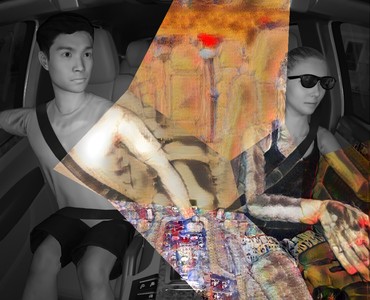}
        \caption{VSD(8,0.75) (RGB)}
        \label{fig:av_V8_p=0.75_color}
    \end{subfigure}
    \hfill
    \begin{subfigure}[t]{0.124\textwidth}
        \centering
        \includegraphics[width=\textwidth]{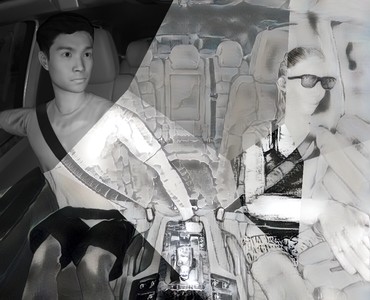}
        \caption{VSD(8,0.75) (Gray)}
        \label{fig:av_V8_p=0.75_gray}
    \end{subfigure}
    \hfill
    \begin{subfigure}[t]{0.124\textwidth}
        \centering
        \includegraphics[width=\textwidth]{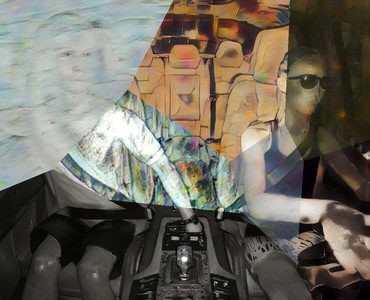}
        \caption{VSD(8,0.75)+ TSA (RGB)}
        \label{fig:av_V8_p=0.75_plusLoRA_color}
    \end{subfigure}
    \hfill
    \begin{subfigure}[t]{0.124\textwidth}
        \centering
        \includegraphics[width=\textwidth]{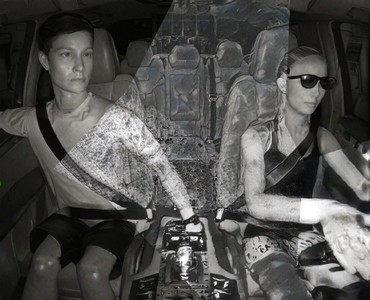}
        \caption{VSD(8,0.75)+ TSA (Gray)}
        \label{fig:av_V8_p=0.75_plusLoRA_gray}
    \end{subfigure}
    \hfill
    \begin{subfigure}[t]{0.124\textwidth}
        \centering
        \includegraphics[width=\textwidth]{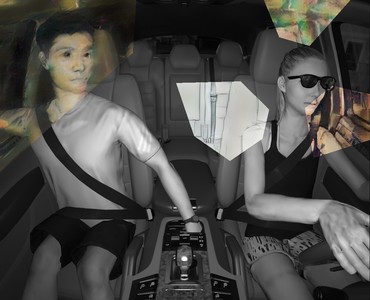}
        \caption{VSD(16,0.25) (RGB)}
        \label{fig:av_V16_p=0.25_color}
    \end{subfigure}
    \\[6pt]
    \begin{subfigure}[t]{0.124\textwidth}
        \centering
        \includegraphics[width=\textwidth]{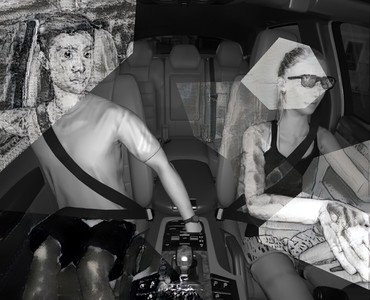}
        \caption{VSD(16,0.25) (Gray)}
        \label{fig:av_V16_p=0.25_gray}
    \end{subfigure}
    \hfill
    \begin{subfigure}[t]{0.124\textwidth}
        \centering
        \includegraphics[width=\textwidth]{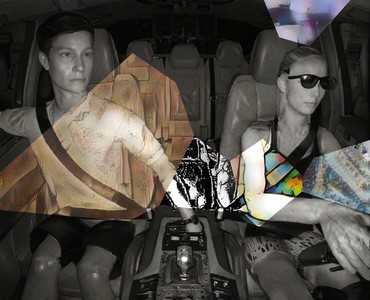}
        \caption{VSD(16,0.25)+ TSA (RGB)}
        \label{fig:av_V16_p=0.25_plusLoRA_color}
    \end{subfigure}
    \hfill
    \begin{subfigure}[t]{0.124\textwidth}
        \centering
        \includegraphics[width=\textwidth]{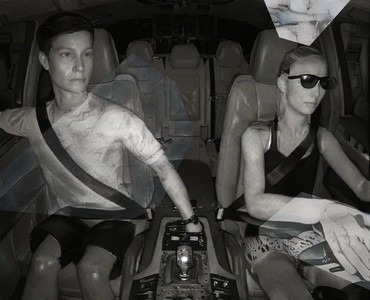}
        \caption{VSD(16,0.25)+ TSA (Gray)}
        \label{fig:av_V16_p=0.25_plusLoRA_gray}
    \end{subfigure}
    \hfill
    \begin{subfigure}[t]{0.124\textwidth}
        \centering
        \includegraphics[width=\textwidth]{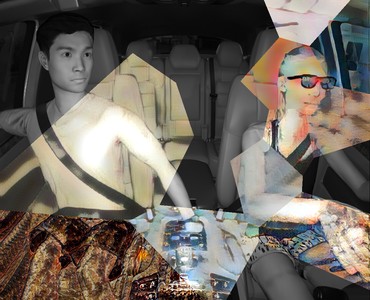}
        \caption{VSD(16,0.75) (RGB)}
        \label{fig:av_V16_p=0.75_color}
    \end{subfigure}
    \hfill
    \begin{subfigure}[t]{0.124\textwidth}
        \centering
        \includegraphics[width=\textwidth]{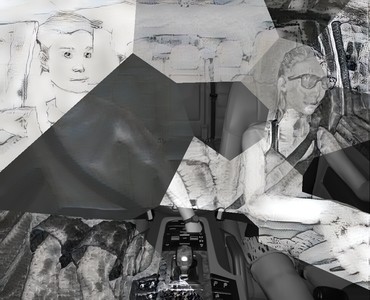}
        \caption{VSD(16,0.75) (Gray)}
        \label{fig:av_V16_p=0.75_gray}
    \end{subfigure}
    \hfill
    \begin{subfigure}[t]{0.124\textwidth}
        \centering
        \includegraphics[width=\textwidth]{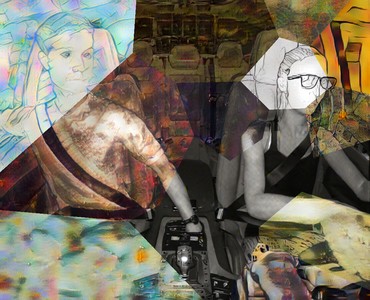}
        \caption{VSD(16,0.75)+ TSA (RGB)}
        \label{fig:av_V16_p=0.75_plusLoRA_color}
    \end{subfigure}
    \hfill
    \begin{subfigure}[t]{0.124\textwidth}
        \centering
        \includegraphics[width=\textwidth]{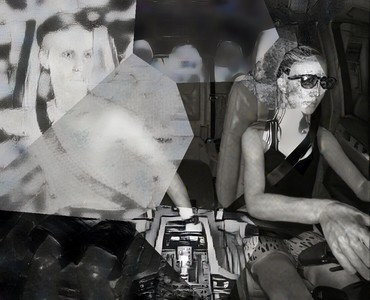}
        \caption{VSD(16,0.75)+ TSA (Gray)}
        \label{fig:av_V16_p=0.75_plusLoRA_gray}
    \end{subfigure}
    \caption{Overview of all augmentation variants for synthetic dataset used for training.}
    \label{fig:overview_datasets_full}
\end{figure*}

We create localized stylization variants using Voronoi partitions with 4, 8, and 16 cells, each combined with stylization probabilities of 0.25 and 0.75 to control how much of the image is altered. 
In addition to these partial stylizations, we also produce a fully stylized dataset in which each image is transformed using a randomly sampled style. 
Furthermore, we evaluate a sequential variant in which TSA is applied first 
followed by stylization. The full set of augmentations is illustrated in \ref{fig:overview_datasets_full}.

To obtain preliminary, pre‑training insights and select suitable candidates, we evaluate the proposed stylization techniques 
using multiple metrics. 
We evaluate pixel‑level similarity between the synthetic baseline and the stylized variants on a per‑image basis using \emph{Peak Signal‑to‑Noise Ratio (PSNR)}, \emph{Structural Similarity Index Measure (SSIM)}, \emph{Root Mean Squared Error (RMSE)}, and \emph{Spectral Angle Mapper (SAM)}, which collectively quantify differences in intensity (PSNR/RMSE), structure (SSIM), and spectral composition (SAM).
Perceptual similarity is assessed using \emph{Learned Perceptual Image Patch Similarity (LPIPS)} and \emph{Deep Image Structure and Texture Similarity (DISTS)}, both of which capture human‑aligned differences in texture, contrast, and structural appearance. To quantify the feature distributional divergence, we investigate the \emph{Fréchet Inception Distance (FID)} for each stylized dataset against both the synthetic and the real reference distributions. This is the only not pixel-wise metric and allows comparison to the target domain.

Table~\ref{tab:synthetic_stylization_similarity_sorted_full} summarizes how each stylization method modifies images across these pixel‑based, perceptual, and distributional dimensions. 
TSA‑only transformations remain comparatively close to the synthetic baseline, indicating moderate appearance changes with preserved structure. 
For the interior data, the fully stylized variant and the Voronoi variants with higher stylization probability introduces the strongest deviations to the synthetic baseline, whereas it is surprisingly that the low stylization methods have a lower FID-score to the real baseline, even below the synthetic baseline. Indicating a closer alignment to the target domain and being a candidate for an improved domain generalization.

\begin{table}[ht]
\centering
\setlength{\tabcolsep}{0.5pt}
\caption{Image similarity of synthetic stylizations against the synthetic baseline (pairwise metrics) and distributional distances (FID). Includes both gray and RGB stylizations. FID from synthetic baseline to real baseline: 156.1.}
\label{tab:synthetic_stylization_similarity_sorted_full}
\begin{tabular}{l c c c c c c cc}
\toprule
\textbf{Method} & \textbf{PSNR\,$\uparrow$} & \textbf{SSIM\,$\uparrow$} & \textbf{RMSE\,$\downarrow$} & \textbf{SAM\,$\downarrow$} & \textbf{LPIPS\,$\downarrow$} & \textbf{DISTS\,$\downarrow$} & \textbf{FID$\rightarrow$Synth\,$\downarrow$} & \textbf{FID$\rightarrow$Real\,$\downarrow$} \\
\midrule
\textit{Synth2Real Baselines} \\
TSA Only & 23.7 & 0.733 & 16.8 & 0.064 & 0.134 & 0.160 & 49.0 & 159.9 \\
VSD(1,1.0) & 12.6 & 0.398 & 69.5 & 0.036 & 0.339 & 0.326 & 281.2 & 218.8 \\
\midrule
\textit{Voronoi (gray) + TSA} \\
VSD(4,0.25) + TSA & 18.5 & 0.627 & 36.1 & 0.055 & 0.211 & 0.202 & 83.1 & 129.5 \\
VSD(4,0.75) + TSA & 12.7 & 0.428 & 64.2 & 0.038 & 0.340 & 0.282 & 240.9 & 175.8 \\
VSD(8,0.25) + TSA & 17.5 & 0.625 & 38.1 & 0.055 & 0.220 & 0.202 & 84.0 & 126.6 \\
VSD(8,0.75) + TSA & 12.2 & 0.421 & 66.4 & 0.038 & 0.353 & 0.279 & 257.8 & 181.8 \\
VSD(16,0.25) + TSA & 16.7 & 0.618 & 40.0 & 0.055 & 0.234 & 0.205 & 90.3 & 126.8 \\
VSD(16,0.75) + TSA & 11.9 & 0.411 & 67.2 & 0.038 & 0.368 & 0.281 & 280.5 & 195.2 \\
\midrule
\textit{Voronoi (rgb) + TSA} \\
VSD(4,0.25) + TSA & 18.3 & 0.619 & 36.4 & 0.094 & 0.249 & 0.218 & 100.2 & 135.8 \\
VSD(4,0.75) + TSA & 12.5 & 0.402 & 64.3 & 0.156 & 0.439 & 0.318 & 291.2 & 209.3 \\
VSD(8,0.25) + TSA & 17.1 & 0.611 & 39.2 & 0.095 & 0.267 & 0.225 & 111.8 & 134.2 \\
VSD(8,0.75) + TSA & 11.9 & 0.396 & 66.7 & 0.154 & 0.455 & 0.320 & 315.5 & 218.3 \\
VSD(16,0.25) + TSA & 16.6 & 0.612 & 39.9 & 0.093 & 0.281 & 0.230 & 120.8 & 134.8 \\
VSD(16,0.75) + TSA & 11.7 & 0.388 & 67.7 & 0.154 & 0.475 & 0.326 & 351.5 & 240.6 \\
\midrule
\textit{Voronoi Only (gray)} \\
VSD(4,0.25) & 17.3 & 0.838 & 29.1 & 0.010 & 0.107 & 0.096 & 37.5 & 122.1 \\
VSD(4,0.75) & 12.8 & 0.533 & 64.1 & 0.023 & 0.285 & 0.244 & 192.3 & 157.1 \\
VSD(8,0.25) & 18.1 & 0.835 & 33.7 & 0.010 & 0.120 & 0.107 & 42.4 & 120.1 \\
VSD(8,0.75) & 12.2 & 0.530 & 65.9 & 0.023 & 0.298 & 0.244 & 202.8 & 159.3 \\
VSD(16,0.25) & 17.8 & 0.828 & 36.4 & 0.010 & 0.138 & 0.121 & 51.4 & 121.7 \\
VSD(16,0.75) & 11.9 & 0.515 & 67.2 & 0.023 & 0.322 & 0.252 & 237.5 & 178.3 \\
\midrule
\textit{Voronoi Only (rgb)} \\
VSD(4,0.25) & 17.1 & 0.831 & 28.4 & 0.050 & 0.147 & 0.121 & 56.1 & 132.6 \\
VSD(4,0.75) & 12.6 & 0.507 & 63.8 & 0.143 & 0.386 & 0.289 & 242.3 & 190.0 \\
VSD(8,0.25) & 17.8 & 0.820 & 34.2 & 0.051 & 0.173 & 0.146 & 73.3 & 135.0 \\
VSD(8,0.75) & 11.9 & 0.493 & 67.0 & 0.142 & 0.413 & 0.300 & 277.8 & 207.2 \\
VSD(16,0.25) & 17.6 & 0.816 & 36.4 & 0.050 & 0.197 & 0.168 & 90.3 & 141.7 \\
VSD(16,0.75) & 11.8 & 0.486 & 67.1 & 0.142 & 0.438 & 0.311 & 314.2 & 229.4 \\
\bottomrule
\end{tabular}
\end{table}

Figure~\ref{fig:label_dists_ranus_gta} shows the class label distributions for the employed GTA5 and RANUS subsets after mapping to the eight shared semantic classes. While the overall scene composition is broadly comparable, notable distributional differences exist between the two domains. Most prominently, road pixels dominate in GTA5 (${\sim}47\%$) whereas vegetation is the largest class in RANUS (${\sim}34\%$), with sky and construction also differing substantially. The human class is effectively absent in both datasets, but particularly so in RANUS, where it constitutes less than $0.04\%$ of all labeled pixels. This extreme imbalance should be taken into account when interpreting per-class segmentation metrics. The overall distributional shift, quantified by a $\chi^2$ distance of $0.307$, confirms that the two datasets present meaningfully different scene compositions despite sharing the same semantic label space.

Table~\ref{tab:similarity_metrics} reports within-domain similarity metrics between paired RGB and generated NIR images. Across most metrics, the RANUS RGB$\leftrightarrow$NIR pair exhibits greater similarity than the GTA5 pair, with lower FID, higher SSIM, and lower perceptual distances under both LPIPS and DISTS. This is consistent with expectations: the NIR LoRA was trained exclusively on real RANUS imagery, and the diffusion model is therefore likely to introduce larger appearance changes when translating GTA5 scenes, where the domain gap between source and target is greater.

\begin{table}[ht]
\centering
\caption{Within-domain image similarity metrics. All pairwise metrics report mean $\pm$ std. Arrows indicate orientation: $\uparrow$ higher is better, $\downarrow$ lower is better.}
\label{tab:similarity_metrics}
\renewcommand{\arraystretch}{1.3}
\begin{tabular}{ l | c c c c c c c}
\toprule
\textbf{Comparison}  & \textbf{PSNR\,$\uparrow$} & \textbf{SSIM\,$\uparrow$} & \textbf{RMSE\,$\downarrow$} & \textbf{SAM\,$\downarrow$} & \textbf{LPIPS\,$\downarrow$} & \textbf{DISTS\,$\downarrow$} & \textbf{FID\,$\downarrow$}\\
\midrule
GTA5 RGB $\leftrightarrow$ NIR  & 15.15 \scriptsize{\textpm 2.11} & 0.45 \scriptsize{\textpm 0.07} & 45.72 \scriptsize{\textpm 10.82} & 0.07 \scriptsize{\textpm 0.02} & 0.38 \scriptsize{\textpm 0.05} & 0.27 \scriptsize{\textpm 0.03} & 69.0\\
RANUS RGB $\leftrightarrow$ NIR  & 14.93 \scriptsize{\textpm 3.91} & 0.62 \scriptsize{\textpm 0.14} & 50.15 \scriptsize{\textpm 21.75} & 0.08 \scriptsize{\textpm 0.04} & 0.29 \scriptsize{\textpm 0.09} & 0.22 \scriptsize{\textpm 0.04} & 49.8 \\
\bottomrule
\end{tabular}
\end{table}

The cross-domain FID scores in Table~\ref{tab:fid_ranus_gta} reveal that the GTA5 NIR$\leftrightarrow$RANUS NIR distance ($105.9$) is notably higher than the corresponding RGB$\leftrightarrow$RGB distance ($82.2$). This result should be interpreted with care, as FID relies on features extracted from an ImageNet-pretrained Inception network, which may not transfer reliably to effectively single-channel NIR images. Beyond this measurement artifact, the pixel statistics in Table~\ref{tab:image_statistics} show that the synthetic GTA5 NIR images are on average darker than their real RANUS counterparts. Two factors likely contribute to this observation. First, the TSA may underestimate the NIR-spectrum luminance, failing to fully capture the high reflectance characteristic of real NIR captures. Second, and arguably more importantly, the class distribution differences discussed above play a significant role: vegetation, which dominates RANUS, exhibits strong NIR reflectance and therefore appears bright in real NIR imagery, whereas road surfaces, which are prevalent in GTA5, are typically low-reflectance and thus appear dark. Disentangling the relative contribution of these two effects is non-trivial, and both should be considered when interpreting the cross-domain FID scores.

\begin{table}[ht]
\centering
\caption{Cross-dataset Frechet Inception Distance (FID).}
\label{tab:fid_ranus_gta}
\renewcommand{\arraystretch}{1.3}
\begin{tabular}{l | c}
\toprule
\textbf{Comparison} & \textbf{FID\,$\downarrow$} \\
\midrule
GTA5 RGB $\leftrightarrow$ RANUS RGB & 82.2 \\
GTA5 NIR $\leftrightarrow$ RANUS NIR & 105.9 \\
\bottomrule
\end{tabular}
\end{table}

\begin{table}[ht]
\centering
\caption{Statistics of pixel values in tested datasets.
  R, G, B are the mean $\pm$ standard deviation of per-pixel channel intensities in [0\,--\,255].
  Saturation is the fraction of pixels whose maximum channel value
  exceeds 250, indicating highlight clipping.
  We can observe that the synthetic NIR images are darker than the real NIR footage on average.}
\label{tab:image_statistics}
\renewcommand{\arraystretch}{1.3}
\begin{tabular}{l | l | c | c c c c}
\toprule
\textbf{Dataset} & \textbf{Mod.} & \textbf{$N$} &
\textbf{R} & \textbf{G} & \textbf{B} & \textbf{Sat.\,[\%]} \\
\midrule
GTA5  & RGB & 6\,766 & $129.11 \pm 61.27$ & $129.62 \pm 60.93$ & $128.02 \pm 61.43$ & $3.25 \pm 4.85$ \\
GTA5  & NIR & 6\,766 & $101.00 \pm 58.20$ & $101.31 \pm 58.17$ & $101.52 \pm 58.19$ & $1.23 \pm 3.06$ \\
\midrule
RANUS & RGB & 3\,772 & $125.68 \pm 71.99$ & $113.38 \pm 72.27$ & $123.43 \pm 74.63$ & $23.15 \pm 15.99$ \\
RANUS & NIR & 3\,772 & $113.32 \pm 56.86$ & $113.32 \pm 56.86$ & $113.32 \pm 56.86$ & $9.53 \pm 13.30$ \\
\bottomrule
\end{tabular}
\end{table}

\begin{table}[ht]
\centering
\setlength{\tabcolsep}{0.5pt}
\caption{Source drift analysis: stylized GTA5-NIR variants vs.\ original GTA5-RGB data. All pairwise metrics are image-level means; FID is computed on full distributions. \emph{FID}$\rightarrow$\emph{Synth}: distance to GTA5-RGB base. \emph{FID}$\rightarrow$\emph{Real}: distance to real RANUS-NIR. Arrows indicate preferred direction.}
\label{tab:synthetic_stylization_similarity_sorted_full_ranus}
\begin{tabular}{l c c c c c c cc}
\toprule
\textbf{Method} & \textbf{PSNR\,$\uparrow$} & \textbf{SSIM\,$\uparrow$} & \textbf{RMSE\,$\downarrow$} & \textbf{SAM\,$\downarrow$} & \textbf{LPIPS\,$\downarrow$} & \textbf{DISTS\,$\downarrow$} & \textbf{FID$\rightarrow$Synth\,$\downarrow$} & \textbf{FID$\rightarrow$Real\,$\downarrow$} \\
\midrule
TSA Only & 15.2 & 0.446 & 45.7 & 0.065 & 0.384 & 0.266 & 69.0 & 105.9 \\
VSD(1,1.0) & 16.6 & 0.576 & 41.9 & 0.068 & 0.273 & 0.249 & 88.6 & 113.5 \\
VSD(4,0.25) + TSA & 14.1 & 0.395 & 51.8 & 0.070 & 0.413 & 0.270 & 77.2 & 104.2 \\
VSD(4,0.75) + TSA & 12.4 & 0.294 & 63.1 & 0.079 & 0.465 & 0.291 & 121.7 & 132.4 \\
VSD(8,0.25) + TSA & 14.0 & 0.392 & 52.2 & 0.069 & 0.421 & 0.268 & 86.2 & 113.3 \\
VSD(8,0.75) + TSA & 12.2 & 0.288 & 63.8 & 0.079 & 0.476 & 0.291 & 139.8 & 149.0 \\
VSD(16,0.25) + TSA & 13.9 & 0.389 & 52.3 & 0.069 & 0.430 & 0.268 & 101.0 & 128.7 \\
VSD(16,0.75) + TSA & 12.0 & 0.281 & 64.6 & 0.079 & 0.493 & 0.296 & 168.9 & 176.6 \\
\bottomrule
\end{tabular}
\end{table}

\subsection{Semantic Segmentation Training}
\label{subsec:training}
    
All semantic segmentation models share identical data preprocessing, augmentation, and training settings to ensure comparability across architectures. We use a crop size of $1024 \times 1024$ for both training and evaluation. The training pipeline includes random scaling within a ratio range of $0.5$ to $2.0$, random horizontal flipping, and photometric distortion. During inference, images are resized to the same $1024 \times 1024$ resolution without further augmentation. 
All remaining models use the identical optimizer setup (AdamW) and follow the same 60k‑iteration training schedule provided by MMSegmentation\footnote{https://github.com/open-mmlab/mmsegmentation}. 
We intentionally select models that represent distinct architectural families to study how different design choices influence model biases, and we leverage pretrained resources from the same source to achieve competitive performance while significantly reducing training time.
We employ \textit{DeepLabV3+} \cite{chen2018encoder} as a classical and widely adopted CNN‑based baseline, featuring an encoder–decoder design with atrous convolutions and a lightweight decoder. For our experiments, we use the ResNet‑50 backbone with ImageNet‑1k pretrained weights.
We utilize a \textit{SegFormer} \cite{xie2021segformer} (MiT‑B5) with ImageNet‑1k pretrained encoder weights. As a lightweight, purely Transformer-based architecture with a minimal decoder, it represents modern efficient segmentation models. Its inclusion enables us to study how compact Transformer encoders differ in bias and robustness from CNNs and heavier Transformers.
\textit{Mask2Former} \cite{cheng2022masked} with a Swin‑Large backbone (pretrained on the full ImageNet‑22k dataset, which is approximately 10 times larger than the ImageNet-1k subset) serves as a high‑capacity, state‑of‑the‑art Transformer model. Its mask‑attention design and large backbone allow us to observe how model scale and broader global attention affect segmentation behavior.


\FloatBarrier
\section{More Numerical Results}\label{app:results}
\subsection{Spectral Dependence Analysis}

\begin{table}[h]
    \centering
    \caption{Comparison of mIoU in \% achieved on target domain data in RGB and NIR for models trained on synthetic RGB data. The DeepLabV3+ and SegFormer models exhibit a noticeably higher gap than Mask2Former. Reports first checkpoint for each.}
    \label{tab:spectral_dependence}
    \begin{tabular}{l  c c c c}
        \toprule
        Model & Train Set & RANUS RGB & RANUS NIR & Gap \\
        \midrule
        DeepLabV3+ & GTA5 RGB & 37.75 & 21.98 & 15.87 \\
        SegFormer & GTA5 RGB & 40.89 & 29.50 & 11.39 \\
        Mask2Former & GTA5 RGB & 61.68 & 59.41 & 2.27 \\
        \bottomrule
    \end{tabular}

\end{table}
\FloatBarrier

\subsection{Detailed Noise Description}
In the following figures the detailed noise is shown:
\begin{figure}[h]
  \centering
  \setlength{\tabcolsep}{2pt}
  \renewcommand{\arraystretch}{0.3}
  \resizebox{\textwidth}{!}{%
    \begin{tabular}{rcccccccc}
      & \small{Original} & \small{$\sigma$=0.03} & \small{$\sigma$=0.05} & \small{$\sigma$=0.1} & \small{$\sigma$=0.2} & \small{$\sigma$=0.35} & \small{$\sigma$=0.6} & \small{$\sigma$=0.9} \\[4pt]
      \raisebox{1.2\height}{\rotatebox[origin=c]{90}{\small{Image}}} &
      \includegraphics[width=0.1250\linewidth]{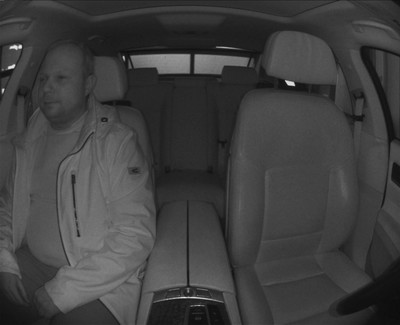} &
      \includegraphics[width=0.1250\linewidth]{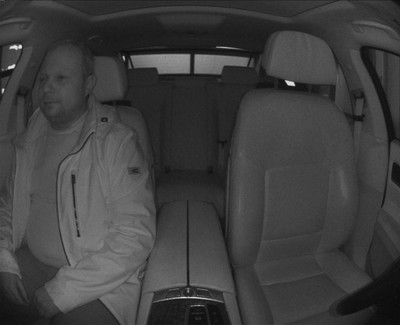} &
      \includegraphics[width=0.1250\linewidth]{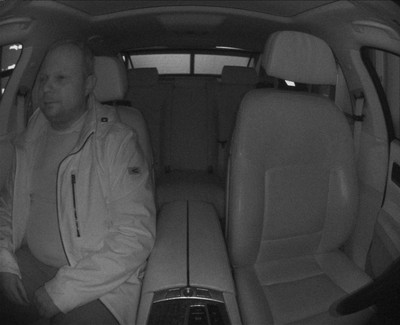} &
      \includegraphics[width=0.1250\linewidth]{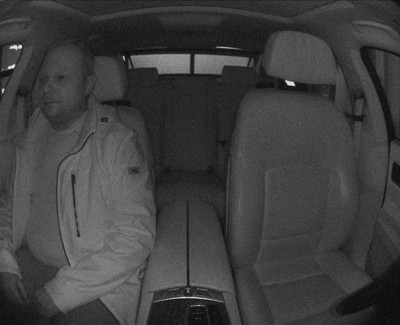} &
      \includegraphics[width=0.1250\linewidth]{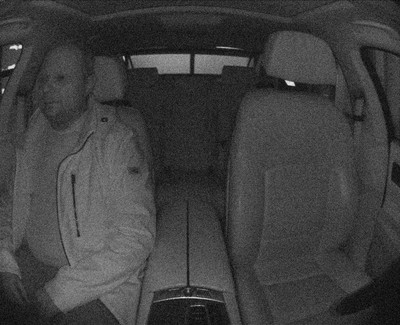} &
      \includegraphics[width=0.1250\linewidth]{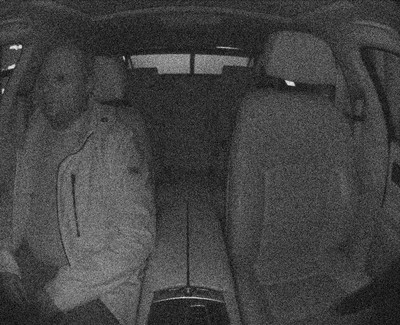} &
      \includegraphics[width=0.1250\linewidth]{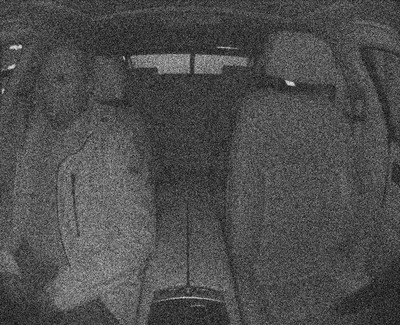} &
      \includegraphics[width=0.1250\linewidth]{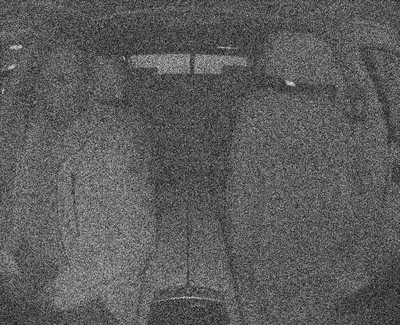} \\
    \end{tabular}%
  }
  \caption{Interior Uniform Noise}
  \label{fig:uniform-noise_interior}
\end{figure}

\begin{figure}[h]
  \centering
  \setlength{\tabcolsep}{2pt}
  \renewcommand{\arraystretch}{0.3}
  \resizebox{\textwidth}{!}{%
    \begin{tabular}{lccccc}
      & \small{Original} & \small{$\sigma$=1} & \small{$\sigma$=5} & \small{$\sigma$=10} & \small{$\sigma$=15} \\[4pt]
      \raisebox{2.2\height}{\rotatebox[origin=c]{90}{\small{Image}}} &
      \includegraphics[width=0.2000\linewidth]{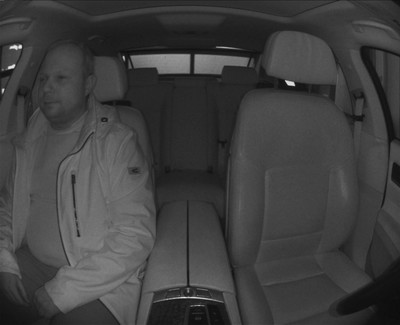} &
      \includegraphics[width=0.2000\linewidth]{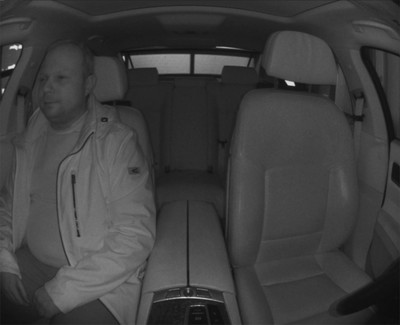} &
      \includegraphics[width=0.2000\linewidth]{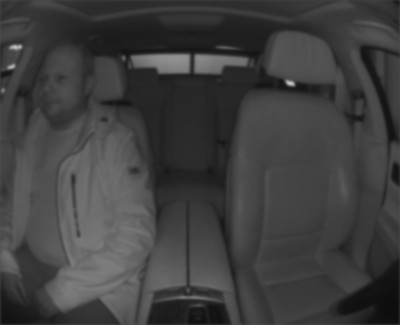} &
      \includegraphics[width=0.2000\linewidth]{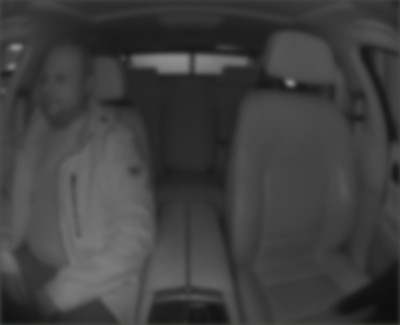} &
      \includegraphics[width=0.2000\linewidth]{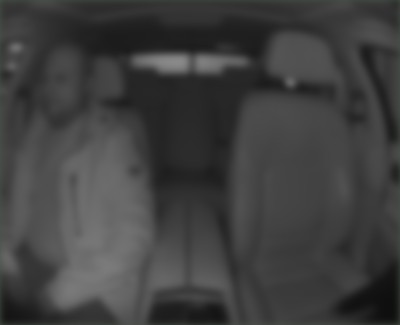} \\
    \end{tabular}%
  }
  \caption{Interior Low-Pass Filtering}
  \label{fig:low-pass_interior}
\end{figure}

\begin{figure}[h]
  \centering
  \setlength{\tabcolsep}{2pt}
  \renewcommand{\arraystretch}{0.3}
  \resizebox{\textwidth}{!}{%
    \begin{tabular}{lcccccc}
      & \small{Original} & \small{s=0.05} & \small{s=0.1} & \small{s=0.5} & \small{s=1.0} & \small{s=2.0} \\[4pt]
      \raisebox{1.8\height}{\rotatebox[origin=c]{90}{\small{Image}}} &
      \includegraphics[width=0.1667\linewidth]{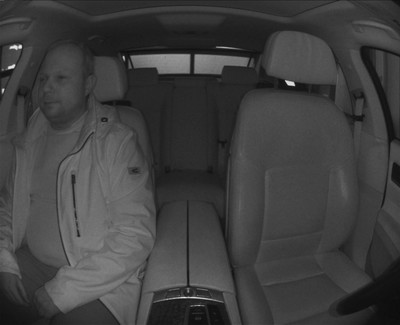} &
      \includegraphics[width=0.1667\linewidth]{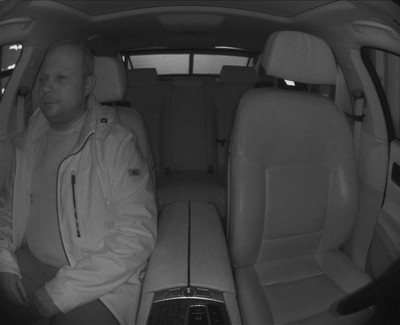} &
      \includegraphics[width=0.1667\linewidth]{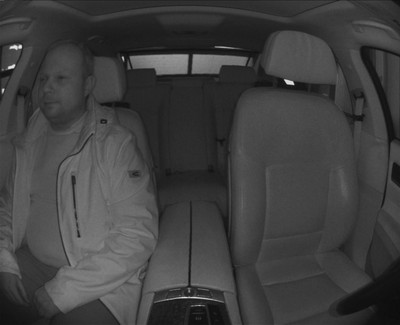} &
      \includegraphics[width=0.1667\linewidth]{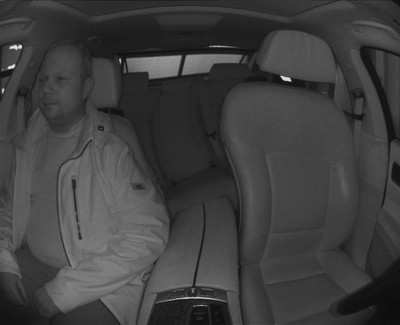} &
      \includegraphics[width=0.1667\linewidth]{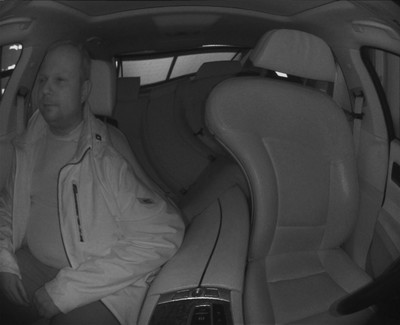} &
      \includegraphics[width=0.1667\linewidth]{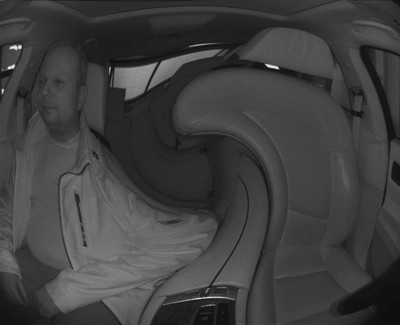} \\[4pt]
      \raisebox{0.8\height}{\rotatebox[origin=c]{90}{\small{Ground Truth}}} &
      \includegraphics[width=0.1667\linewidth]{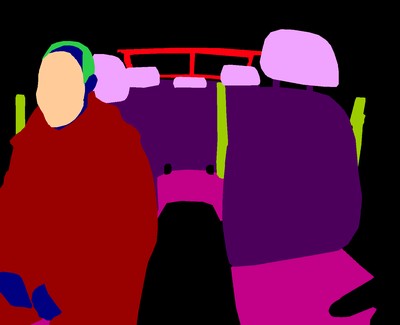} &
      \includegraphics[width=0.1667\linewidth]{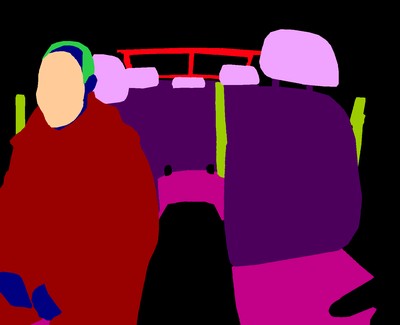} &
      \includegraphics[width=0.1667\linewidth]{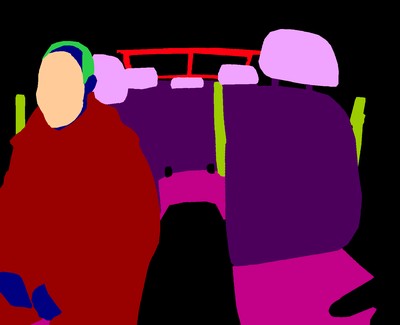} &
      \includegraphics[width=0.1667\linewidth]{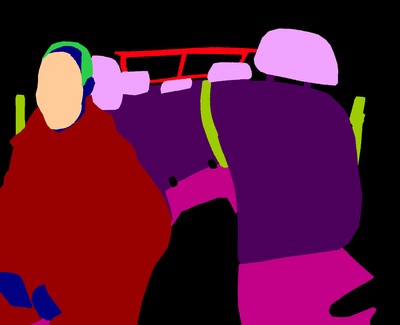} &
      \includegraphics[width=0.1667\linewidth]{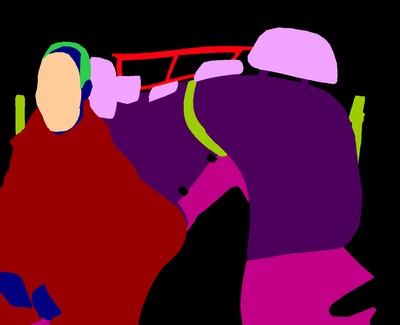} &
      \includegraphics[width=0.1667\linewidth]{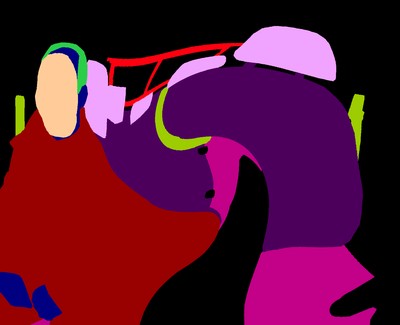} \\
    \end{tabular}%
  }
  \caption{Interior Swirl Distortion}
  \label{fig:swirl_interior}
\end{figure}

\begin{figure}[h]
  \centering
  \setlength{\tabcolsep}{2pt}
  \renewcommand{\arraystretch}{0.3}
  \resizebox{\textwidth}{!}{%
    \begin{tabular}{lcccccc}
      & \small{Original} & \small{$\alpha$=10} & \small{$\alpha$=20} & \small{$\alpha$=50} & \small{$\alpha$=100} & \small{$\alpha$=150} \\[4pt]
      \raisebox{1.8\height}{\rotatebox[origin=c]{90}{\small{Image}}} &
      \includegraphics[width=0.1667\linewidth]{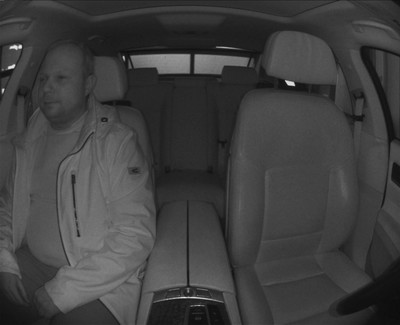} &
      \includegraphics[width=0.1667\linewidth]{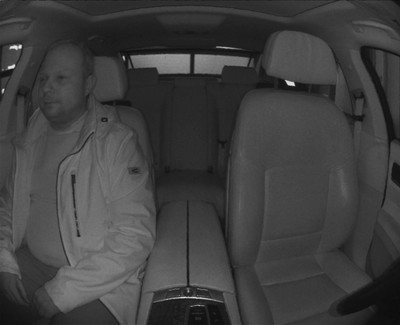} &
      \includegraphics[width=0.1667\linewidth]{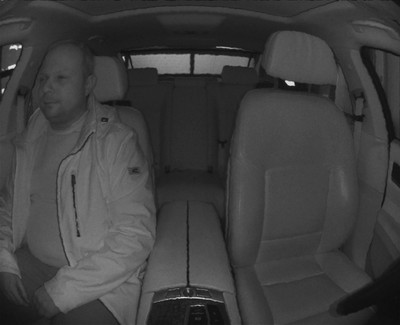} &
      \includegraphics[width=0.1667\linewidth]{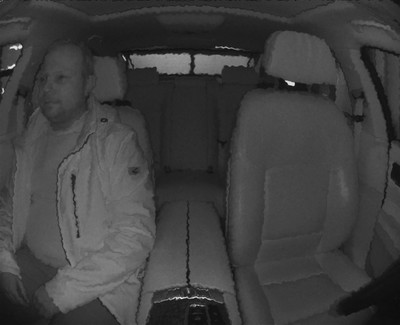} &
      \includegraphics[width=0.1667\linewidth]{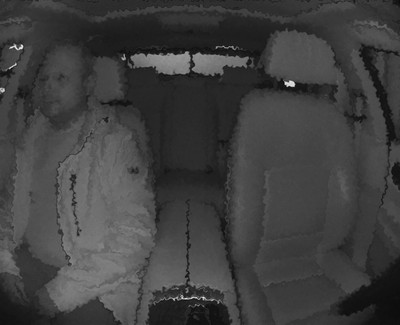} &
      \includegraphics[width=0.1667\linewidth]{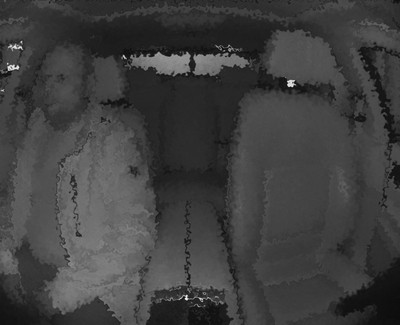} \\[4pt]
      \raisebox{0.8\height}{\rotatebox[origin=c]{90}{\small{Ground Truth}}} &
      \includegraphics[width=0.1667\linewidth]{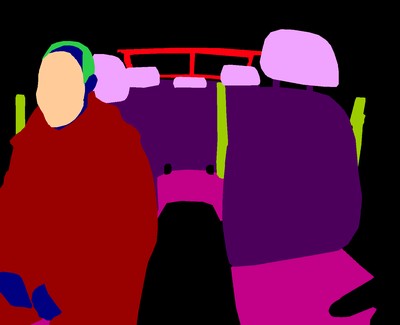} &
      \includegraphics[width=0.1667\linewidth]{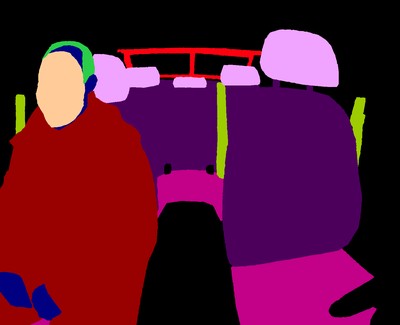} &
      \includegraphics[width=0.1667\linewidth]{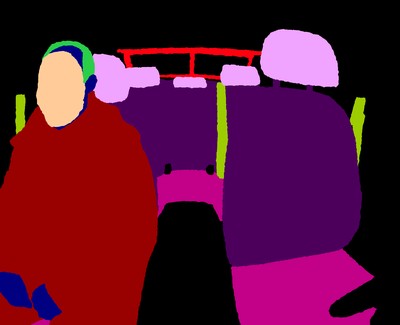} &
      \includegraphics[width=0.1667\linewidth]{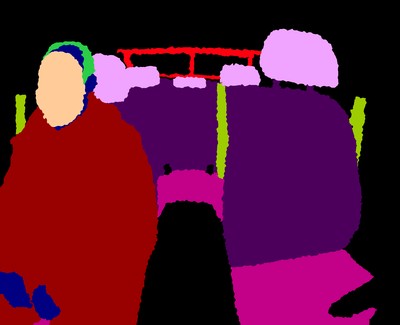} &
      \includegraphics[width=0.1667\linewidth]{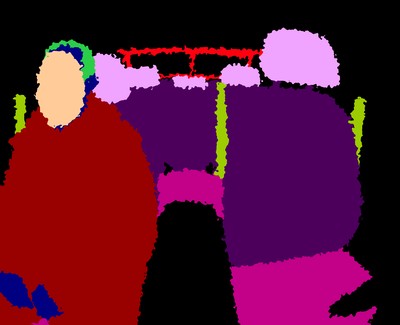} &
      \includegraphics[width=0.1667\linewidth]{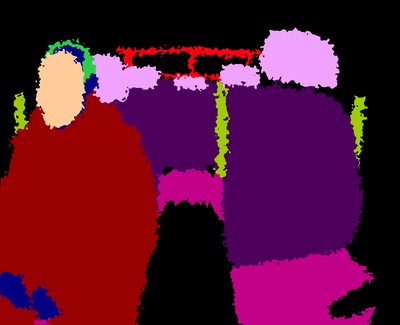} \\
    \end{tabular}%
  }
  \caption{Interior Elastic Deformation}
  \label{fig:elastic_interior}
\end{figure}

\begin{figure}[h]
  \centering
  \setlength{\tabcolsep}{1pt}
  \renewcommand{\arraystretch}{0.3}
    \resizebox{\textwidth}{!}{%

  \begin{tabular}{lcccccc}
     & \small{Original} & \small{$\sigma$=1} & \small{$\sigma$=5} & \small{$\sigma$=10} & \small{$\sigma$=15} & \small{$\sigma$=20} \\[2pt]
    \rotatebox{90}{\small{Image}} &
    \includegraphics[width=0.1533\linewidth]{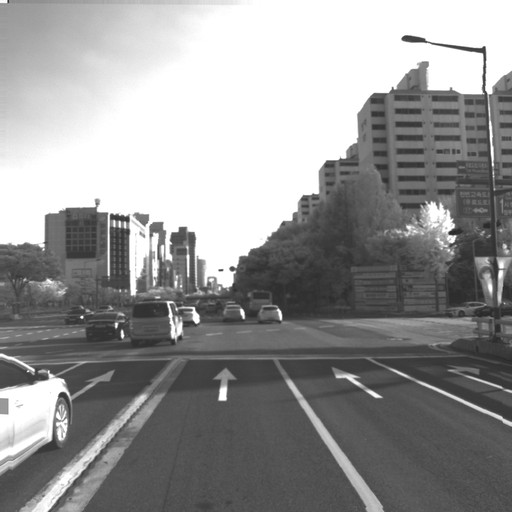} &
    \includegraphics[width=0.1533\linewidth]{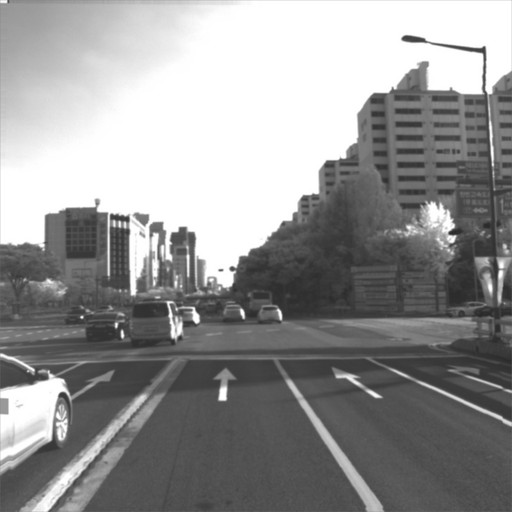} &
    \includegraphics[width=0.1533\linewidth]{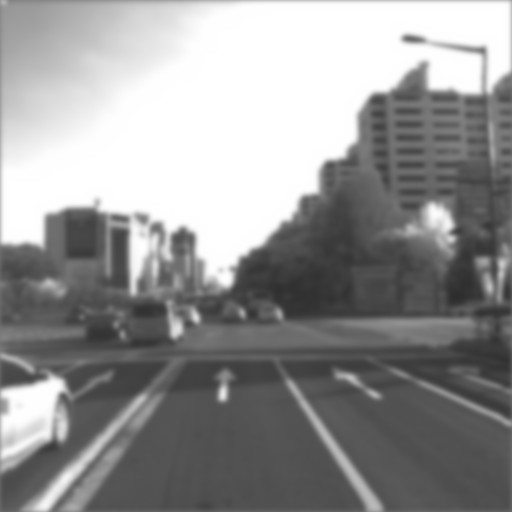} &
    \includegraphics[width=0.1533\linewidth]{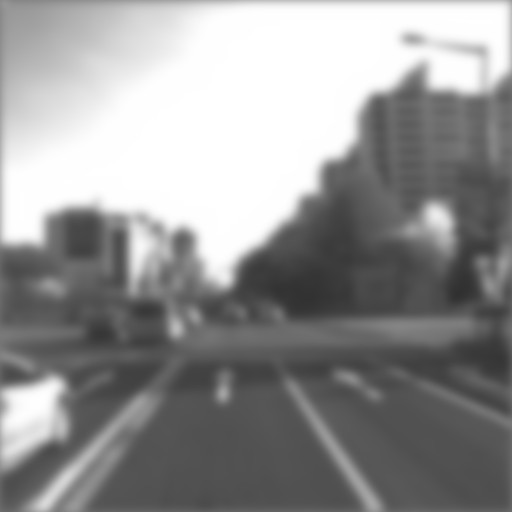} &
    \includegraphics[width=0.1533\linewidth]{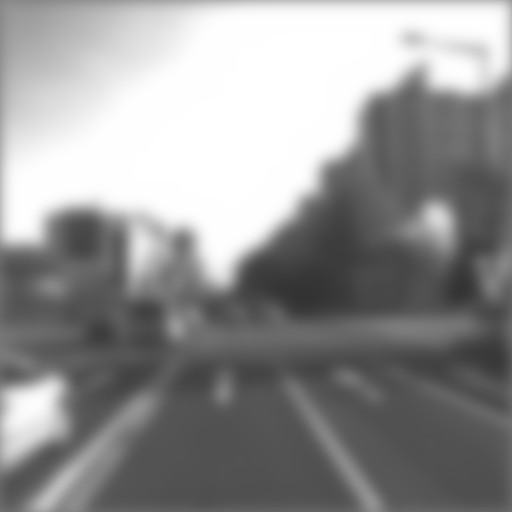} &
    \includegraphics[width=0.1533\linewidth]{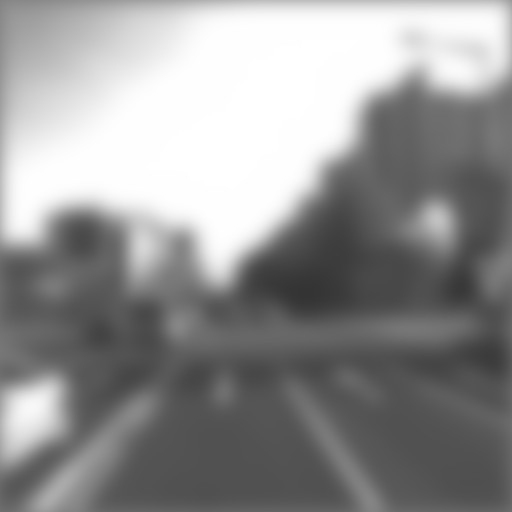} \\
  \end{tabular}
  }
  \caption{RANUS Low-pass filtering}
  \label{fig:low-pass_ranus}
\end{figure}

\begin{figure}[h]
  \centering
  \setlength{\tabcolsep}{1pt}
  \renewcommand{\arraystretch}{0.3}
    \resizebox{\textwidth}{!}{%

  \begin{tabular}{lcccccccc}
     & \small{Original} & \small{n=0.03} & \small{n=0.05} & \small{n=0.1} & \small{n=0.2} & \small{n=0.35} & \small{n=0.6} & \small{n=0.9} \\[2pt]
    \rotatebox{90}{\small{Image}} &
    \includegraphics[width=0.125\linewidth]{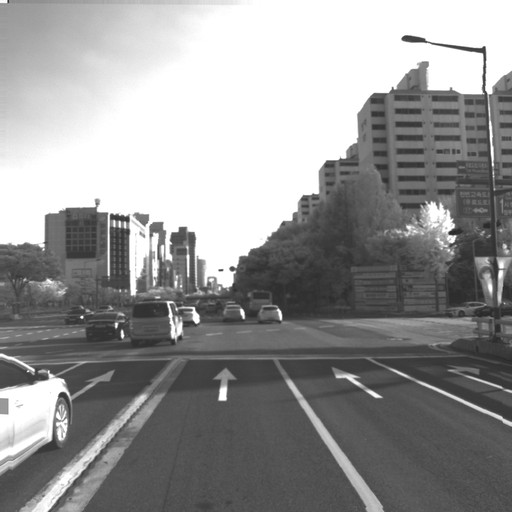} &
    \includegraphics[width=0.125\linewidth]{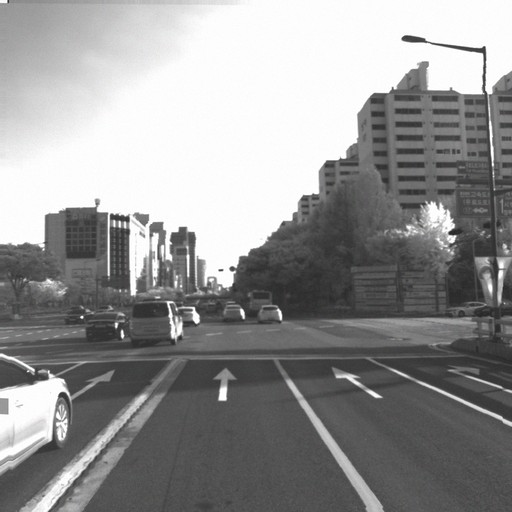} &
    \includegraphics[width=0.125\linewidth]{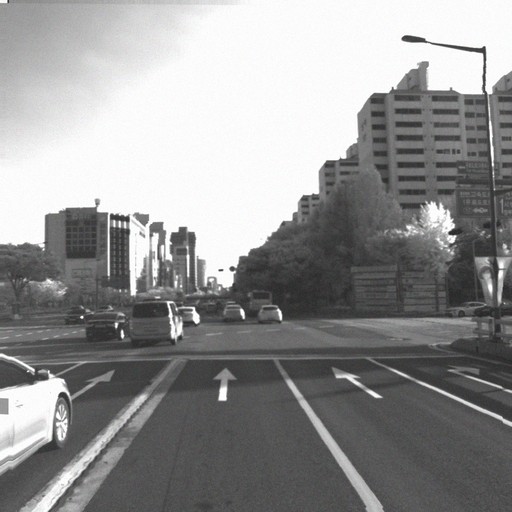} &
    \includegraphics[width=0.125\linewidth]{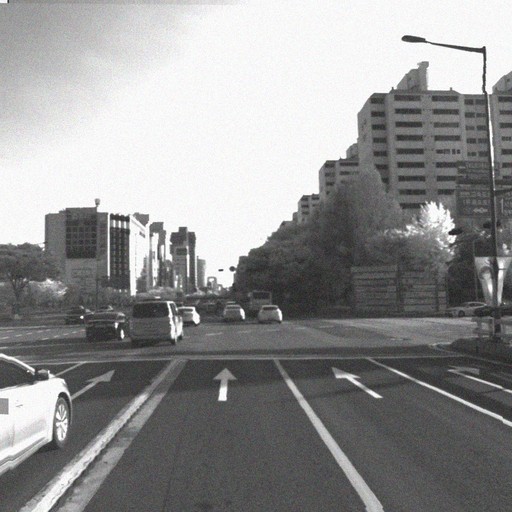} &
    \includegraphics[width=0.125\linewidth]{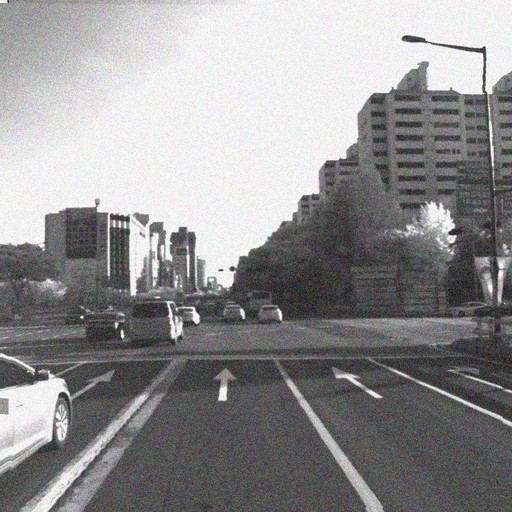} &
    \includegraphics[width=0.125\linewidth]{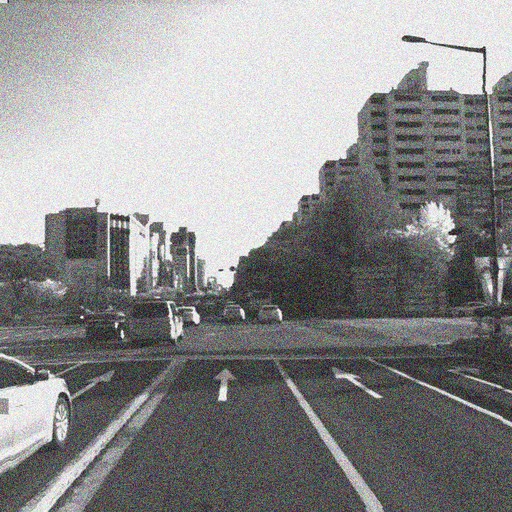} &
    \includegraphics[width=0.125\linewidth]{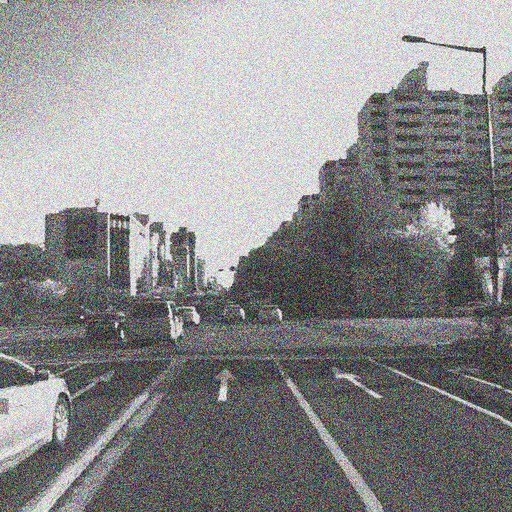} &
    \includegraphics[width=0.125\linewidth]{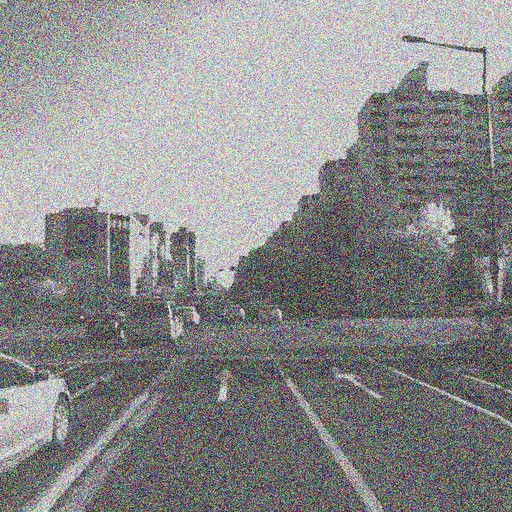} \\
  \end{tabular}
  }
  \caption{RANUS Uniform noise}
  \label{fig:uniform-noise_ranus}
\end{figure}

\begin{figure}[h]
  \centering
  \setlength{\tabcolsep}{1pt}
  \renewcommand{\arraystretch}{0.3}
      \resizebox{\textwidth}{!}{%

  \begin{tabular}{lcccccc}
     & \small{Original} & \small{s=0.05} & \small{s=0.1} & \small{s=0.5} & \small{s=1.0} & \small{s=2.0} \\[2pt]
    \rotatebox{90}{\small{Image}} &
    \includegraphics[width=0.1533\linewidth]{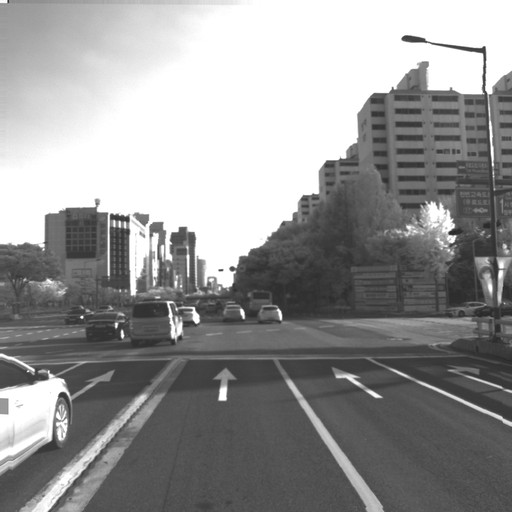} &
    \includegraphics[width=0.1533\linewidth]{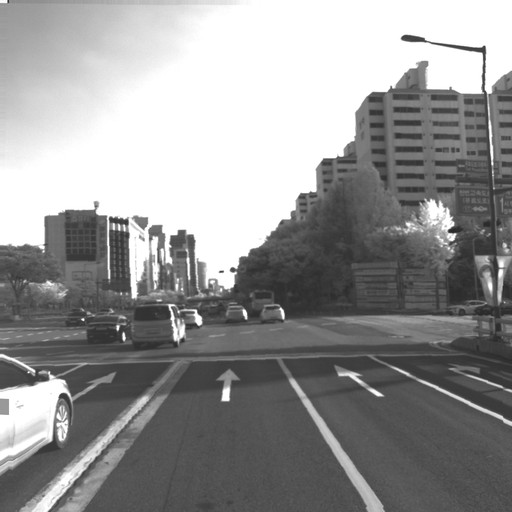} &
    \includegraphics[width=0.1533\linewidth]{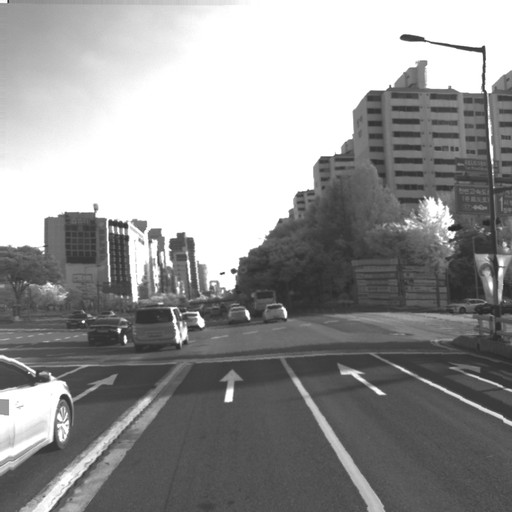} &
    \includegraphics[width=0.1533\linewidth]{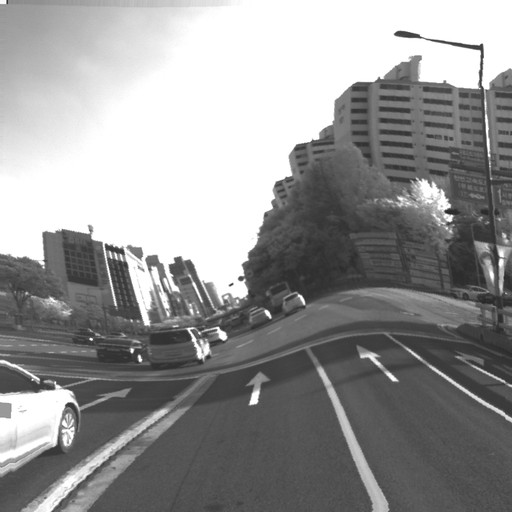} &
    \includegraphics[width=0.1533\linewidth]{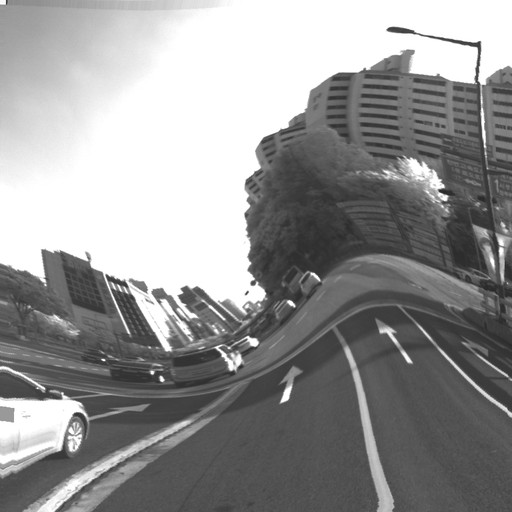} &
    \includegraphics[width=0.1533\linewidth]{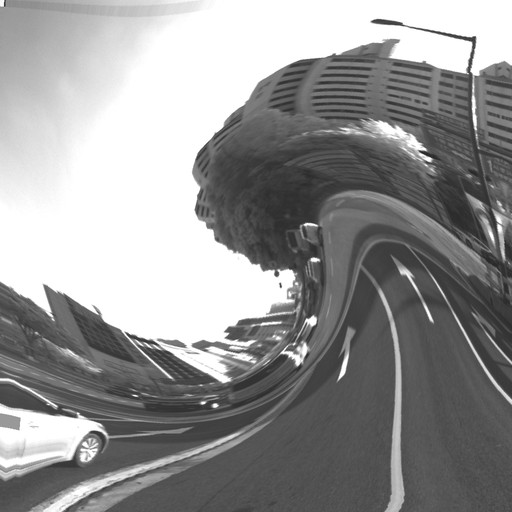} \\[1pt]
    \rotatebox{90}{\small{Ground Truth}} &
    \includegraphics[width=0.1533\linewidth]{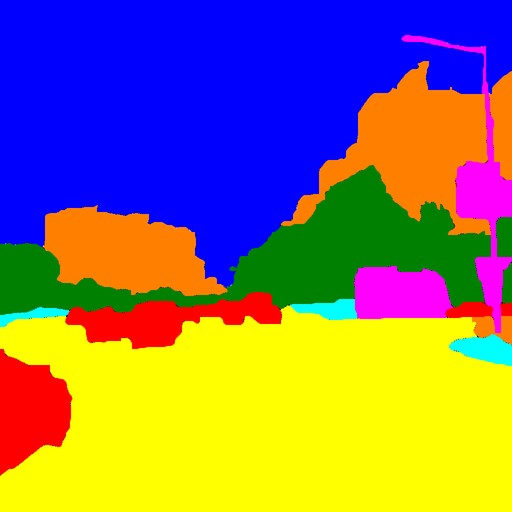} &
    \includegraphics[width=0.1533\linewidth]{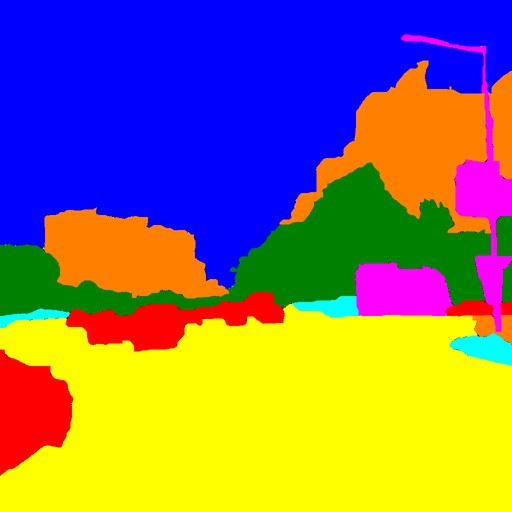} &
    \includegraphics[width=0.1533\linewidth]{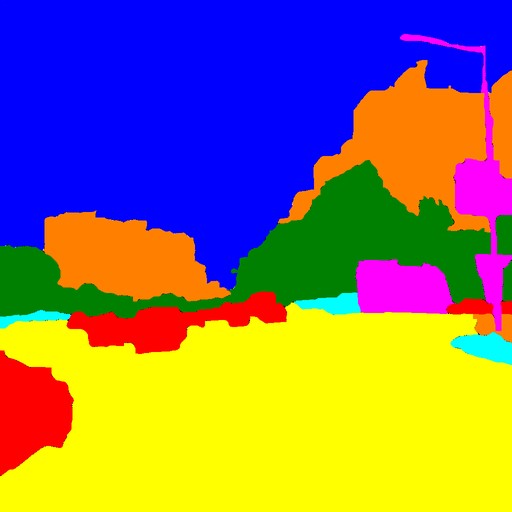} &
    \includegraphics[width=0.1533\linewidth]{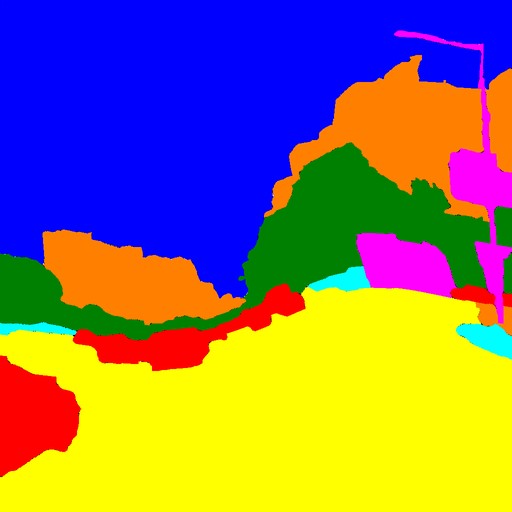} &
    \includegraphics[width=0.1533\linewidth]{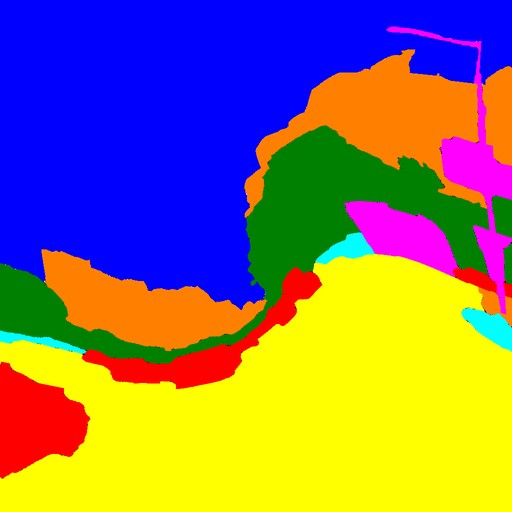} &
    \includegraphics[width=0.1533\linewidth]{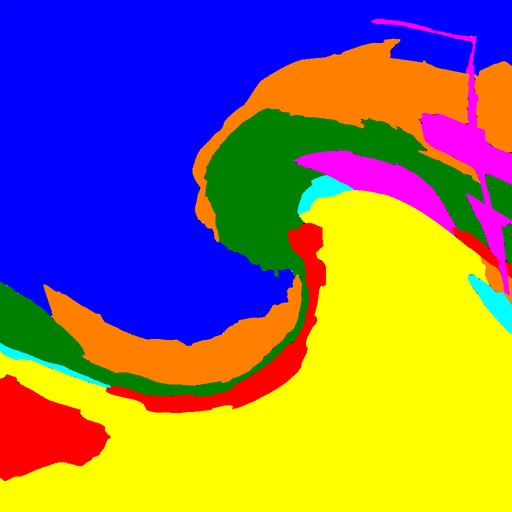} \\
  \end{tabular}
  }
  \caption{RANUS Swirl distortion}
  \label{fig:swirl_ranus}
\end{figure}

\begin{figure}[h]
  \centering
  \setlength{\tabcolsep}{1pt}
  \renewcommand{\arraystretch}{0.3}
    \resizebox{\textwidth}{!}{%

  \begin{tabular}{lcccccc}
     & \small{Original} & \small{$\alpha$=10} & \small{$\alpha$=20} & \small{$\alpha$=50} & \small{$\alpha$=100} & \small{$\alpha$=150} \\[2pt]
    \rotatebox{90}{\small{Image}} &
    \includegraphics[width=0.1533\linewidth]{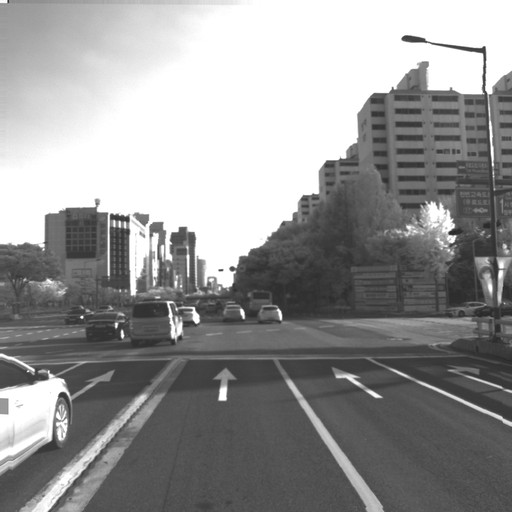} &
    \includegraphics[width=0.1533\linewidth]{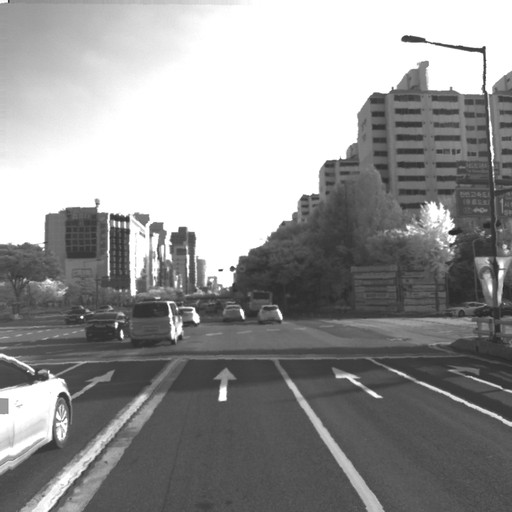} &
    \includegraphics[width=0.1533\linewidth]{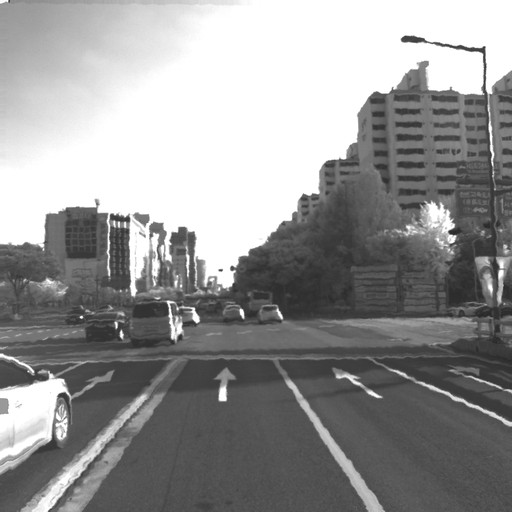} &
    \includegraphics[width=0.1533\linewidth]{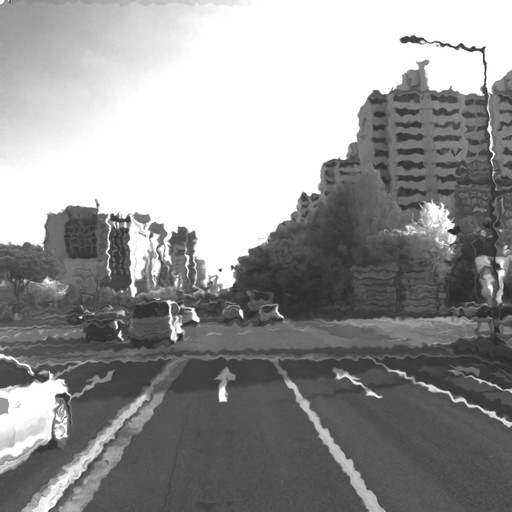} &
    \includegraphics[width=0.1533\linewidth]{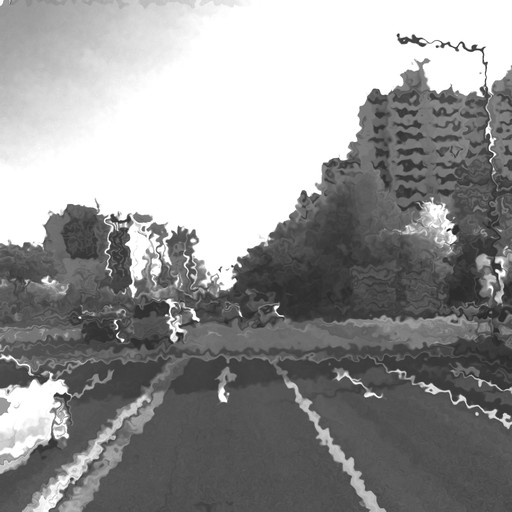} &
    \includegraphics[width=0.1533\linewidth]{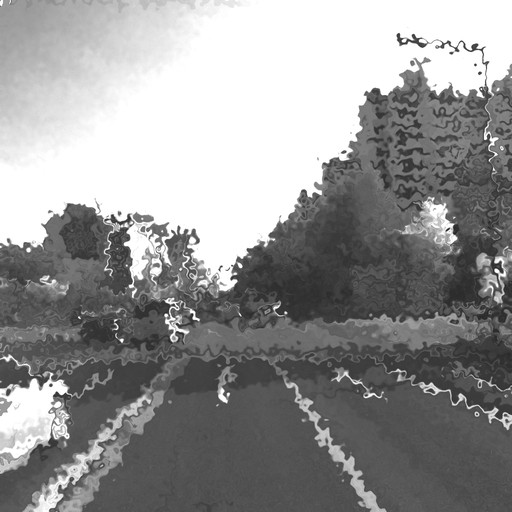} \\[1pt]
    \rotatebox{90}{\small{Ground Truth}} &
    \includegraphics[width=0.1533\linewidth]{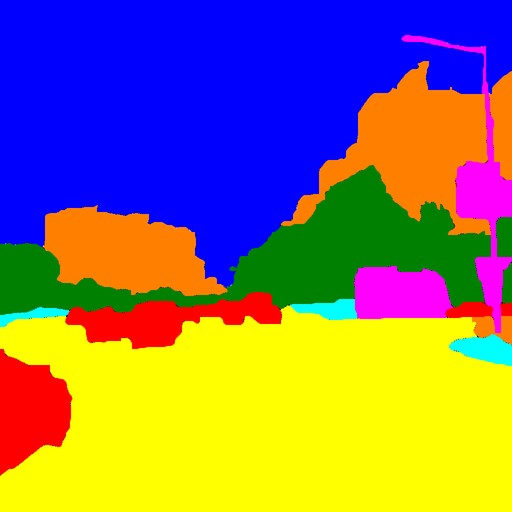} &
    \includegraphics[width=0.1533\linewidth]{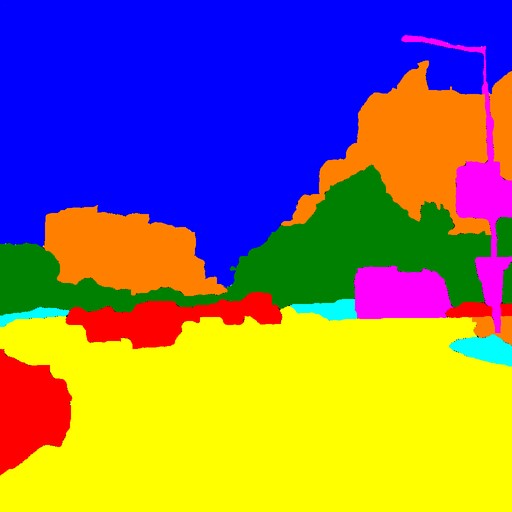} &
    \includegraphics[width=0.1533\linewidth]{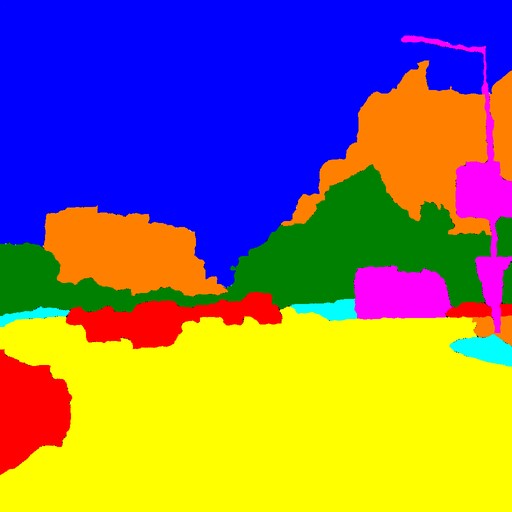} &
    \includegraphics[width=0.1533\linewidth]{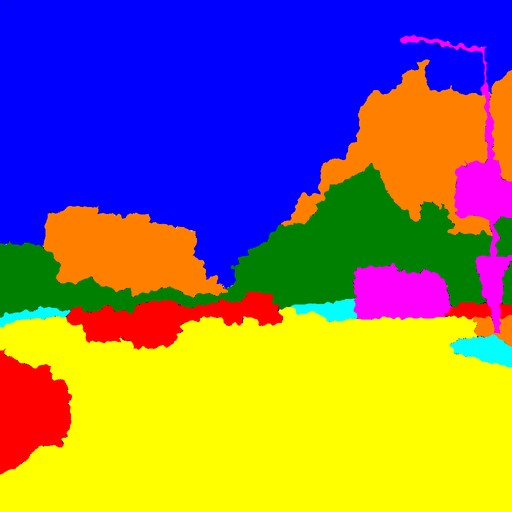} &
    \includegraphics[width=0.1533\linewidth]{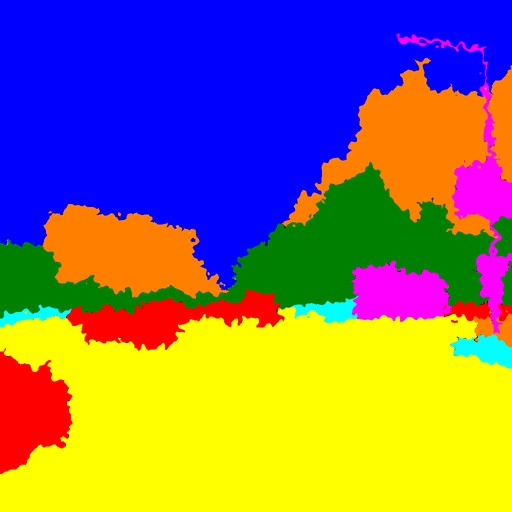} &
    \includegraphics[width=0.1533\linewidth]{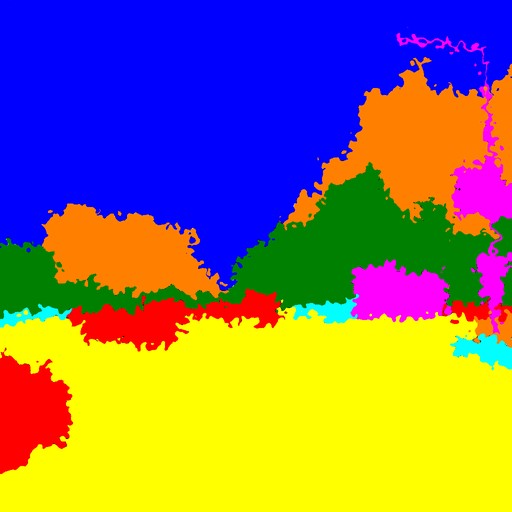} \\
  \end{tabular}
  }
  \caption{RANUS Elastic deformation}
  \label{fig:elastic_ranus}
\end{figure}

\FloatBarrier
\subsection{Qualitative Domain Improvements}

\CompareAcrossModels

\FloatBarrier

\subsection{Detailed Robustness}
\label{sec:detailed_robust}
\begin{table}[h]
\centering
\caption{Robustness summary on interior data across all models and segmentation architectures. $mPC_{AUC}$ is the area-normalized mean mIoU across distortions; Norm $mPC_{AUC}$ is normalized to the Real Baseline (100 = baseline).}
\label{tab:robustness_norm_combined}
\small
\setlength{\tabcolsep}{4pt}
\begin{tabular}{l rr rr rr}
\toprule
& \multicolumn{2}{c}{\textbf{DeepLabV3+}} 
& \multicolumn{2}{c}{\textbf{Segformer}} 
& \multicolumn{2}{c}{\textbf{Mask2Former}} \\
\cmidrule(lr){2-3} \cmidrule(lr){4-5} \cmidrule(lr){6-7}
Model 
  & $mPC_{AUC}$ & Norm (\%) 
  & $mPC_{AUC}$ & Norm (\%) 
  & $mPC_{AUC}$ & Norm (\%) \\
\midrule
Real Baseline         & 41.43 & 100.0 & 55.89 & 100.0 & 60.78 & 100.0 \\
\midrule
VSD(1,1.0)         & 13.22 &  29.0 & 38.02 &  69.2 & 45.83 &  76.0 \\
Synthetic Baseline    & 11.35 &  25.4 & 25.06 &  44.5 & 34.49 &  57.4 \\
TSA Only             & 14.51 &  31.4 & 28.38 &  50.0 & 34.85 &  56.7 \\
\midrule
VSD(4,0.25)            & 16.34 &  36.5 & 35.36 &  63.9 & 42.25 &  70.5 \\
VSD(4,0.25) [Gray]     & 17.51 &  38.8 & 35.03 &  63.4 & 42.36 &  70.7 \\
VSD(4,0.25)+ TSA       & 20.25 &  47.7 & 35.48 &  63.5 & 41.51 &  68.9 \\
VSD(4,0.25)+ TSA [Gray]& 21.70 &  52.8 & 36.97 &  66.8 & 42.75 &  71.1 \\
VSD(4,0.75)            & 18.46 &  41.2 & 35.99 &  65.2 & 43.40 &  72.2 \\
VSD(4,0.75) [Gray]     & 18.60 &  43.3 & 36.01 &  65.4 & 43.35 &  72.3 \\
VSD(4,0.75)+ TSA       & 21.15 &  51.8 & 35.81 &  64.7 & 44.72 &  74.7 \\
VSD(4,0.75)+ TSA [Gray]& 21.59 &  53.2 & 35.35 &  64.1 & 43.79 &  73.0 \\
\midrule
VSD(8,0.25)            & 16.00 &  35.7 & 34.86 &  62.7 & 41.71 &  69.2 \\
VSD(8,0.25) [Gray]     & 17.60 &  39.5 & 34.40 &  62.3 & 41.43 &  68.9 \\
VSD(8,0.25)+ TSA       & 20.39 &  49.4 & 35.21 &  63.1 & 42.59 &  70.7 \\
VSD(8,0.25)+ TSA [Gray]& 21.32 &  53.1 & 35.45 &  63.8 & 41.81 &  69.4 \\
VSD(8,0.75)            & 19.01 &  46.4 & 35.30 &  63.7 & 43.09 &  71.5 \\
VSD(8,0.75) [Gray]     & 18.43 &  42.9 & 35.46 &  64.2 & 42.25 &  70.2 \\
VSD(8,0.75)+ TSA       & 21.45 &  53.5 & 35.94 &  65.0 & 43.91 &  73.0 \\
VSD(8,0.75)+ TSA [Gray]& 22.19 &  57.4 & 35.56 &  64.3 & 43.56 &  72.7 \\
\midrule
VSD(16,0.25)           & 15.07 &  35.1 & 33.13 &  59.6 & 39.75 &  66.0 \\
VSD(16,0.25) [Gray]    & 17.47 &  39.6 & 33.22 &  59.6 & 39.26 &  65.4 \\
VSD(16,0.25)+ TSA      & 20.98 &  52.3 & 34.83 &  62.3 & 40.95 &  67.9 \\
VSD(16,0.25)+ TSA [Gray]& 21.43 & 53.3 & 35.56 &  63.8 & 40.91 &  67.7 \\
VSD(16,0.75)           & 18.18 &  43.1 & 34.55 &  62.2 & 41.67 &  69.3 \\
VSD(16,0.75) [Gray]    & 18.83 &  43.1 & 34.62 &  62.5 & 42.00 &  69.7 \\
VSD(16,0.75)+ TSA      & 20.69 &  52.8 & 33.69 &  60.4 & 43.10 &  71.7 \\
VSD(16,0.75)+ TSA [Gray]& 22.94 & 58.8 & 33.86 &  61.0 & 42.11 &  70.1 \\
\bottomrule
\end{tabular}
\end{table}

\begin{table}[h]
\centering
\caption{Robustness summary on exterior data across all models and segmentation architectures. $mPC_{AUC}$ is the area-normalized mean mIoU across distortions; Norm $mPC_{AUC}$ is normalized to the Real Baseline (100 = baseline). $VSD(n,p)$ without TSA refers to
Voronoi style diversification with RGB source and style images.}
\label{tab:robustness_summary_exterior_all}
\small
\setlength{\tabcolsep}{5pt}
\renewcommand{\arraystretch}{1.1}
\begin{tabular}{l cc cc cc}
\toprule
\textbf{Method} & \multicolumn{2}{c}{\textbf{DeepLabV3+}} & \multicolumn{2}{c}{\textbf{SegFormer}} & \multicolumn{2}{c}{\textbf{Mask2Former}} \\
\cmidrule(lr){2-3}
\cmidrule(lr){4-5}
\cmidrule(lr){6-7}
 & mPC-AUC & Norm (\%) & mPC-AUC & Norm (\%) & mPC-AUC & Norm (\%) \\
\midrule
Real Baseline & 36.96 & 100.0 & 38.88 & 100.0 & 53.28 & 100.0 \\
Synthetic Baseline & 10.58 & 31.0 & 14.34 & 35.0 & 36.01 & 68.0 \\
\midrule
TSA Only & 18.96 & 52.5 & 20.28 & 51.5 & 34.21 & 64.2 \\
VSD(1,1.0) & 15.95 & 43.7 & 20.88 & 61.9 & 44.32 & 84.0 \\
VSD(4,0.25) + TSA & 22.94 & 64.1 & 23.20 & 66.1 & 39.83 & 75.2 \\
VSD(4,0.75) + TSA & 23.44 & 66.3 & 25.38 & 72.9 & 41.97 & 79.4 \\
VSD(8,0.25) + TSA & 24.03 & 68.1 & 23.77 & 66.4 & 40.99 & 77.4 \\
VSD(8,0.75) + TSA & 20.63 & 59.6 & 25.30 & 71.5 & 42.44 & 80.3 \\
VSD(16,0.25) + TSA & 22.07 & 62.1 & 23.98 & 65.8 & 40.10 & 75.9 \\
VSD(16,0.75) + TSA & 19.30 & 52.1 & 24.43 & 68.1 & 39.57 & 74.9 \\
\midrule
VSD(1,1.0) & 15.07 & 41.5 & 23.02 & 68.1 & 42.15 & 80.3 \\
VSD(4,0.25) & 16.64 & 49.6 & 19.28 & 55.7 & 40.49 & 77.0 \\
VSD(4,0.75) & 19.96 & 58.8 & 21.41 & 59.6 & 39.60 & 75.5 \\
VSD(8,0.25) & 14.76 & 44.3 & 18.71 & 51.1 & 38.31 & 72.3 \\
VSD(8,0.75) & 16.84 & 46.6 & 21.73 & 60.9 & 39.43 & 74.9 \\
VSD(16,0.25) & 14.07 & 41.2 & 18.35 & 49.3 & 38.40 & 72.8 \\
VSD(16,0.75) & 18.09 & 51.6 & 20.93 & 57.8 & 38.78 & 73.7 \\
\bottomrule
\end{tabular}
\end{table}

\begin{figure*}[h]
  \centering
  \begin{subfigure}{0.44\textwidth}
    \centering
    \includegraphics[width=\linewidth]{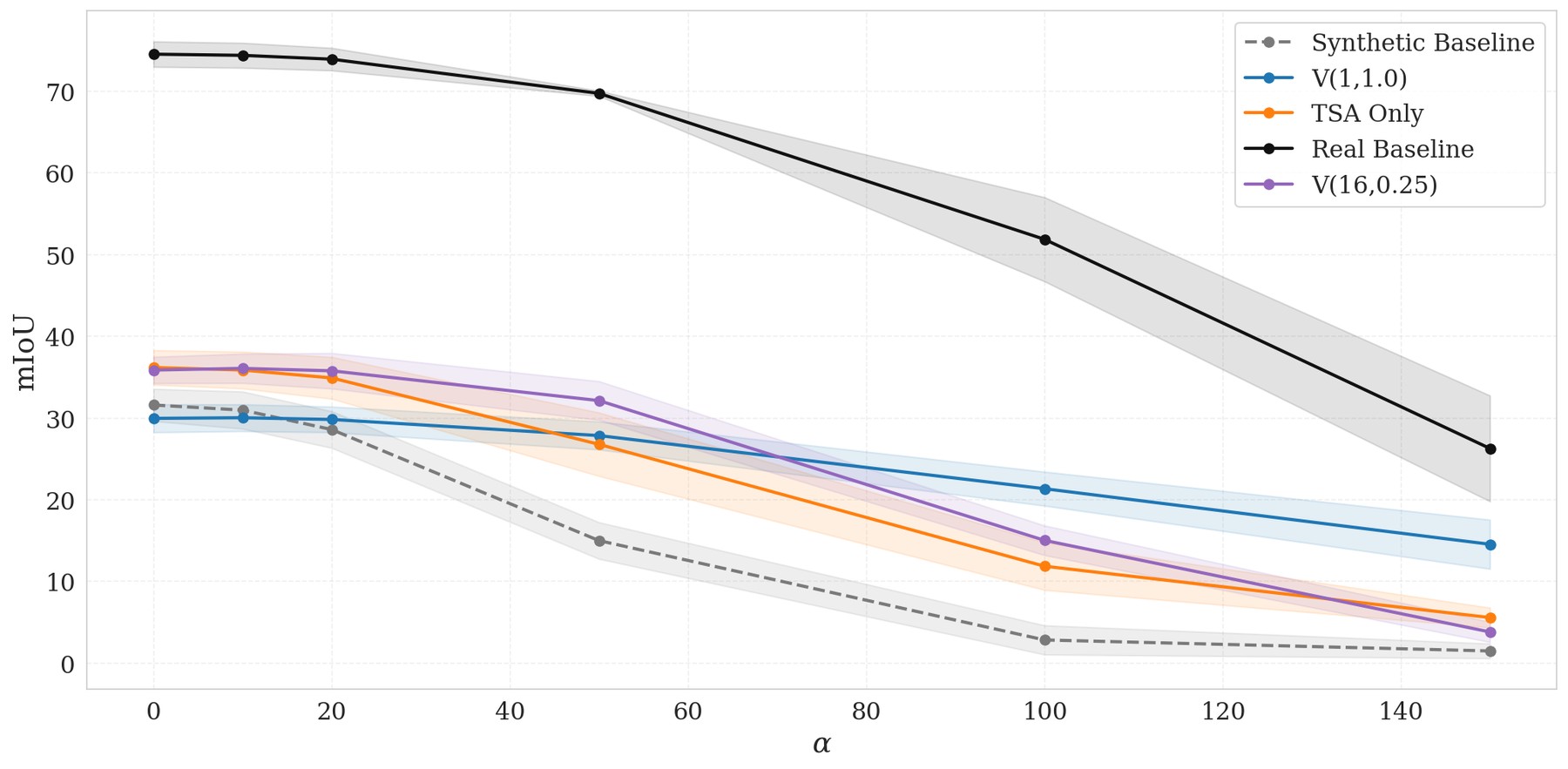}
    \caption{Elastic deformation}
  \end{subfigure}\hfill
  \begin{subfigure}{0.44\textwidth}
    \centering
    \includegraphics[width=\linewidth]{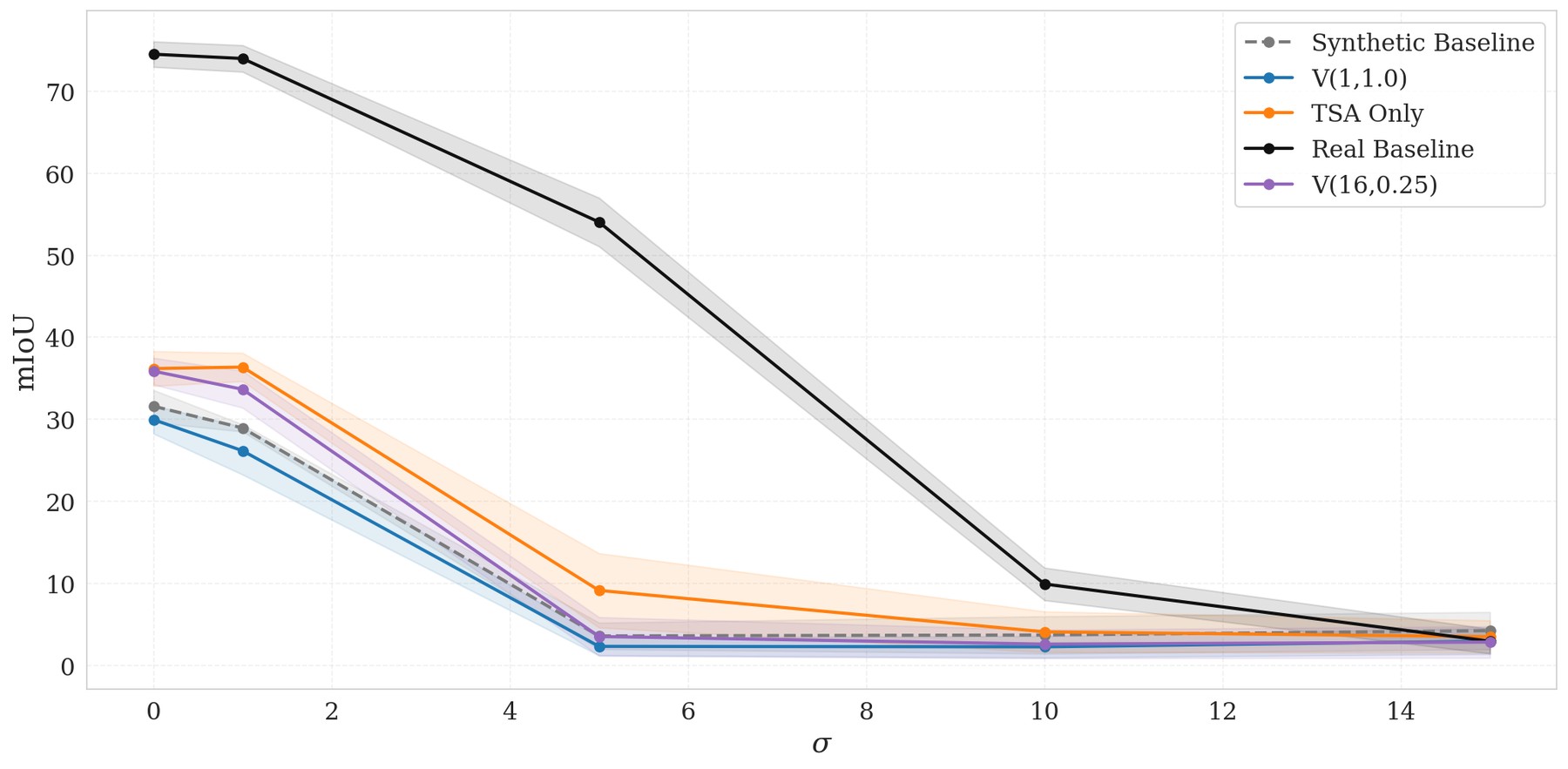}
    \caption{Low-pass filter}
  \end{subfigure}

  \vspace{0.6em}

  \begin{subfigure}{0.44\textwidth}
    \centering
    \includegraphics[width=\linewidth]{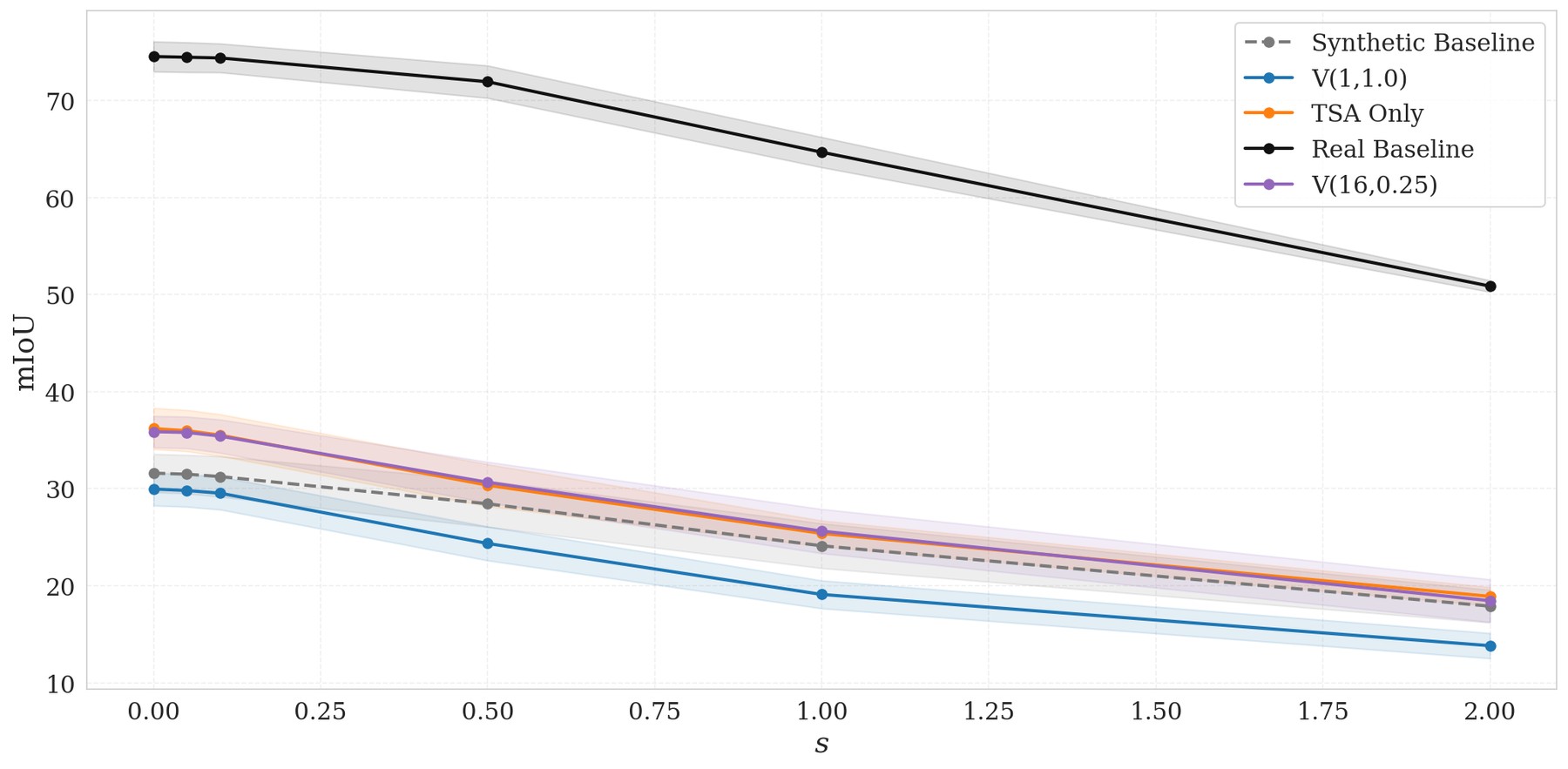}
    \caption{Swirl}
  \end{subfigure}\hfill
  \begin{subfigure}{0.44\textwidth}
    \centering
    \includegraphics[width=\linewidth]{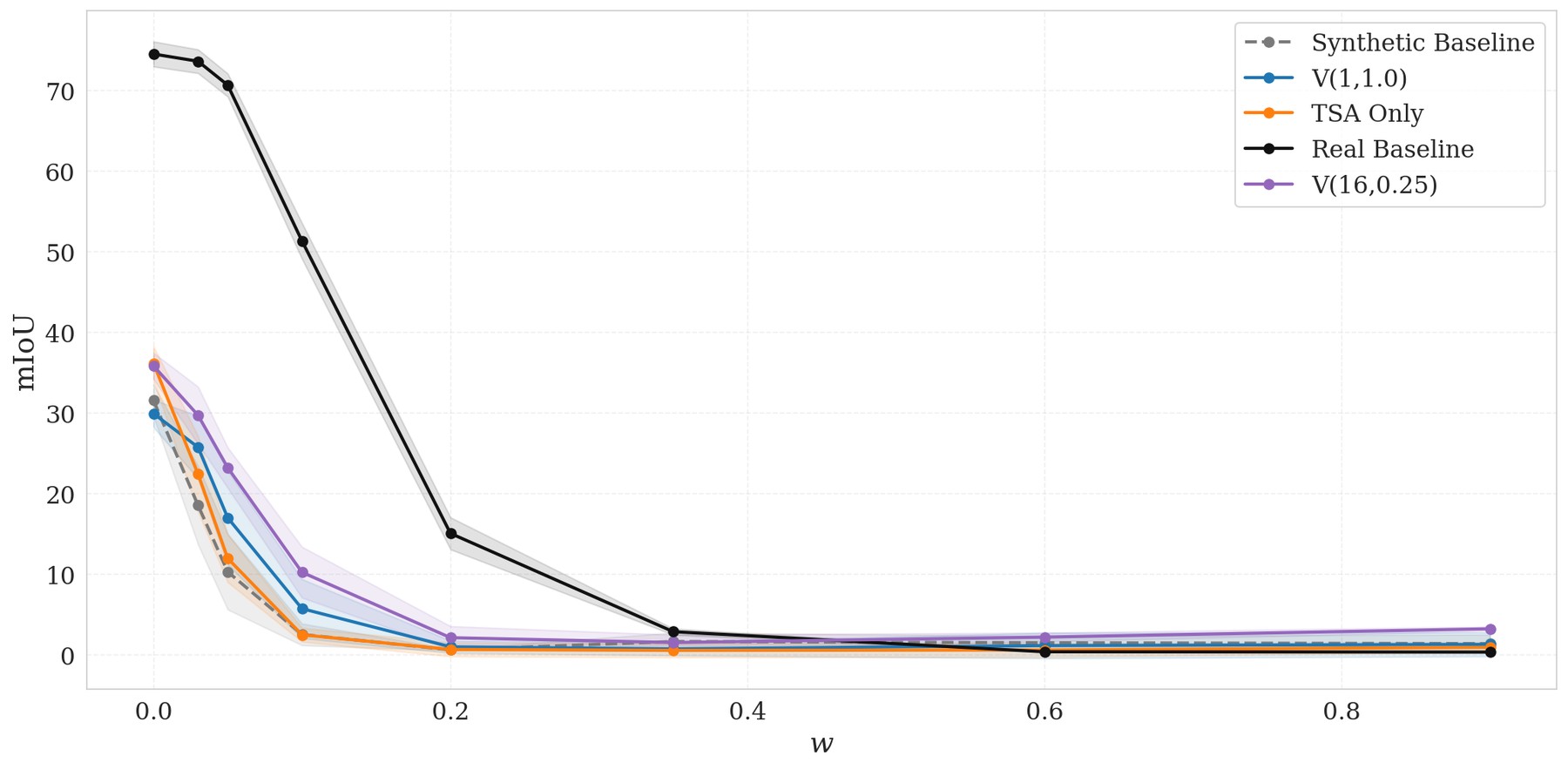}
    \caption{Uniform noise}
  \end{subfigure}

  \caption{Distortion robustness curves on interior data for \textbf{DeepLabV3+}: mIoU over for four representative distortions. First data point presents mIoU on clean data.}
  \label{fig:distortion_deeplab_2x2}
\end{figure*}

\begin{figure*}[h]
  \centering
  \begin{subfigure}{0.44\textwidth}
    \centering
    \includegraphics[width=\linewidth]{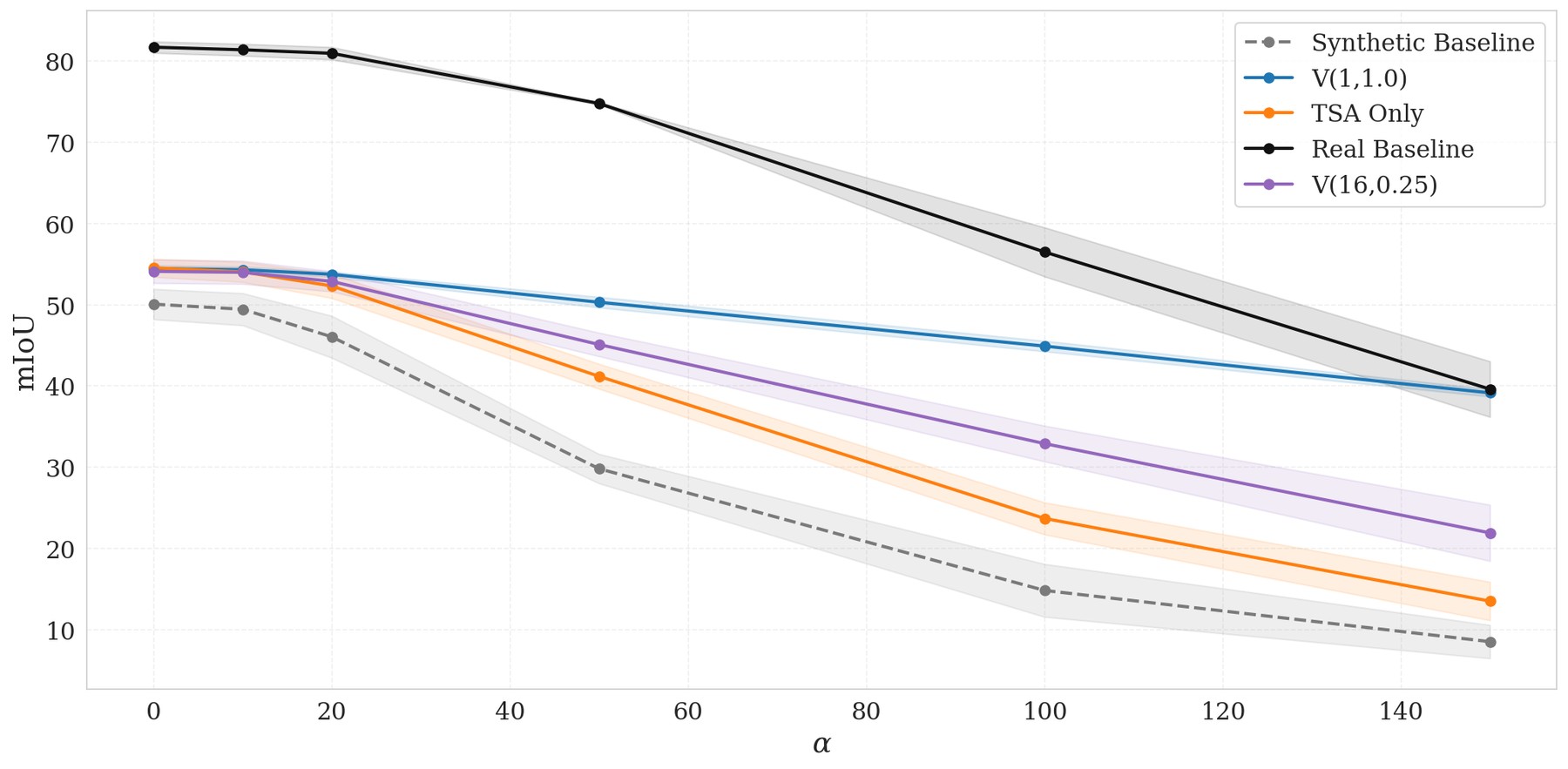}
    \caption{Elastic deformation}
  \end{subfigure}\hfill
  \begin{subfigure}{0.44\textwidth}
    \centering
    \includegraphics[width=\linewidth]{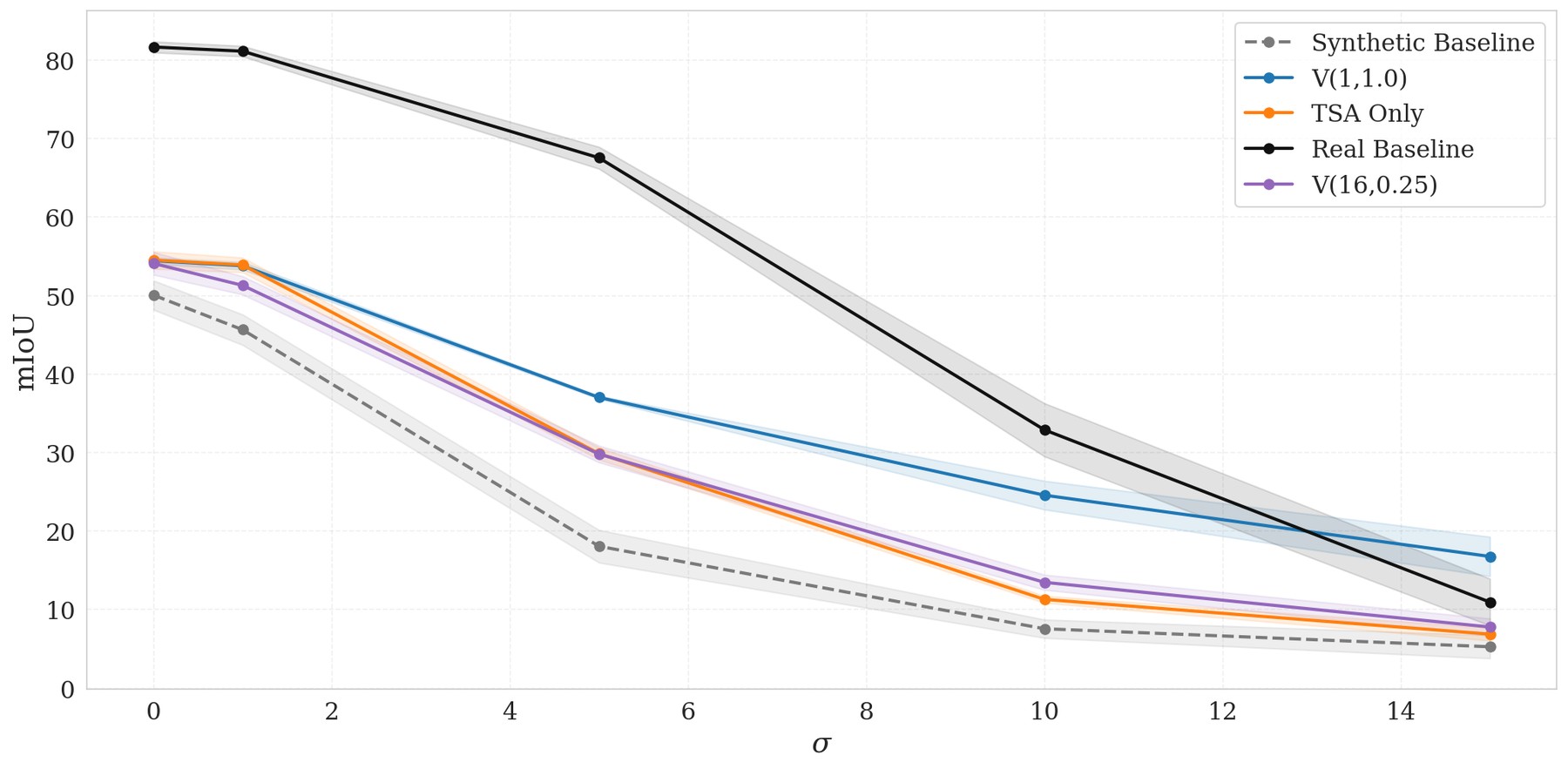}
    \caption{Low-pass filter}
  \end{subfigure}

  \vspace{0.6em}

  \begin{subfigure}{0.44\textwidth}
    \centering
    \includegraphics[width=\linewidth]{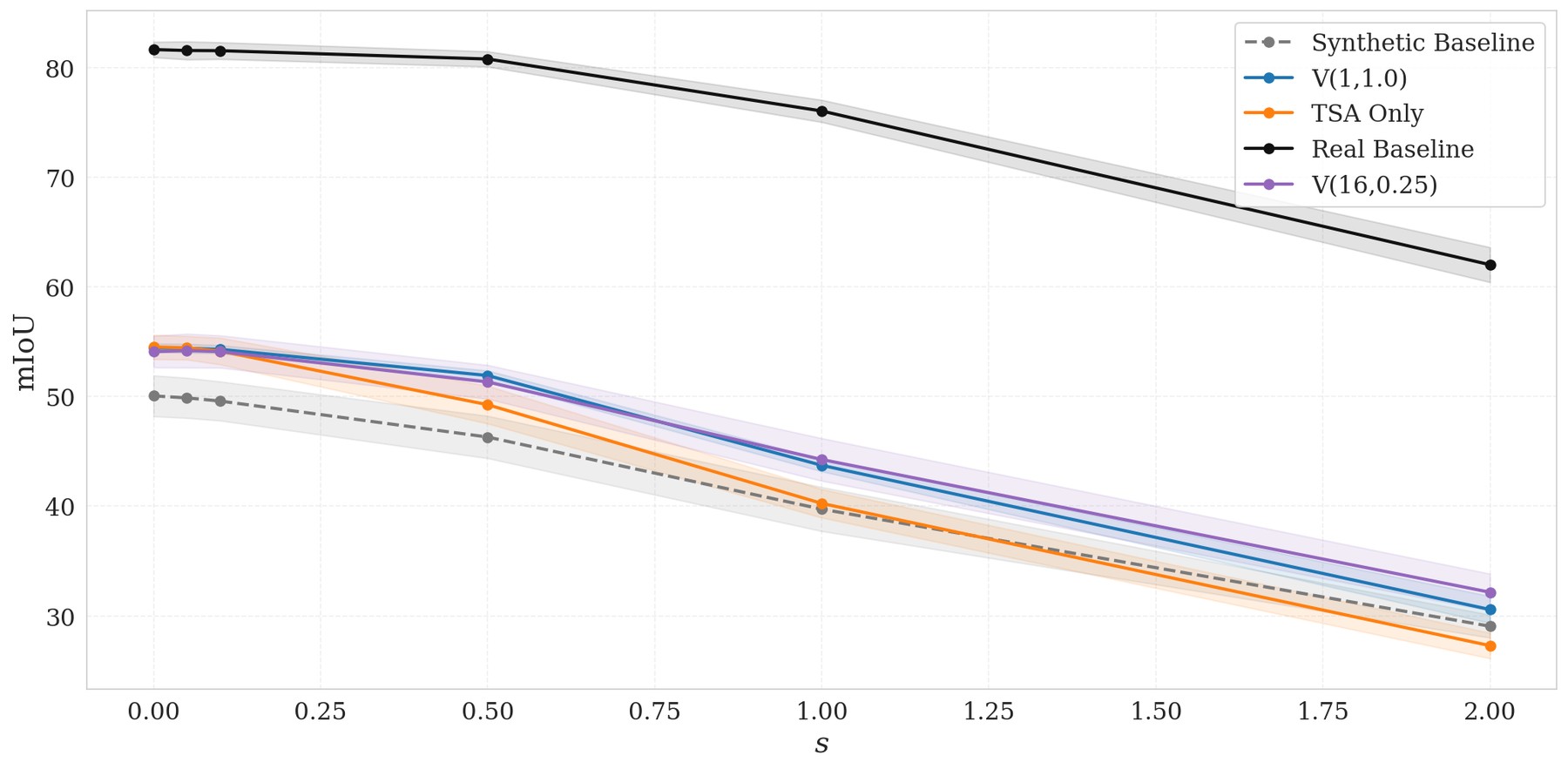}
    \caption{Swirl}
  \end{subfigure}\hfill
  \begin{subfigure}{0.44\textwidth}
    \centering
    \includegraphics[width=\linewidth]{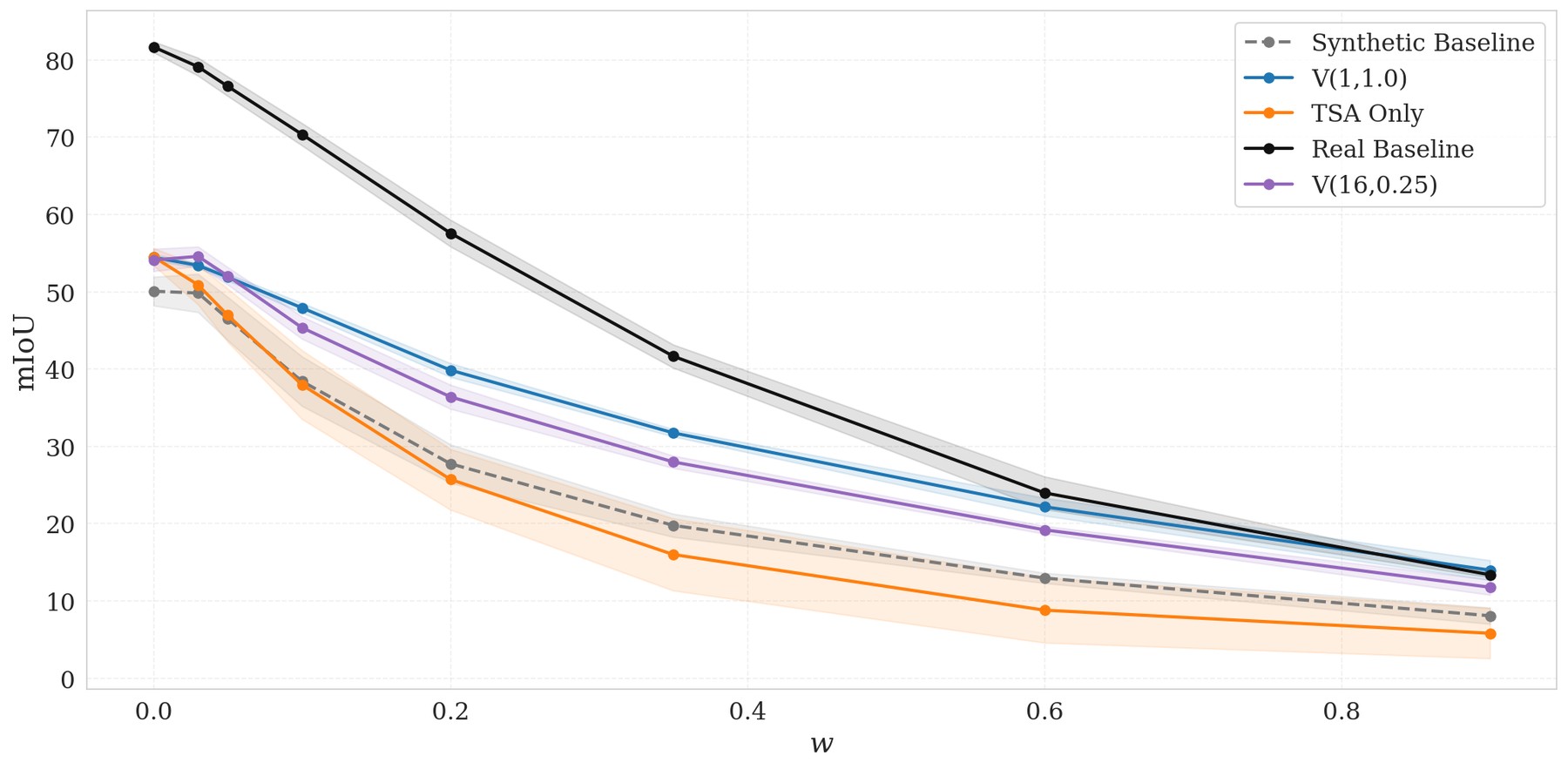}
    \caption{Uniform noise}
  \end{subfigure}

  \caption{Distortion robustness curves on interior data for \textbf{SegFormer}: mIoU over for four representative distortions. First data point presents mIoU on clean data.}
  \label{fig:distortion_segformer_2x2}
\end{figure*}

\begin{figure*}[h]
  \centering
  \begin{subfigure}{0.44\textwidth}
    \centering
    \includegraphics[width=\linewidth]{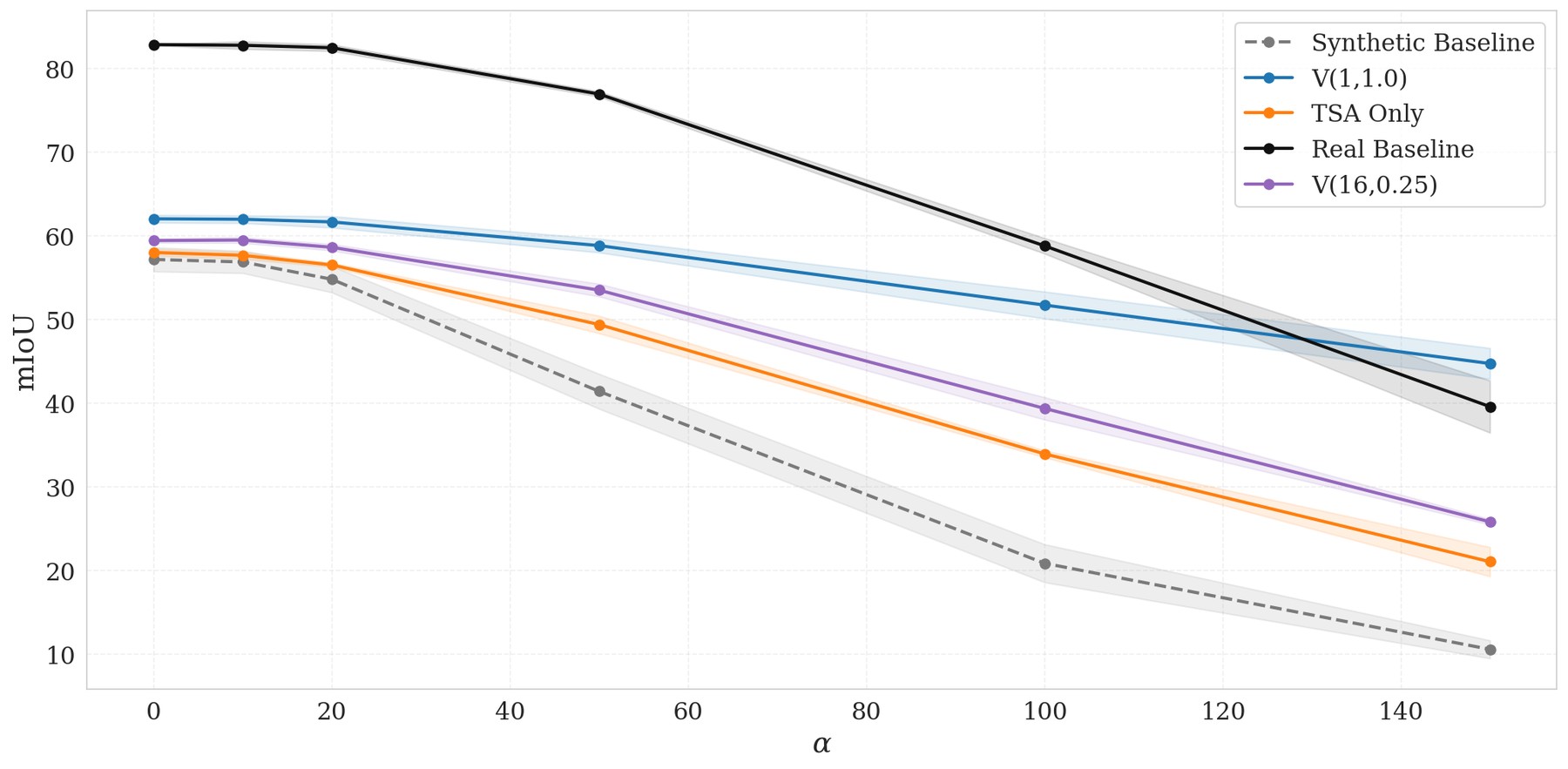}
    \caption{Elastic deformation}
  \end{subfigure}\hfill
  \begin{subfigure}{0.44\textwidth}
    \centering
    \includegraphics[width=\linewidth]{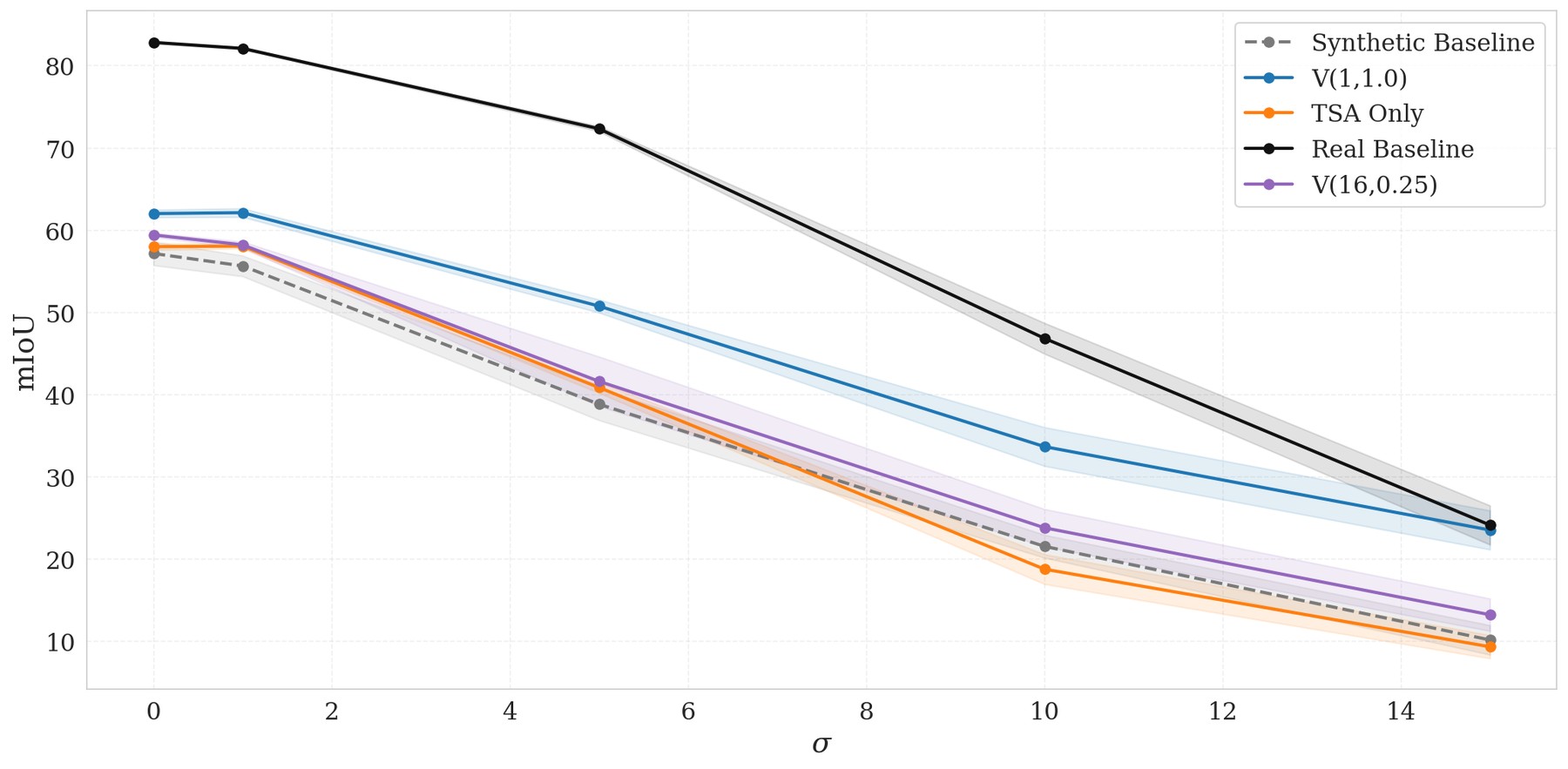}
    \caption{Low-pass filter}
  \end{subfigure}

  \vspace{0.6em}

  \begin{subfigure}{0.44\textwidth}
    \centering
    \includegraphics[width=\linewidth]{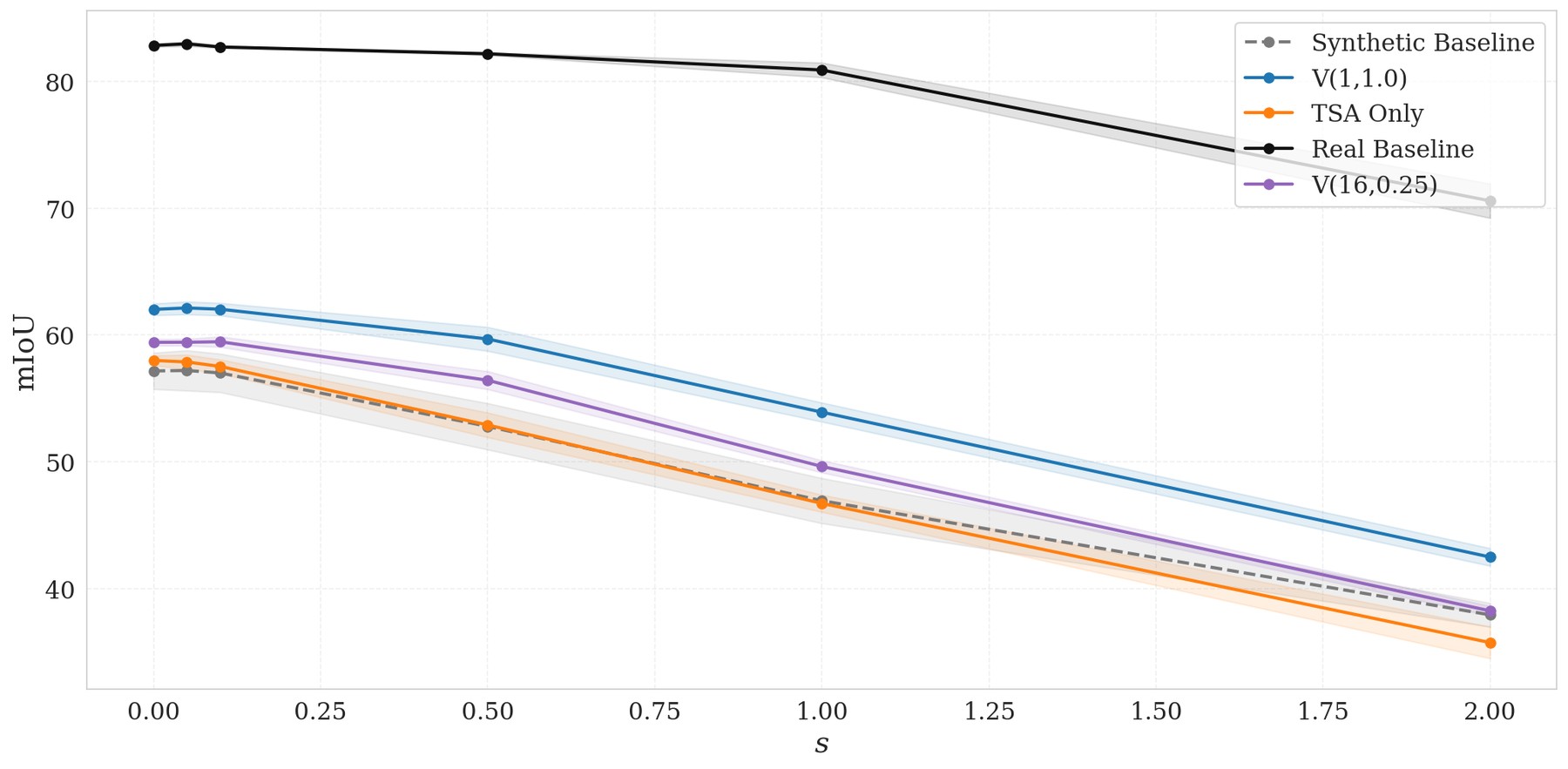}
    \caption{Swirl}
  \end{subfigure}\hfill
  \begin{subfigure}{0.44\textwidth}
    \centering
    \includegraphics[width=\linewidth]{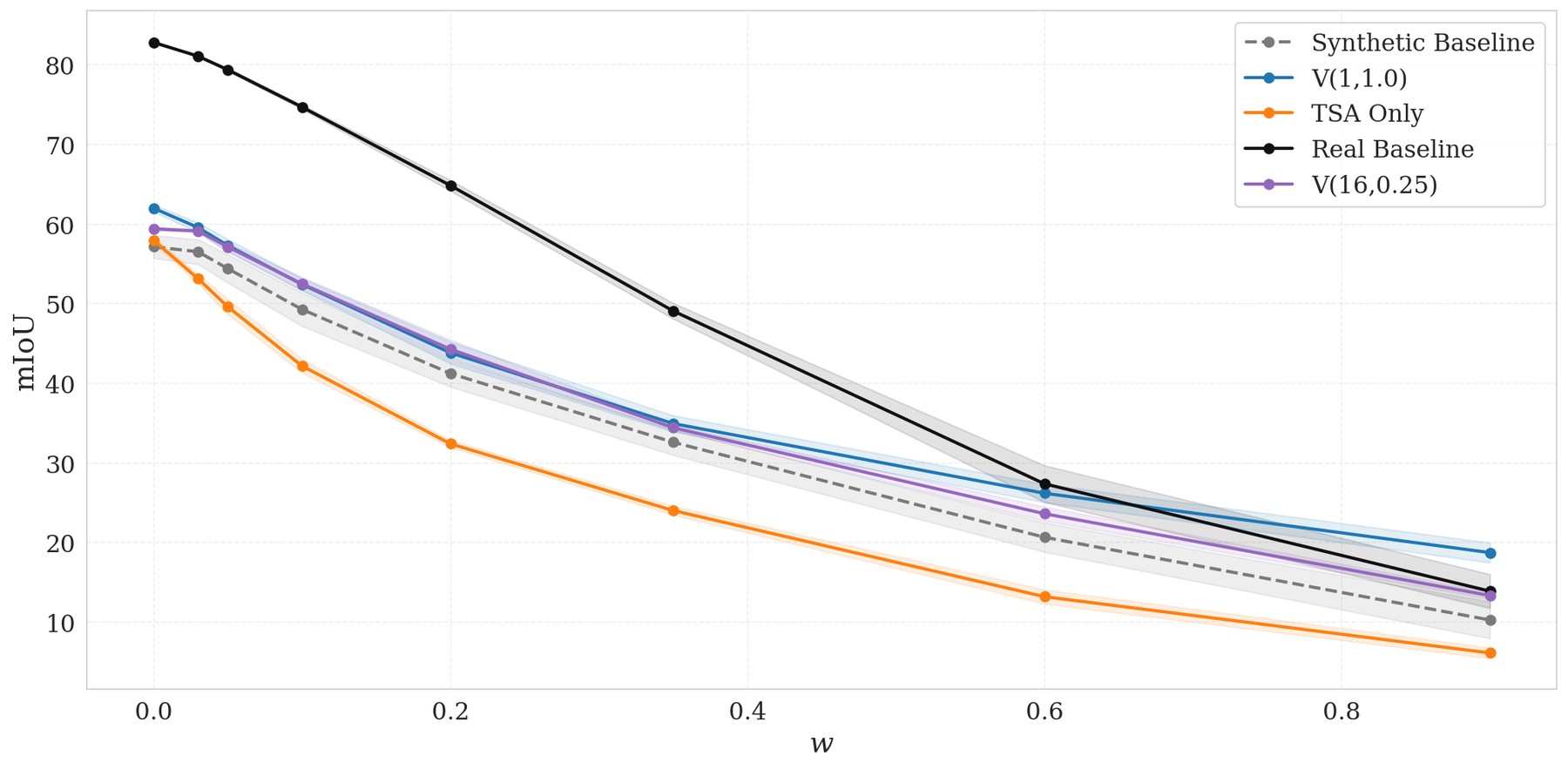}
    \caption{Uniform noise}
  \end{subfigure}

  \caption{Distortion robustness curves on interior data for \textbf{Mask2Former}: mIoU over four representative distortions. First data point presents mIoU on clean data.}
  \label{fig:distortion_m2f_2x2}
\end{figure*}

\begin{figure*}[h]
  \centering
  \begin{subfigure}{0.44\textwidth}
    \centering
    \includegraphics[width=\linewidth]{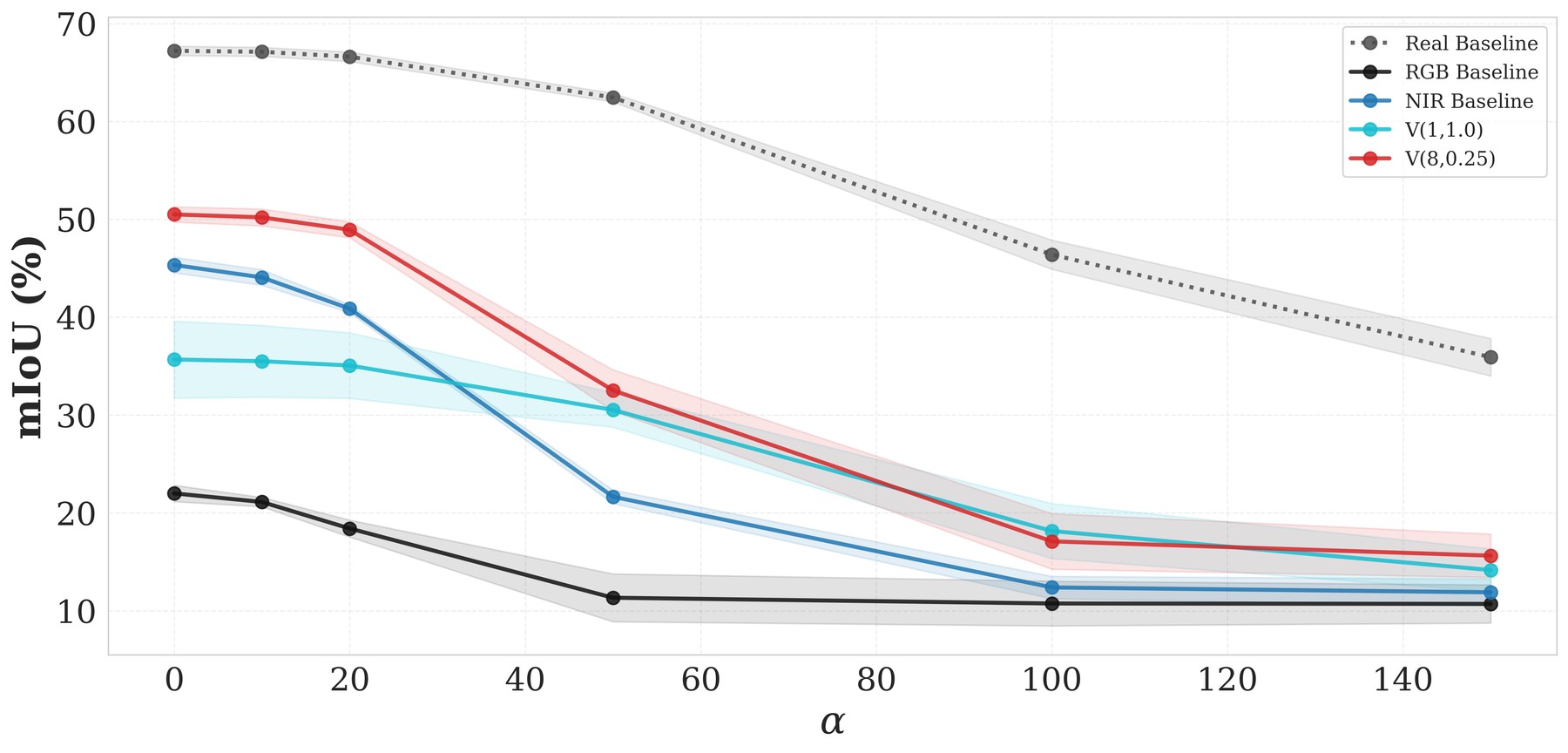}
    \caption{Elastic deformation}
  \end{subfigure}\hfill
  \begin{subfigure}{0.44\textwidth}
    \centering
    \includegraphics[width=\linewidth]{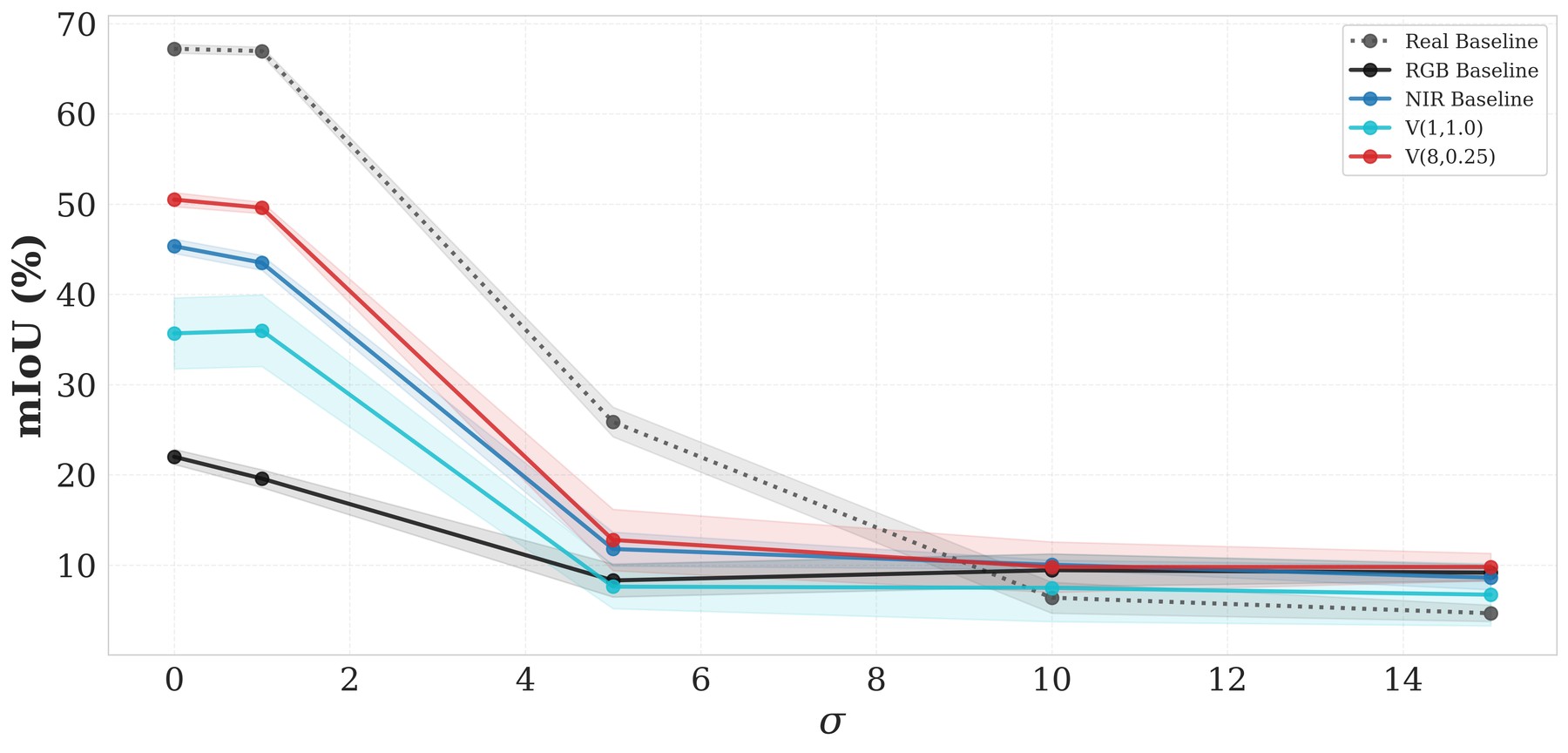}
    \caption{Low-pass filter}
  \end{subfigure}

  \vspace{0.6em}

  \begin{subfigure}{0.44\textwidth}
    \centering
    \includegraphics[width=\linewidth]{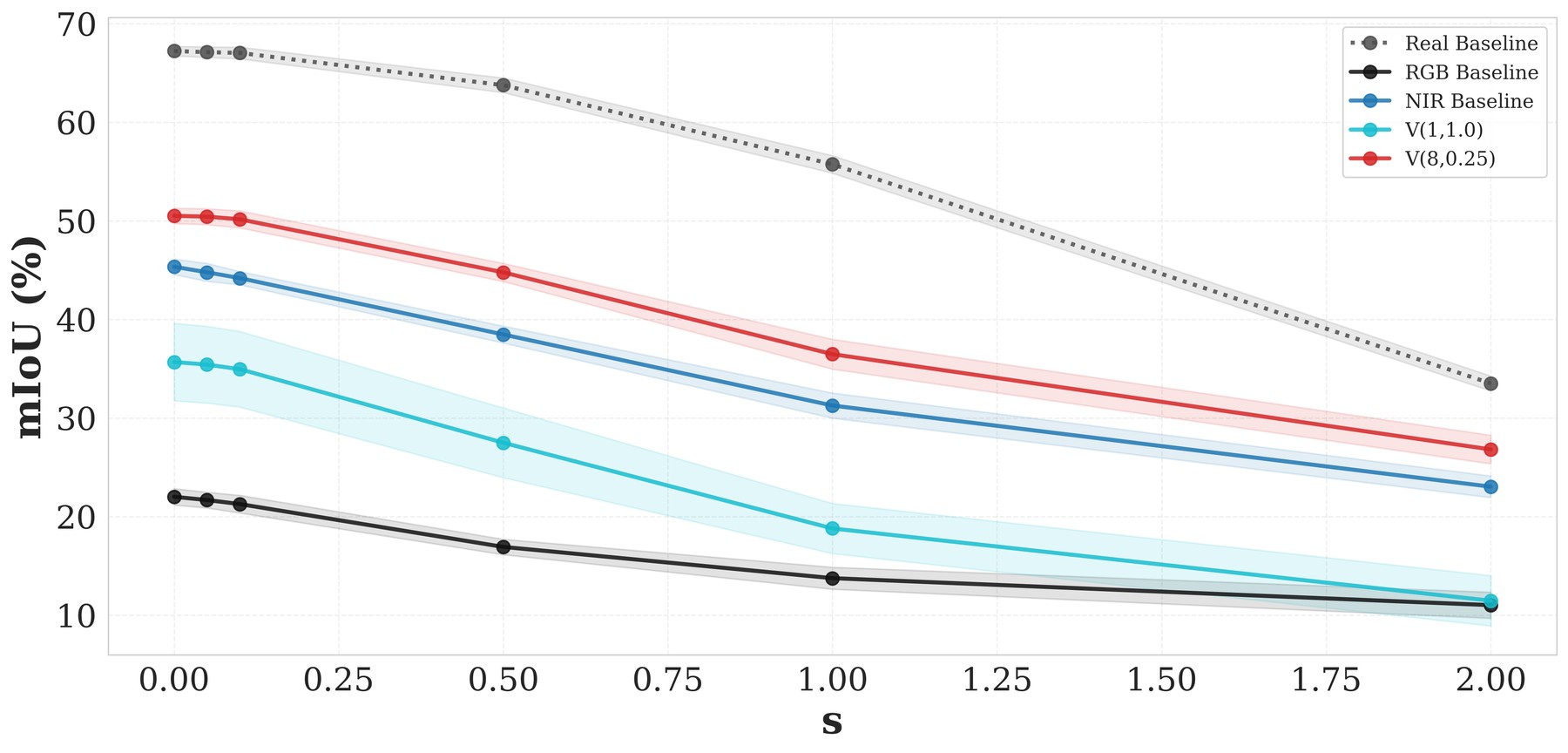}
    \caption{Swirl}
  \end{subfigure}\hfill
  \begin{subfigure}{0.44\textwidth}
    \centering
    \includegraphics[width=\linewidth]{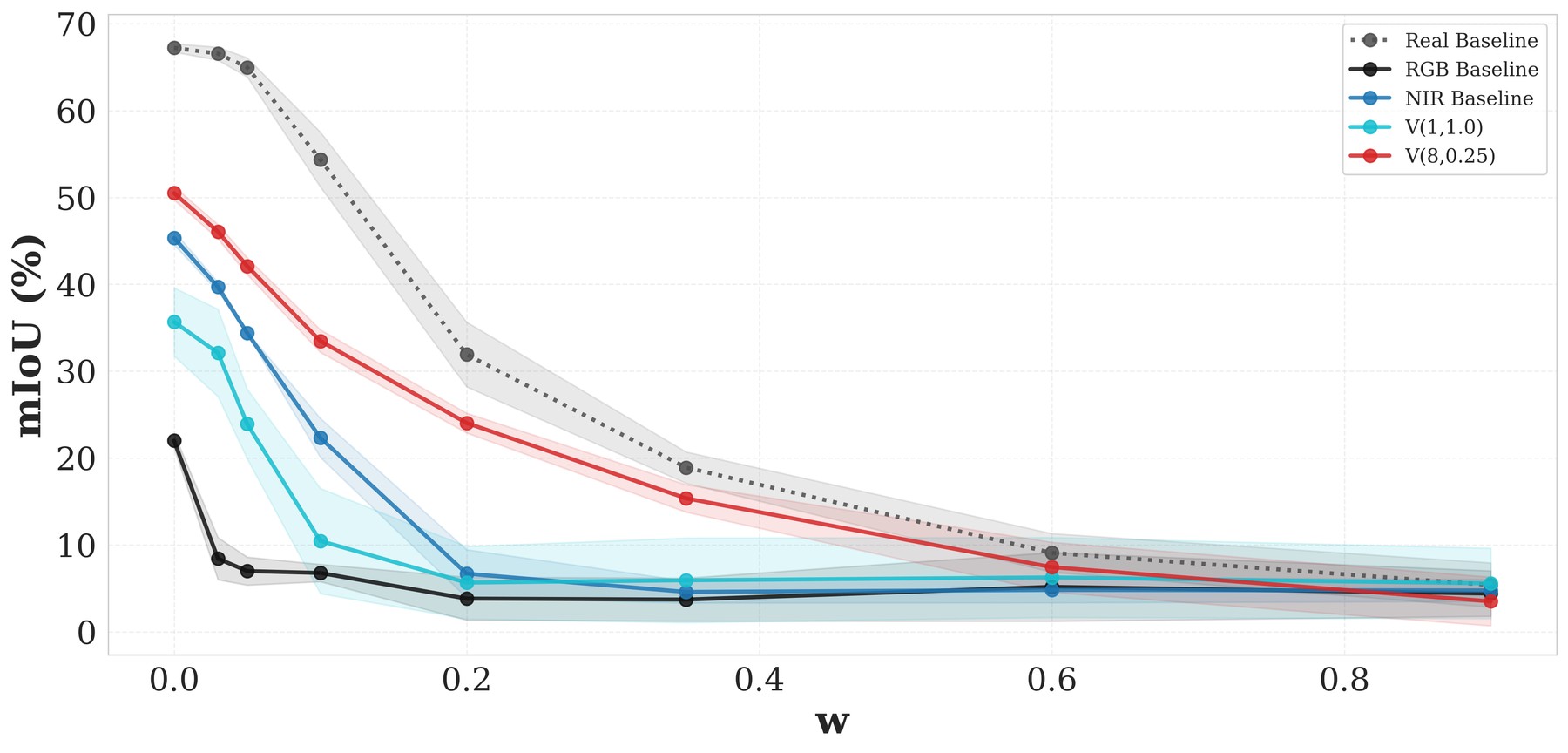}
    \caption{Uniform noise}
  \end{subfigure}

  \caption{Distortion robustness curves on exterior data for \textbf{DeepLabV3+}: mIoU over for four representative distortions. First data point presents mIoU on clean data.}
  \label{fig:distortion_deeplab_2x2_ext}
\end{figure*}

\begin{figure*}[h]
  \centering
  \begin{subfigure}{0.44\textwidth}
    \centering
    \includegraphics[width=\linewidth]{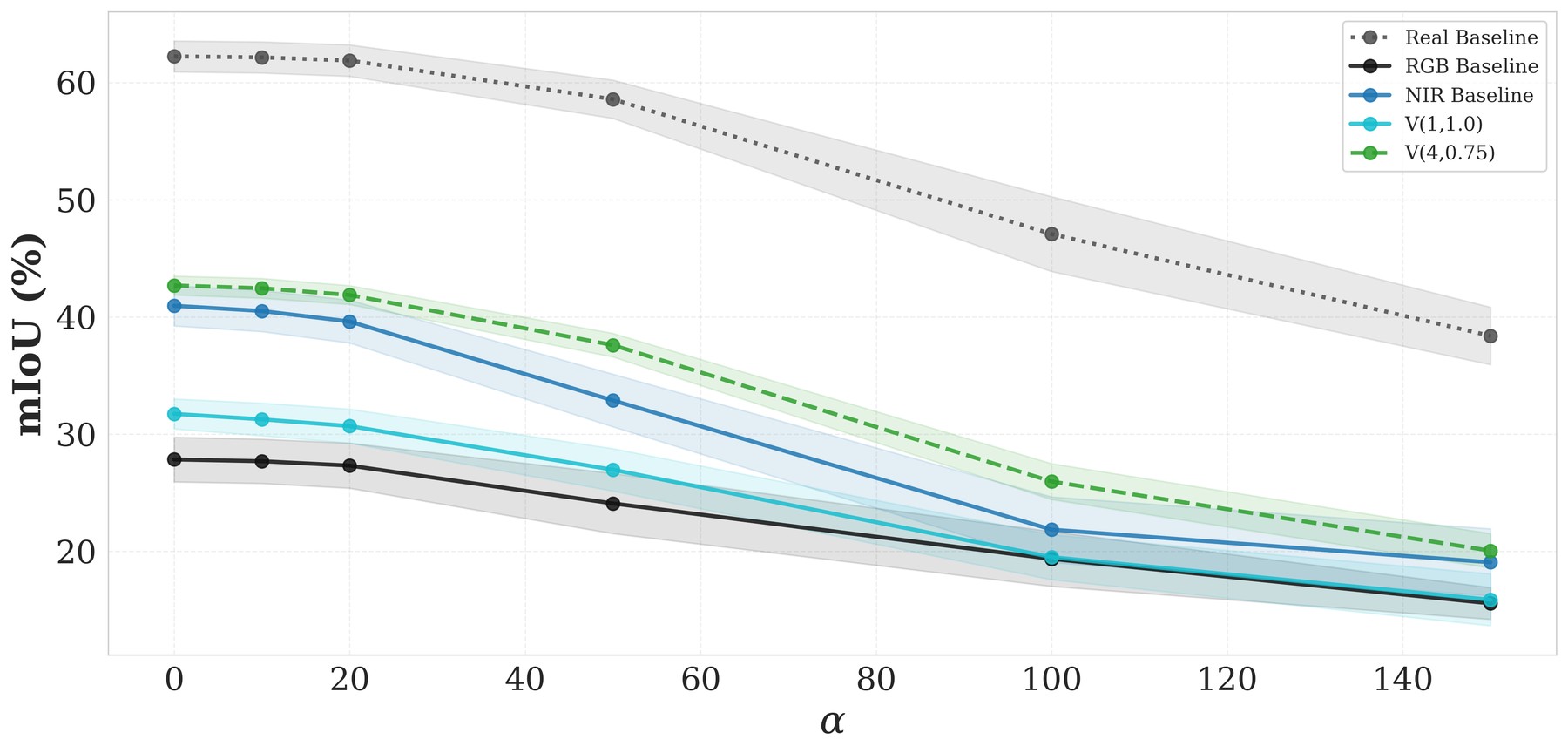}
    \caption{Elastic deformation}
  \end{subfigure}\hfill
  \begin{subfigure}{0.44\textwidth}
    \centering
    \includegraphics[width=\linewidth]{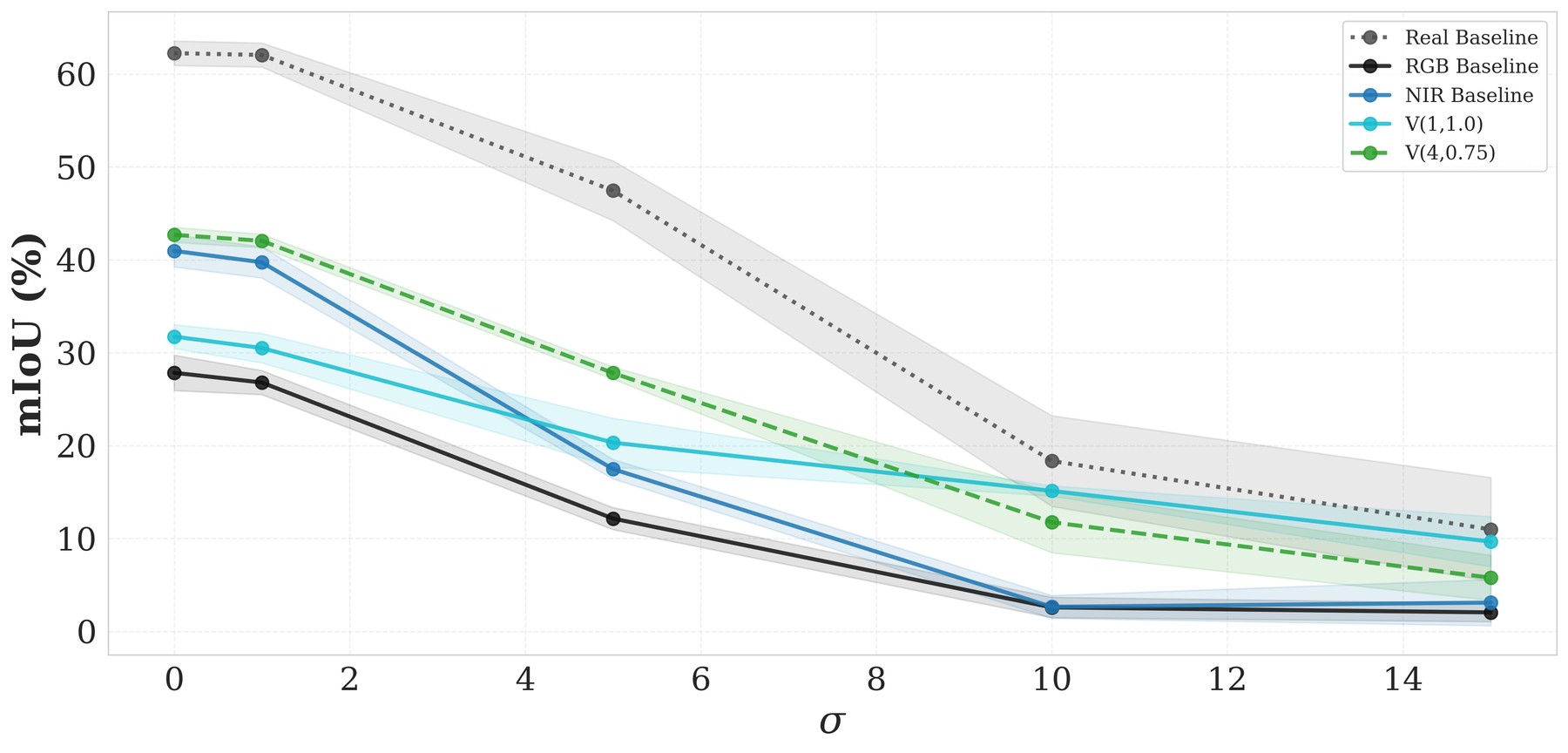}
    \caption{Low-pass filter}
  \end{subfigure}

  \vspace{0.6em}

  \begin{subfigure}{0.44\textwidth}
    \centering
    \includegraphics[width=\linewidth]{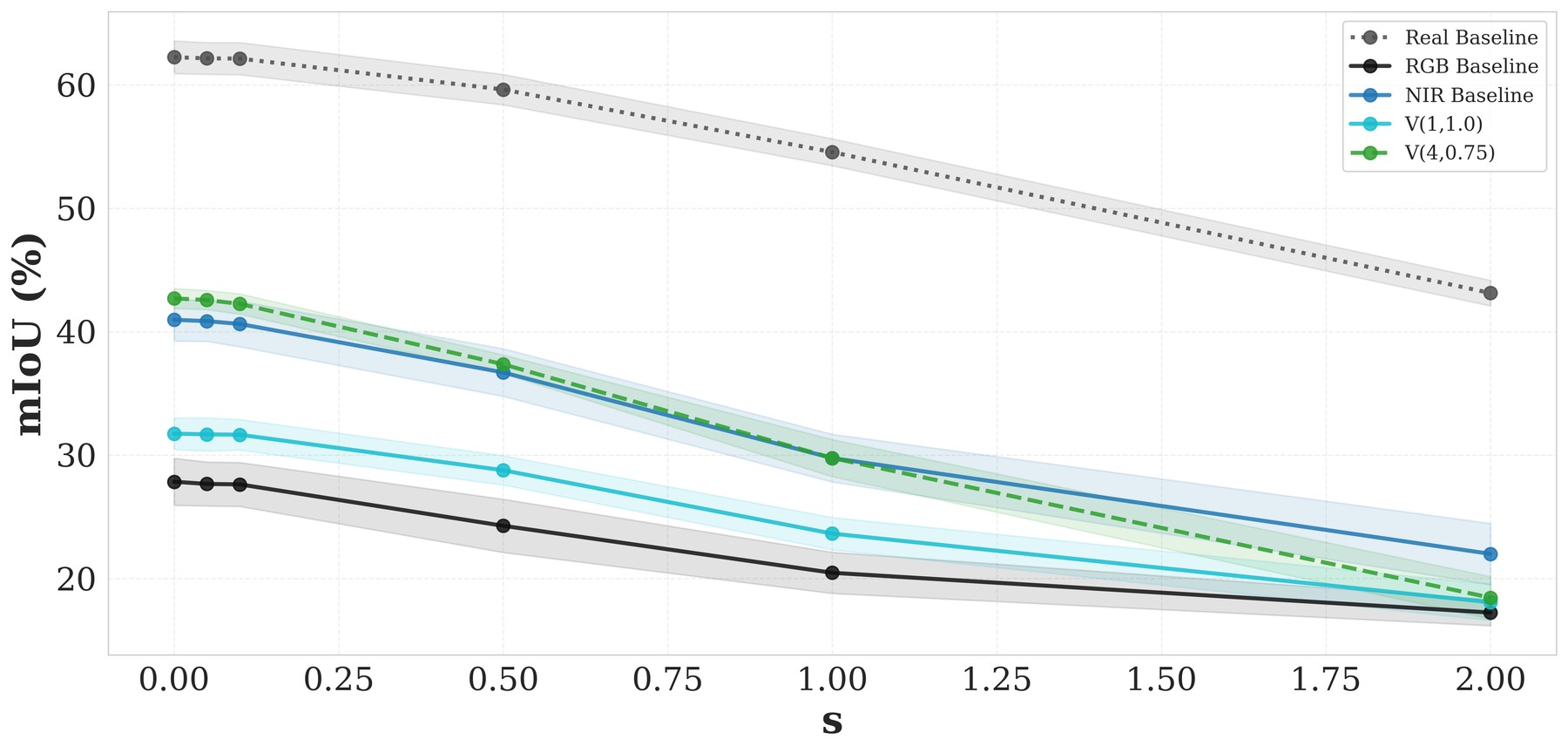}
    \caption{Swirl}
  \end{subfigure}\hfill
  \begin{subfigure}{0.44\textwidth}
    \centering
    \includegraphics[width=\linewidth]{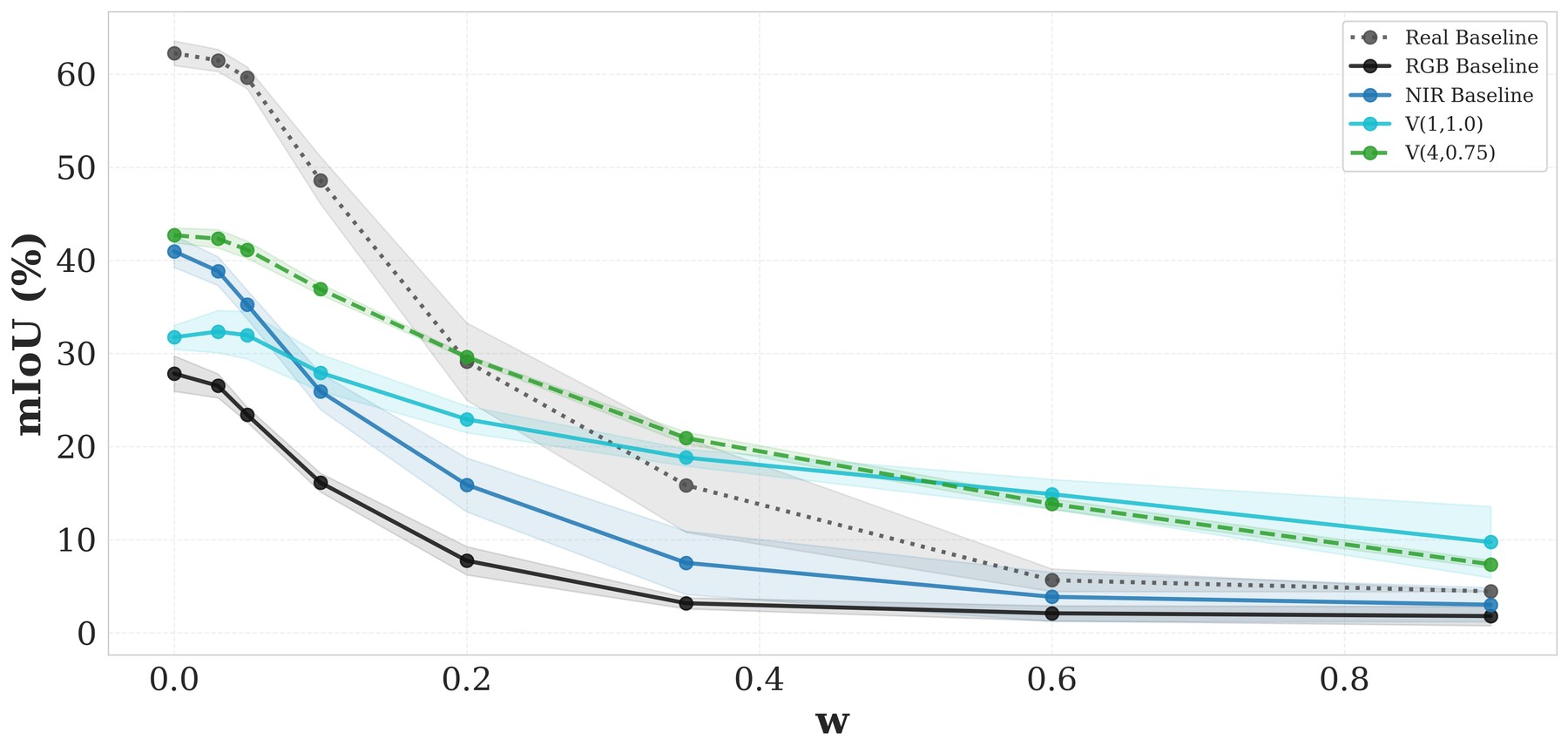}
    \caption{Uniform noise}
  \end{subfigure}

  \caption{Distortion robustness curves on exterior data for \textbf{SegFormer}: mIoU over for four representative distortions. First data point presents mIoU on clean data.}
  \label{fig:distortion_segformer_2x2_ext}
\end{figure*}

\begin{figure*}[h]
  \centering
  \begin{subfigure}{0.44\textwidth}
    \centering
    \includegraphics[width=\linewidth]{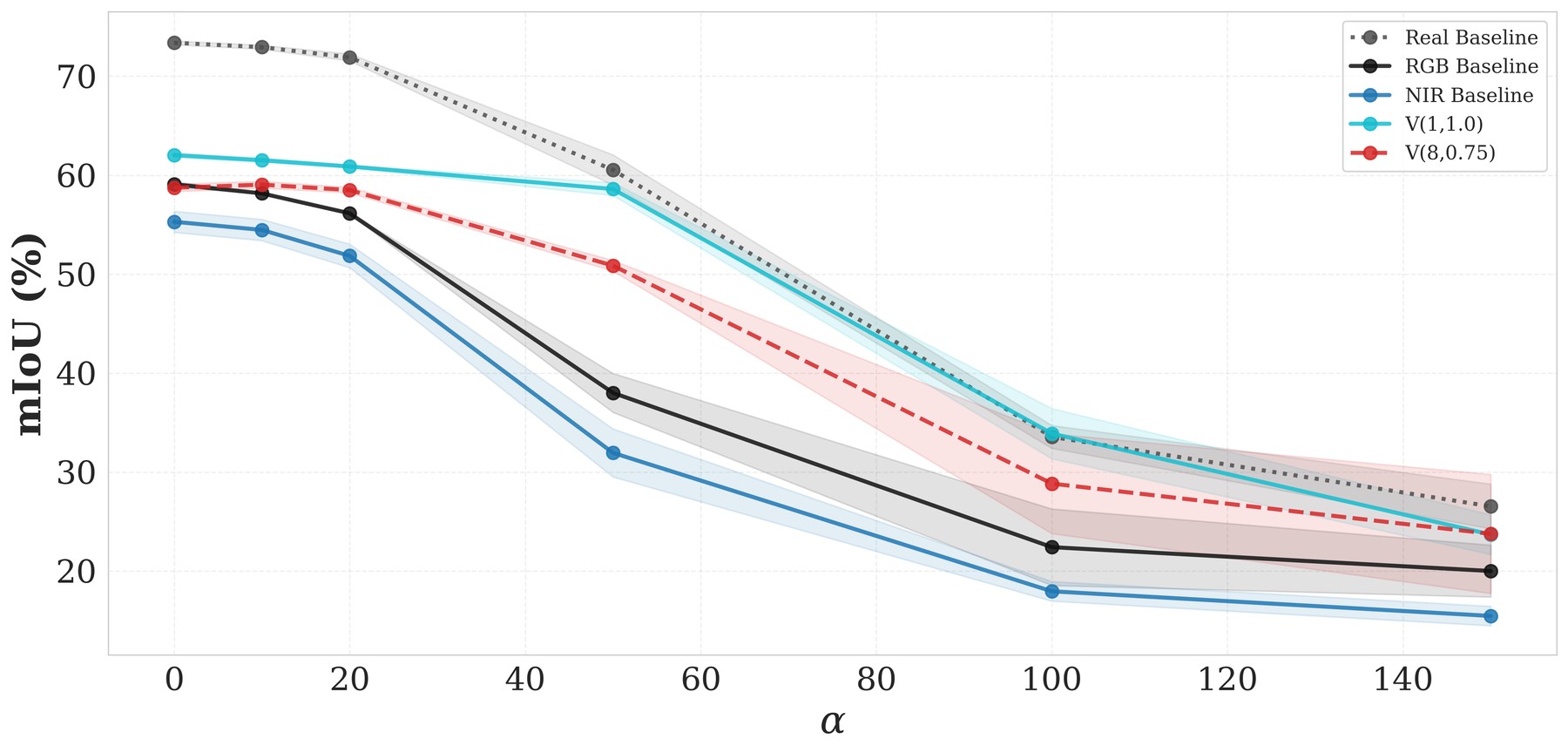}
    \caption{Elastic deformation}
  \end{subfigure}\hfill
  \begin{subfigure}{0.44\textwidth}
    \centering
    \includegraphics[width=\linewidth]{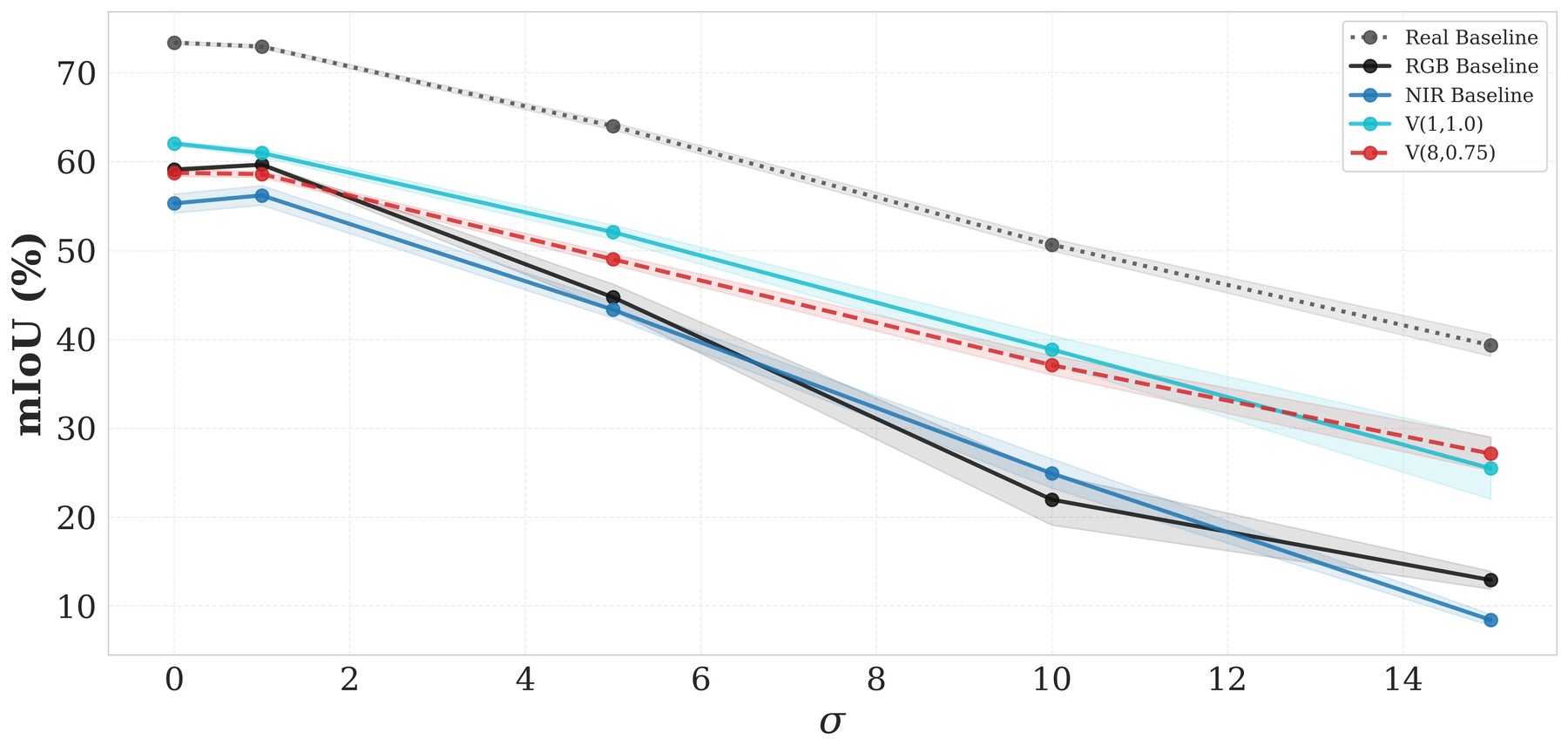}
    \caption{Low-pass filter}
  \end{subfigure}

  \vspace{0.6em}

  \begin{subfigure}{0.44\textwidth}
    \centering
    \includegraphics[width=\linewidth]{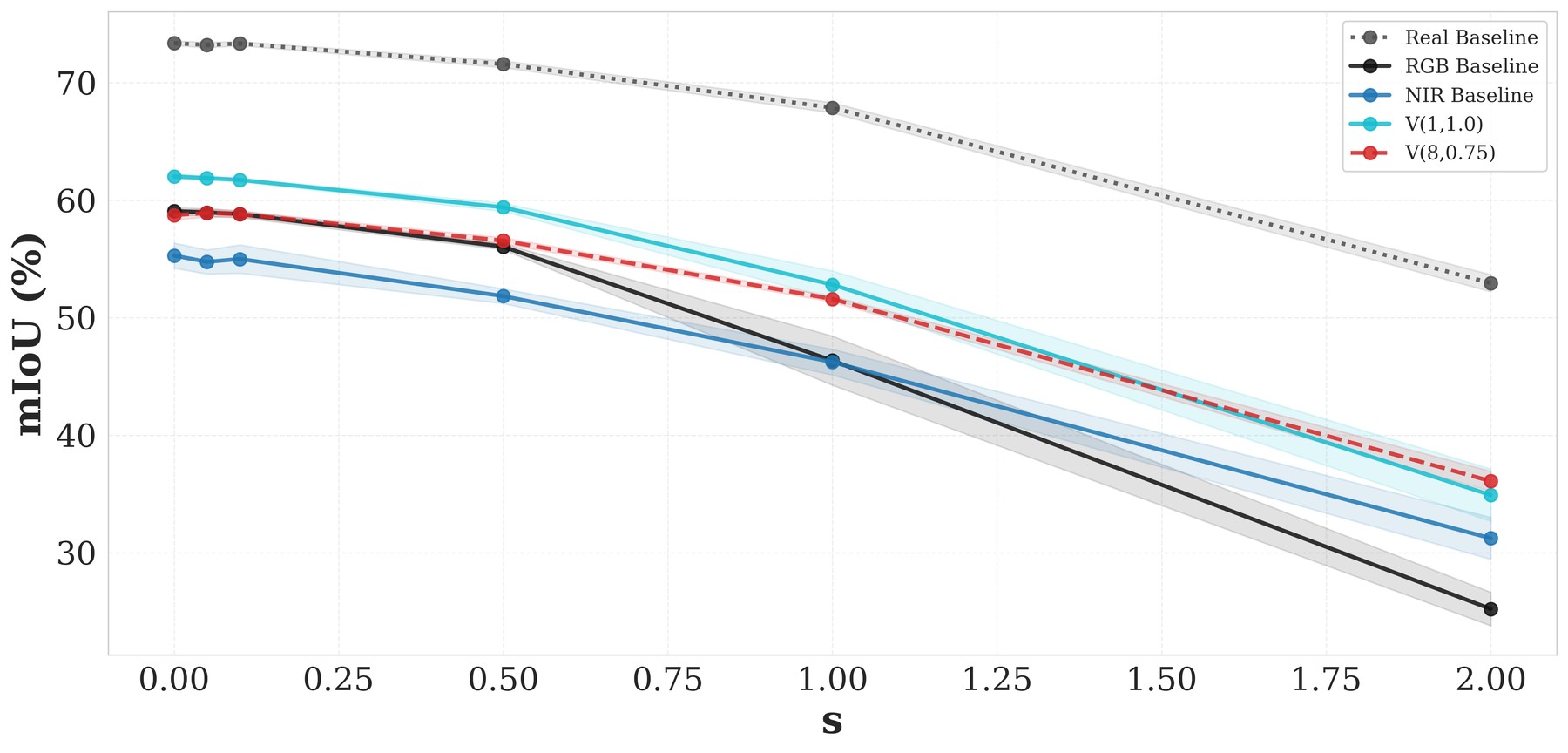}
    \caption{Swirl}
  \end{subfigure}\hfill
  \begin{subfigure}{0.44\textwidth}
    \centering
    \includegraphics[width=\linewidth]{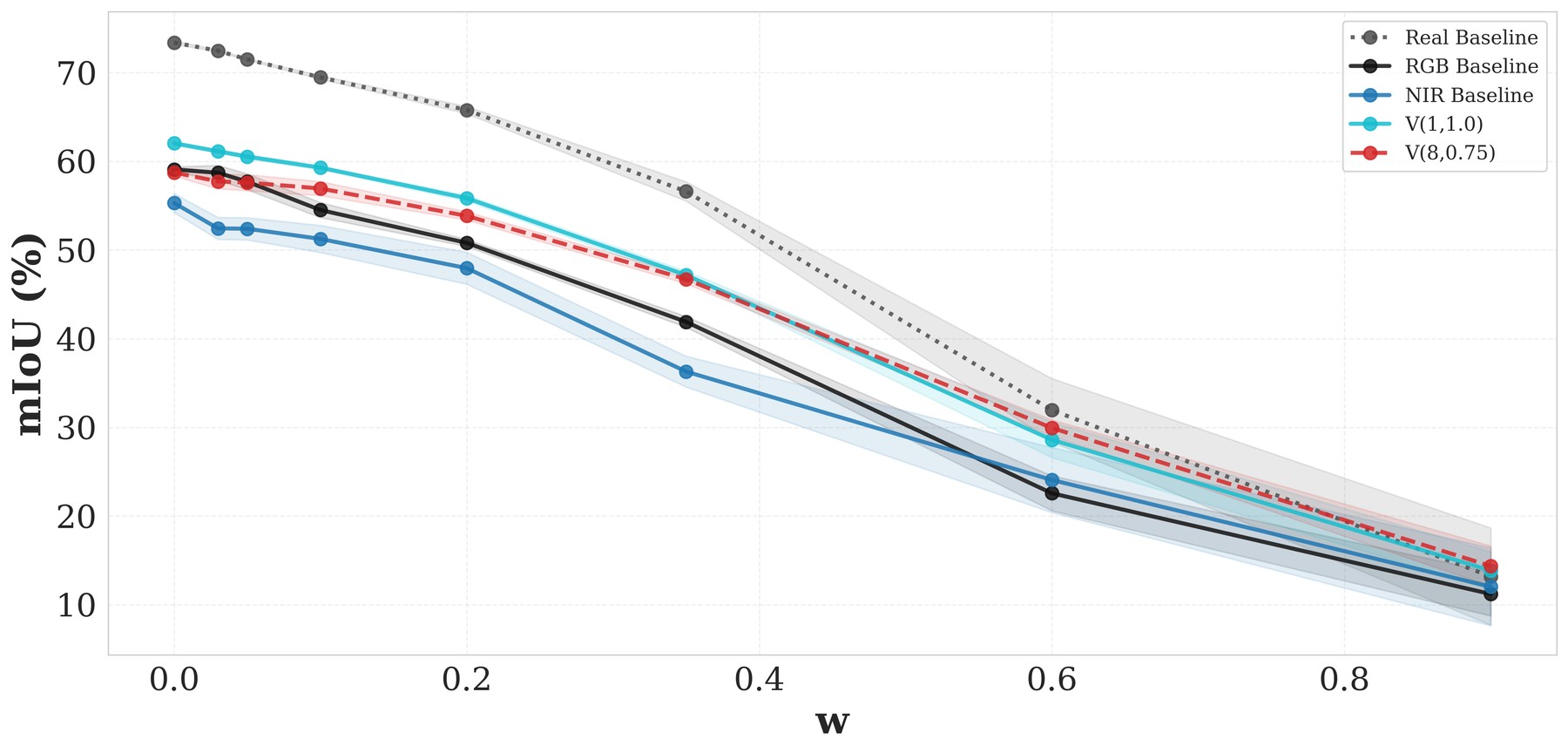}
    \caption{Uniform noise}
  \end{subfigure}

  \caption{Distortion robustness curves on exterior data for \textbf{Mask2Former}: mIoU over four representative distortions. First data point presents mIoU on clean data.}
  \label{fig:distortion_m2f_2x2_ext}
\end{figure*}

\begin{table}[h]
\centering
\caption{mIoU on real-world validation sets (median over 3 runs, $\pm$ shows distance to min/max). Best per column in \textbf{bold}.}
\label{tab:full_main_results_3runs}
\small
\setlength{\tabcolsep}{5pt}
\renewcommand{\arraystretch}{1.1}
\resizebox{\textwidth}{!}{%

\begin{tabular}{l ccc ccc}
\toprule
& \multicolumn{3}{c}{\textbf{Interior}}
& \multicolumn{3}{c}{\textbf{Exterior}} \\
\cmidrule(lr){2-4}\cmidrule(lr){5-7}
\textbf{Method} & DeepLabV3+ & SegFormer & M2Former & DeepLabV3+ & SegFormer & M2Former \\
\midrule
Real Baseline                  & $75.12^{+0.40}_{-2.15}$ & $81.71^{+0.57}_{-0.67}$ & $82.87^{+0.05}_{-0.14}$ & $66.99^{+0.78}_{-0.10}$ & $62.55^{+0.83}_{-1.76}$ & $73.48^{+0.02}_{-0.32}$ \\
Synthetic Baseline  & $32.00^{+1.11}_{-2.31}$ & $49.18^{+2.79}_{-0.13}$ & $56.79^{+1.78}_{-0.67}$ & $21.57^{+1.39}_{-0.11}$ & $28.14^{+1.44}_{-2.34}$ & $58.95^{+0.46}_{-0.06}$ \\
\midrule
TSA Only                      & $35.18^{+3.17}_{-0.14}$ & $53.97^{+1.69}_{-0.02}$ & $58.15^{+0.08}_{-0.56}$ & $44.95^{+1.30}_{-0.14}$ & $41.84^{+0.22}_{-2.86}$ & $55.26^{+1.12}_{-1.02}$ \\
VSD(1,1.0)                  & $29.72^{+1.86}_{-1.13}$ & $54.25^{+0.60}_{-0.01}$ & $\textbf{62.04}^{+0.37}_{-0.44}$ & $41.59^{+2.95}_{-2.79}$ & $40.53^{+0.20}_{-1.14}$ & $\textbf{61.91}^{+0.39}_{-0.04}$ \\
VSD(4,0.25) + TSA             & $43.59^{+1.02}_{-0.84}$ & $56.50^{+1.07}_{-0.30}$ & $59.92^{+1.05}_{-0.37}$ & $48.54^{+0.46}_{-1.04}$ & $40.44^{+1.67}_{-0.69}$ & $55.43^{+0.84}_{-0.18}$ \\
VSD(4,0.75) + TSA             & $42.33^{+0.09}_{-1.18}$ & $54.94^{+1.28}_{-0.65}$ & $60.86^{+0.20}_{-0.08}$ & $49.00^{+0.13}_{-0.60}$ & $42.53^{+1.04}_{-0.55}$ & $57.57^{+2.04}_{-1.03}$ \\
VSD(8,0.25) + TSA             & $43.92^{+0.30}_{-0.44}$ & $56.75^{+0.77}_{-0.78}$ & $60.13^{+0.54}_{-0.26}$ & $\textbf{50.44}^{+0.87}_{-0.69}$ & $42.02^{+1.38}_{-0.10}$ & $57.16^{+1.06}_{-0.97}$ \\
VSD(8,0.75) + TSA             & $41.51^{+0.17}_{-0.22}$ & $55.59^{+0.18}_{-0.25}$ & $60.83^{+1.58}_{-0.58}$ & $48.64^{+0.20}_{-3.44}$ & $41.04^{+2.52}_{-0.57}$ & $58.81^{+0.25}_{-0.50}$ \\
VSD(16,0.25) + TSA             & $\textbf{44.26}^{+0.93}_{-0.41}$ & $\textbf{57.30}^{+0.08}_{-0.32}$ & $60.55^{+0.77}_{-0.74}$ & $48.31^{+0.03}_{-1.76}$ & $42.44^{+2.99}_{-0.61}$ & $56.29^{+0.26}_{-1.84}$ \\
VSD(16,0.75) + TSA             & $43.62^{+0.70}_{-2.17}$ & $54.32^{+1.09}_{-0.94}$ & $60.62^{+0.99}_{-0.84}$ & $46.36^{+0.51}_{-0.53}$ & $\textbf{43.69}^{+0.32}_{-1.66}$ & $55.35^{+0.30}_{-1.58}$ \\
\bottomrule
\end{tabular}}
\end{table}

\begin{table}[h]
\centering
\caption{Mid-Stop after $30,\!000$ iterations (halft the training schedule). mIoU on real-world validation sets (median over 3 runs, $\pm$ shows distance to min/max). $n$: Voronoi cells; $\alpha$: stylization probability. Best per column in \textbf{bold}.}
\label{tab:30k_results_3runs}
\begin{tabular}{l ccc ccc}
\toprule
& \multicolumn{3}{c}{\textbf{Interior}}
& \multicolumn{3}{c}{\textbf{Exterior}} \\
\cmidrule(lr){2-4}\cmidrule(lr){5-7}
\textbf{Method} & DeepLabV3+ & SegFormer & M2Former & DeepLabV3+ & SegFormer & M2Former \\
\midrule
Real Baseline                  & $75.12^{+0.40}_{-2.56}$ & $81.04^{+1.24}_{-1.92}$ & $82.73^{+0.19}_{-0.18}$ & $61.58^{+0.51}_{-0.64}$ & $59.50^{+0.63}_{-4.14}$ & $73.23^{+0.19}_{-0.03}$\\
Synthetic Baseline & $32.00^{+1.11}_{-1.25}$ & $49.18^{+2.79}_{-1.03}$ & $56.79^{+1.78}_{-3.32}$ & $27.45^{+3.01}_{-4.21}$ & $27.68^{+3.46}_{-4.33}$ & $58.21^{+1.90}_{-0.79}$\\
\midrule
TSA Only                      & $35.18^{+3.17}_{-8.59}$ & $53.95^{+1.71}_{-2.28}$ & $57.91^{+0.32}_{-0.32}$ & $45.46^{+3.42}_{-6.51}$ & $41.37^{+0.29}_{-0.64}$ & $56.58^{+1.72}_{-3.22}$ \\
VSD(1,1.0)                  & $29.72^{+1.86}_{-7.58}$ & $54.25^{+0.60}_{-1.93}$ & $\textbf{61.60}^{+0.81}_{-2.27}$ & $36.02^{+4.13}_{-3.79}$ & $35.08^{+0.50}_{-5.94}$ & $61.05^{+0.57}_{-0.82}$ \\
VSD(4,0.25) + TSA             & $42.75^{+1.86}_{-8.23}$ & $56.20^{+1.37}_{-4.37}$ & $59.92^{+0.13}_{-0.37}$ & $45.03^{+0.50}_{-0.04}$ & $41.28^{+0.17}_{-8.79}$ & $56.45^{+2.65}_{-0.70}$ \\
VSD(4,0.75) + TSA             & $42.33^{+0.09}_{-6.38}$ & $54.94^{+1.28}_{-1.43}$ & $60.78^{+0.28}_{-1.49}$ & $\textbf{46.50}^{+0.40}_{-3.32}$ & $41.42^{+2.72}_{-6.31}$ & $56.39^{+3.28}_{-0.46}$ \\
VSD(8,0.25) + TSA             & $43.48^{+0.74}_{-7.94}$ & $56.75^{+0.77}_{-2.35}$ & $60.67^{+0.09}_{-0.80}$ & $44.01^{+1.92}_{-0.04}$ & $41.45^{+4.10}_{-0.14}$ & $57.39^{+0.77}_{-1.10}$ \\
VSD(8,0.75) + TSA             & $41.29^{+0.39}_{-4.75}$ & $55.59^{+0.18}_{-1.77}$ & $60.83^{+1.58}_{-2.18}$ & $45.35^{+0.79}_{-4.56}$ & $41.62^{+1.04}_{-1.91}$ & $57.24^{+2.67}_{-0.54}$ \\
VSD(16,0.25) + TSA             & $43.85^{+0.41}_{-5.51}$ & $\textbf{56.98}^{+0.32}_{-3.34}$ & $60.55^{+0.50}_{-0.74}$ & $43.66^{+1.01}_{-11.42}$ & $40.58^{+1.69}_{-1.70}$ & $57.15^{+0.87}_{-0.01}$ \\
VSD(16,0.75) + TSA             & $43.62^{+0.70}_{-7.72}$ & $54.32^{+1.09}_{-4.99}$ & $59.78^{+1.83}_{-0.72}$ & $43.31^{+1.17}_{-2.01}$ & $\textbf{42.39}^{+0.55}_{-0.28}$ & $56.54^{+2.30}_{-2.67}$ \\
\midrule
\textit{Voronoi(Gray) Only} \\
VSD(4,0.25)                    & $38.55^{+2.83}_{-1.47}$ & $53.95^{+0.86}_{-1.25}$ & $60.42^{+0.27}_{-0.15}$ & --- & --- & --- \\
VSD(4,0.75)                    & $38.13^{+2.84}_{-7.82}$ & $53.70^{+1.67}_{-0.16}$ & $60.39^{+0.10}_{-0.30}$ & --- & --- & --- \\
VSD(8,0.25)                    & $35.93^{+2.37}_{-0.07}$ & $54.82^{+0.34}_{-5.09}$ & $60.30^{+1.30}_{-0.27}$ & --- & --- & --- \\
VSD(8,0.75)                    & $38.70^{+0.76}_{-8.69}$ & $53.80^{+1.31}_{-4.20}$ & $60.75^{+0.59}_{-0.65}$ & --- & --- & --- \\
VSD(16,0.25)                    & $37.99^{+1.15}_{-3.50}$ & $54.55^{+0.14}_{-7.02}$ & $59.64^{+2.79}_{-0.05}$ & --- & --- & --- \\
VSD(16,0.75)                    & $39.51^{+2.11}_{-10.91}$ & $54.03^{+0.76}_{-3.64}$ & $60.66^{+1.06}_{-0.37}$ & --- & --- & --- \\
\midrule
\textit{Voronoi(RGB) + TSA} \\
VSD(4,0.25) + TSA             & $43.12^{+0.65}_{-10.60}$ & $56.43^{+0.88}_{-5.41}$ & $58.94^{+0.97}_{-0.43}$ & --- & --- & --- \\
VSD(4,0.75) + TSA             & $41.48^{+1.06}_{-7.50}$ & $55.41^{+0.58}_{-2.56}$ & $60.70^{+0.15}_{-0.10}$ & --- & --- & --- \\
VSD(8,0.25) + TSA             & $43.43^{+0.96}_{-5.98}$ & $55.83^{+0.21}_{-1.08}$ & $60.77^{+0.34}_{-0.78}$ & --- & --- & --- \\
VSD(8,0.75) + TSA             & $40.52^{+1.86}_{-2.13}$ & $56.51^{+0.03}_{-5.58}$ & $60.83^{+1.41}_{-0.55}$ & --- & --- & --- \\
VSD(16,0.25) + TSA             & $\textbf{44.23}^{+1.20}_{-6.67}$ & $56.20^{+1.32}_{-3.31}$ & $59.95^{+0.70}_{-0.21}$ & --- & --- & --- \\
VSD(16,0.75) + TSA             & $41.93^{+0.54}_{-6.47}$ & $54.66^{+0.17}_{-1.60}$ & $60.41^{+0.67}_{-0.23}$ & --- & --- & --- \\
\midrule
\textit{Voronoi(RGB) Only} \\
VSD(4,0.25)                    & $38.10^{+1.27}_{-2.98}$ & $55.00^{+0.94}_{-2.16}$ & $60.67^{+0.06}_{-1.83}$ & $38.45^{+1.64}_{-4.04}$ & $32.28^{+0.64}_{-3.97}$ & $61.21^{+0.71}_{-0.18}$ \\
VSD(4,0.75)                    & $40.48^{+0.98}_{-3.57}$ & $54.22^{+1.33}_{-1.32}$ & $60.56^{+0.23}_{-0.84}$ & $41.12^{+5.21}_{-3.44}$ & $31.95^{+3.73}_{-2.07}$ & $61.73^{+1.00}_{-2.57}$ \\
VSD(8,0.25)                    & $36.95^{+0.17}_{-0.41}$ & $55.80^{+0.05}_{-3.94}$ & $60.76^{+0.24}_{-0.62}$ & $34.44^{+3.72}_{-1.01}$ & $30.65^{+2.58}_{-0.20}$ & $60.92^{+0.43}_{-0.25}$ \\
VSD(8,0.75)                    & $38.11^{+1.51}_{-1.01}$ & $52.66^{+1.65}_{-0.84}$ & $60.99^{+0.03}_{-0.02}$ & $38.16^{+2.55}_{-3.57}$ & $35.66^{+1.32}_{-0.68}$ & $61.52^{+0.46}_{-0.80}$ \\
VSD(16,0.25)                    & $35.18^{+0.76}_{-0.77}$ & $54.01^{+1.41}_{-3.94}$ & $59.64^{+0.37}_{-0.46}$ & $31.56^{+0.88}_{-0.72}$ & $29.36^{+1.87}_{-0.08}$ & $\textbf{62.80}^{+0.24}_{-0.06}$ \\
VSD(16,0.75)                    & $38.61^{+2.03}_{-7.77}$ & $53.96^{+1.07}_{-0.18}$ & $60.43^{+0.19}_{-0.37}$ & $37.20^{+2.77}_{-0.07}$ & $32.35^{+3.81}_{-3.70}$ & $60.27^{+1.30}_{-0.42}$ \\
\bottomrule
\end{tabular}
\end{table}

\begin{table}[h]
\centering
\caption{Ablation on rgb stylization and TSA on the interior data. mIoU on real-world validation set (median over 3 runs, $\pm$ shows distance to min/max). Best per column in \textbf{bold}.}
\label{tab:ablation_all_main_results_3runs}
\small
\setlength{\tabcolsep}{5pt}
\begin{tabular}{l ccc}
\toprule
\textbf{Method} & DeepLabV3+ & SegFormer & M2Former \\
\midrule
\textit{Voronoi Only} \\
VSD(4,0.25)        & $39.73^{+1.65}_{-1.18}$ & $54.81^{+0.77}_{-0.86}$ & $60.69^{+0.09}_{-0.42}$ \\
VSD(4,0.75)        & $39.78^{+1.19}_{-1.65}$ & $53.84^{+1.53}_{-0.14}$ & $60.49^{+0.76}_{-0.40}$ \\
VSD(8,0.25)        & $38.30^{+3.43}_{-2.37}$ & $54.82^{+0.34}_{-0.49}$ & $60.30^{+0.52}_{-0.27}$ \\
VSD(8,0.75)        & $38.70^{+0.76}_{-0.10}$ & $55.09^{+0.02}_{-1.29}$ & $60.39^{+0.36}_{-0.29}$ \\
VSD(16,0.25)       & $39.14^{+1.47}_{-1.15}$ & $54.55^{+0.14}_{-0.24}$ & $59.64^{+0.04}_{-0.05}$ \\
VSD(16,0.75)       & $41.10^{+0.52}_{-1.59}$ & $54.03^{+0.76}_{-0.60}$ & $60.66^{+0.56}_{-0.37}$ \\
\midrule
\textit{Voronoi(RGB) + TSA} \\
VSD(4,0.25) + TSA  & $43.12^{+0.65}_{-0.23}$ & $\textbf{57.31}^{+0.39}_{-0.88}$ & $58.94^{+0.96}_{-0.43}$ \\
VSD(4,0.75) + TSA  & $41.70^{+0.84}_{-0.22}$ & $55.99^{+0.20}_{-0.58}$ & $60.85^{+0.68}_{-0.25}$ \\
VSD(8,0.25) + TSA  & $43.43^{+0.96}_{-0.42}$ & $56.04^{+1.11}_{-0.21}$ & $60.77^{+0.34}_{-0.17}$ \\
VSD(8,0.75) + TSA  & $42.38^{+1.66}_{-1.86}$ & $56.51^{+0.03}_{-0.53}$ & $\textbf{61.30}^{+0.94}_{-0.47}$ \\
VSD(16,0.25) + TSA & $\textbf{45.43}^{+2.31}_{-1.20}$ & $56.69^{+0.83}_{-0.49}$ & $60.39^{+0.26}_{-0.65}$ \\
VSD(16,0.75) + TSA & $41.93^{+0.54}_{-0.90}$ & $54.83^{+0.41}_{-0.17}$ & $60.33^{+0.08}_{-0.15}$ \\
\midrule
\textit{Voronoi(RGB) Only} \\
VSD(4,0.25)        & $38.10^{+1.27}_{-3.01}$ & $55.61^{+0.33}_{-0.61}$ & $60.67^{+0.06}_{-1.17}$ \\
VSD(4,0.75)        & $40.48^{+0.98}_{-4.08}$ & $55.04^{+0.51}_{-0.82}$ & $60.63^{+0.16}_{-0.07}$ \\
VSD(8,0.25)        & $37.12^{+3.28}_{-0.58}$ & $55.85^{+0.44}_{-0.05}$ & $60.71^{+0.29}_{-0.57}$ \\
VSD(8,0.75)        & $39.62^{+0.41}_{-1.51}$ & $54.31^{+0.75}_{-1.65}$ & $60.99^{+0.26}_{-0.02}$ \\
VSD(16,0.25)       & $35.94^{+1.33}_{-1.53}$ & $54.01^{+1.41}_{-1.13}$ & $59.43^{+0.21}_{-0.25}$ \\
VSD(16,0.75)       & $38.61^{+2.03}_{-2.70}$ & $53.96^{+1.06}_{-0.18}$ & $60.43^{+0.63}_{-0.37}$ \\
\bottomrule
\end{tabular}
\end{table}

\begin{table}[h]
\centering
\caption{Ablation on TSA on the exterior data. mIoU on real-world validation set (median over 3 runs, $\pm$ shows distance to min/max). Best per column in \textbf{bold}. Voronoi(RGB) only is applied to source RGB images.}
\label{tab:ablation_all_main_results_3runs_ext}
\small
\setlength{\tabcolsep}{5pt}
\begin{tabular}{l ccc}
\toprule
\textbf{Method} & DeepLabV3+ & SegFormer & M2Former \\
\midrule
\textit{Voronoi(Gray) + TSA } \\
VSD(4,0.25) + TSA        & $48.54^{+0.46}_{-1.04}$ & $40.44^{+1.67}_{-0.69}$ & $55.43^{+0.84}_{-0.18}$ \\ 
VSD(4,0.75) + TSA       & $49.00^{+0.13}_{-0.60}$ & $42.53^{+1.04}_{-0.55}$ & $57.57^{+2.04}_{-1.03}$ \\ 
VSD(8,0.25) + TSA       & $\textbf{50.44}^{+0.87}_{-0.69}$ & $42.02^{+1.38}_{-0.10}$ & $57.16^{+1.06}_{-0.97}$ \\ 
VSD(8,0.75) + TSA       & $48.64^{+0.20}_{-3.44}$ & $41.04^{+2.52}_{-0.57}$ & $58.81^{+0.25}_{-0.50}$ \\ 
VSD(16,0.25) + TSA     & $48.31^{+0.03}_{-1.76}$ & $42.44^{+2.99}_{-0.61}$ & $56.29^{+0.26}_{-1.84}$ \\ 
VSD(16,0.75) + TSA     & $46.36^{+0.51}_{-0.53}$ & $\textbf{43.69}^{+0.32}_{-1.66}$ & $55.35^{+0.30}_{-1.58}$ \\ 
\midrule
\textit{Voronoi(RGB) Only} \\
VSD(4,0.25)                 & $27.46^{+5.57}_{-0.07}$ & $31.08^{+3.18}_{-0.46}$ & $61.67^{+1.48}_{-1.06}$ \\ 
VSD(4,0.75)                 & $41.06^{+0.80}_{-2.42}$ & $36.82^{+0.98}_{-1.13}$ & $61.38^{+0.17}_{-2.09}$ \\ 
VSD(8,0.25)                 & $30.22^{+1.12}_{-4.94}$ & $36.22^{+0.21}_{-3.16}$ & $61.31^{+1.11}_{-0.34}$ \\ 
VSD(8,0.75)                 & $38.91^{+0.89}_{-0.00}$ & $36.44^{+2.29}_{-0.74}$ & $61.88^{+0.28}_{-0.42}$ \\ 
VSD(16,0.25)                & $30.60^{+1.81}_{-6.60}$ & $32.57^{+2.13}_{-1.29}$ & $62.01^{+0.67}_{-0.35}$ \\ 
VSD(16,0.75)                & $39.13^{+2.25}_{-3.21}$ & $37.59^{+0.37}_{-1.82}$ & $\textbf{62.07}^{+0.58}_{-0.55}$ \\ 
\bottomrule
\end{tabular}
\end{table}

\FloatBarrier

\begin{table}[h]
\centering
\caption{Cue-Decomposition Shape Bias $S_{\mathrm{cd}}$: EED vs Voronoi-shuffled 128 (mean over 3 runs) on interior data.}
\label{tab:cdsb_voronoi_128_full}
\small
\setlength{\tabcolsep}{5pt}
\begin{tabular}{l ccc ccc ccc }
\toprule
 & \multicolumn{3}{c}{\textbf{DeepLabV3+}} & \multicolumn{3}{c}{\textbf{SegFormer}} & \multicolumn{3}{c}{\textbf{M2Former}} \\
\cmidrule(lr){2-4}\cmidrule(lr){5-7}\cmidrule(lr){8-10}
\textbf{Method} & EED & Vor & $S_{\mathrm{cd}}$ & EED & Vor & $S_{\mathrm{cd}}$ & EED & Vor & $S_{\mathrm{cd}}$ \\
\midrule
Real Baseline & 34.9 & 8.4 & 0.620 & 40.1 & 16.4 & 0.489 & 44.2 & 25.0 & 0.410 \\
Synthetic Baseline & 6.8 & 6.3 & 0.296 & 14.1 & 8.5 & 0.394 & 21.1 & 12.5 & 0.399 \\ \midrule
TSA Only & 16.3 & 6.6 & 0.490 & 22.8 & 7.0 & 0.562 & 28.4 & 9.2 & 0.549 \\
VSD(1,1.0) & 9.5 & 3.6 & 0.513 & 30.7 & 6.9 & 0.637 & 38.4 & 10.3 & 0.595 \\ \midrule
VSD(4,0.25) + TSA & 20.1 & 7.1 & 0.527 & 27.0 & 9.3 & 0.532 & 33.8 & 12.7 & 0.511 \\
VSD(4,0.75) + TSA & 20.2 & 5.7 & 0.583 & 25.5 & 6.8 & 0.595 & 35.0 & 11.1 & 0.553 \\
VSD(8,0.25) + TSA & 17.8 & 6.8 & 0.508 & 26.8 & 9.3 & 0.531 & 31.8 & 11.7 & 0.517 \\
VSD(8,0.75) + TSA & 20.1 & 5.8 & 0.576 & 28.3 & 7.9 & 0.583 & 34.9 & 11.8 & 0.538 \\
VSD(16,0.25) + TSA & 17.1 & 6.8 & 0.497 & 25.2 & 9.0 & 0.524 & 30.3 & 12.3 & 0.492 \\
VSD(16,0.75) + TSA & 20.2 & 6.3 & 0.558 & 25.2 & 7.2 & 0.580 & 33.6 & 11.9 & 0.526 \\ \midrule
VSD(4,0.25) & 10.2 & 6.9 & 0.368 & 20.9 & 10.1 & 0.447 & 28.9 & 14.8 & 0.435 \\
VSD(4,0.75) & 18.4 & 5.3 & 0.579 & 24.3 & 9.1 & 0.512 & 32.6 & 14.1 & 0.477 \\
VSD(8,0.25) & 9.8 & 6.3 & 0.379 & 19.1 & 10.0 & 0.429 & 27.4 & 15.2 & 0.415 \\
VSD(8,0.75) & 18.6 & 6.3 & 0.537 & 24.3 & 9.8 & 0.494 & 32.8 & 13.5 & 0.488 \\
VSD(16,0.25) & 7.9 & 6.1 & 0.338 & 18.8 & 10.7 & 0.409 & 27.3 & 14.8 & 0.421 \\
VSD(16,0.75) & 18.1 & 5.4 & 0.570 & 23.6 & 8.9 & 0.511 & 31.8 & 13.6 & 0.479 \\
\bottomrule
\end{tabular}
\end{table}

\begin{table}[h]
\centering
\caption{Cue-Decomposition Shape Bias $S_{\mathrm{cd}}$: EED vs Voronoi-shuffled 128 (mean over 3 runs). Evaluated on exterior RANUS NIR. $VSD(n,p)$ without TSA refers to
Voronoi style diversification with RGB source and style images.}
\label{tab:scd_voronoi_128}
\small
\setlength{\tabcolsep}{5pt}
\begin{tabular}{l ccc ccc ccc }
\toprule
 & \multicolumn{3}{c}{\textbf{DeepLabV3+}} & \multicolumn{3}{c}{\textbf{SegFormer}} & \multicolumn{3}{c}{\textbf{M2Former}} \\
\cmidrule(lr){2-4}\cmidrule(lr){5-7}\cmidrule(lr){8-10}
\textbf{Method} & EED & Vor & $S_{\mathrm{cd}}$ & EED & Vor & $S_{\mathrm{cd}}$ & EED & Vor & $S_{\mathrm{cd}}$ \\
\midrule
Real Baseline &            21.1 & 7.6 & 0.371 & 30.1 & 8.4 & 0.431 & 46.2 & 8.4 & 0.537 \\
Synthetic Baseline              & 11.6 & 3.2 & 0.430 & 12.7 & 6.2 & 0.301 & 20.2 & 2.8 & 0.601 \\
\midrule
TSA Only              & 17.2 & 5.7 & 0.390 & 26.2 & 6.2 & 0.470 & 34.8 & 3.6 & 0.668 \\
VSD(1,1.0)             & 11.8 & 3.2 & 0.437 & 17.5 & 3.5 & 0.516 & 41.7 & 4.0 & 0.684 \\
VSD(4,0.25) + TSA               & 18.9 & 7.4 & 0.349 & 27.3 & 6.2 & 0.483 & 43.0 & 6.3 & 0.591 \\
VSD(4,0.75) + TSA               & 24.6 & 5.7 & 0.477 & 29.6 & 4.6 & 0.577 & 42.6 & 8.0 & 0.529 \\
VSD(8,0.25) + TSA               & 21.3 & 7.1 & 0.388 & 27.6 & 6.1 & 0.487 & 43.1 & 7.0 & 0.564 \\
VSD(8,0.75) + TSA               & 21.6 & 6.2 & 0.424 & 29.7 & 5.0 & 0.556 & 43.1 & 7.9 & 0.534 \\
VSD(16,0.25) + TSA               & 20.7 & 6.3 & 0.411 & 28.2 & 6.3 & 0.486 & 41.3 & 7.3 & 0.542 \\
VSD(16,0.75) + TSA               & 21.3 & 5.5 & 0.452 & 30.2 & 5.9 & 0.517 & 42.5 & 7.1 & 0.557 \\
\midrule
VSD(1,1.0)             & 20.4 & 3.2 & 0.598 & 26.3 & 4.4 & 0.582 & 43.3 & 2.1 & 0.830 \\
VSD(4,0.25)                 & 8.6 & 4.5 & 0.310 & 19.2 & 5.4 & 0.453 & 41.5 & 8.7 & 0.526 \\
VSD(4,0.75)                 & 12.7 & 3.1 & 0.488 & 25.8 & 4.7 & 0.562 & 39.7 & 8.6 & 0.519 \\
VSD(8,0.25)                 & 8.5 & 2.5 & 0.439 & 16.5 & 5.2 & 0.428 & 39.0 & 8.9 & 0.506 \\
VSD(8,0.75)                 & 12.0 & 3.6 & 0.436 & 22.8 & 5.1 & 0.513 & 38.9 & 9.2 & 0.496 \\
VSD(16,0.25)                & 8.1 & 5.2 & 0.269 & 14.0 & 6.6 & 0.331 & 35.5 & 9.5 & 0.467 \\
VSD(16,0.75)                & 12.5 & 2.8 & 0.507 & 23.9 & 5.5 & 0.503 & 37.0 & 8.9 & 0.494 \\
\bottomrule
\end{tabular}
\end{table}

\begin{table}[h]
\centering
\caption{Per-class IoU on the Interior validation set per model: Synthetic baseline vs per-architecture best (aggregation across 3 runs = median). Gain = Best $-$ Synthetic.}
\label{tab:per_class_by_model_interior}
\small
\setlength{\tabcolsep}{5pt}
\begin{tabular}{l c c@{\hspace{10pt}} c c c @{\hspace{10pt}}c c c @{\hspace{10pt}}c}
\toprule
& \multicolumn{3}{c}{\textbf{DeepLabV3+}} & \multicolumn{3}{c}{\textbf{SegFormer}} & \multicolumn{3}{c}{\textbf{M2Former}} \\
\cmidrule(lr){2-4}\cmidrule(lr){5-7}\cmidrule(lr){8-10}
\textbf{Class} & Syn & Best & Gain & Syn & Best & Gain & Syn & Best & Gain \\
\midrule
seat backrest & 13.79 & 65.90 & 52.11 & 71.93 & 86.92 & 14.99 & 68.57 & 85.89 & 17.32 \\
seat pad & 12.92 & 40.73 & 27.81 & 49.42 & 66.94 & 17.52 & 43.76 & 69.82 & 26.06 \\
child seat & 3.25 & 9.13 & 5.88 & 10.89 & 35.12 & 24.23 & 10.07 & 24.42 & 14.35 \\
seat belt & 18.57 & 43.86 & 25.29 & 39.72 & 48.87 & 9.15 & 55.93 & 54.25 & -1.68 \\
face mask & 20.19 & 38.64 & 18.45 & 43.23 & 56.14 & 12.91 & 70.38 & 66.56 & -3.82 \\
background & 58.81 & 78.51 & 19.70 & 82.93 & 87.91 & 4.98 & 85.74 & 88.41 & 2.67 \\
seat headrest & 44.83 & 60.10 & 15.27 & 75.23 & 74.85 & -0.38 & 73.66 & 76.82 & 3.16 \\
person & 38.53 & 42.41 & 3.88 & 54.47 & 60.91 & 6.44 & 61.66 & 64.06 & 2.40 \\
head-ware & 3.27 & 5.89 & 2.62 & 10.51 & 15.27 & 4.76 & 18.97 & 21.10 & 2.13 \\
clothing & 59.76 & 65.80 & 6.04 & 70.38 & 72.59 & 2.21 & 79.32 & 79.09 & -0.23 \\
face & 65.91 & 73.03 & 7.12 & 77.18 & 78.74 & 1.56 & 83.35 & 82.60 & -0.75 \\
glasses & 39.84 & 30.80 & -9.04 & 39.53 & 39.55 & 0.02 & 48.91 & 60.89 & 11.98 \\
hair & 54.56 & 55.91 & 1.35 & 67.10 & 64.59 & -2.51 & 72.56 & 73.72 & 1.16 \\
other object & 9.42 & 6.93 & -2.49 & 13.55 & 11.78 & -1.77 & 22.55 & 24.57 & 2.02 \\
\midrule
\textit{mIoU} & 32.00 & 44.26 & 12.26 & 49.18 & 57.30 & 8.12 & 56.79 & 62.04 & 5.25 \\
\bottomrule
\end{tabular}
\vspace{0.25em}
\\ \footnotesize Best configs: DeepLabV3+: VSD(16,0.25) + TSA; SegFormer: VSD(16,0.25) + TSA; M2Former: VSD(1,1.0). Aggregation across runs: median.
\end{table}

\begin{table}[h]
\centering
\caption{Per-class IoU on the Exterior validation set per model: Synthetic GTA5 RGB baseline vs per-architecture best (aggregation across 3 runs = median). Gain = Best $-$ Synthetic.}
\label{tab:per_class_by_model_exterior}
\small
\begin{tabular}{l c c@{\hspace{10pt}} c c c @{\hspace{10pt}}c c c @{\hspace{10pt}}c}
\toprule
& \multicolumn{3}{c}{\textbf{DeepLabV3+}} & \multicolumn{3}{c}{\textbf{SegFormer}} & \multicolumn{3}{c}{\textbf{M2Former}} \\
\cmidrule(lr){2-4}\cmidrule(lr){5-7}\cmidrule(lr){8-10}
\textbf{Class} & Syn & Best & Gain & Syn & Best & Gain & Syn & Best & Gain \\
\midrule
construction & 13.4 & 42.9 & 29.5 & 29.6 & 41.6 & 11.9 & 56.5 & 64.3 & 7.7 \\
ground & 4.9 & 21.1 & 16.2 & 2.8 & 11.0 & 8.2 & 21.4 & 31.2 & 9.8 \\
human & 0.2 & 1.7 & 1.5 & 0.2 & 0.1 & -0.2 & 20.4 & 14.4 & -6.1 \\
object & 17.0 & 23.4 & 6.4 & 17.0 & 16.2 & -0.8 & 30.9 & 31.7 & 0.9 \\
road & 27.0 & 84.7 & 57.7 & 52.1 & 79.2 & 27.1 & 89.8 & 91.0 & 1.2 \\
sky & 51.2 & 87.3 & 36.0 & 68.7 & 89.3 & 20.6 & 90.8 & 94.1 & 3.2 \\
vegetation & 12.5 & 72.8 & 60.2 & 7.5 & 54.2 & 46.7 & 76.4 & 82.8 & 6.4 \\
vehicle & 49.7 & 70.2 & 20.5 & 44.8 & 54.3 & 9.5 & 86.4 & 86.8 & 0.5 \\
\midrule
\textit{mIoU} & 22.0 & 50.5 & 28.5 & 27.8 & 43.2 & 15.4 & 59.1 & 62.0 & 2.9 \\
\bottomrule
\end{tabular}
\vspace{0.25em}
\\ \footnotesize Best configs: DeepLabV3+: VSD(8,0.25) + TSA; SegFormer: VSD(16,0.75) + TSA; M2Former: VSD(1,1.0). Aggregation across runs: median.
\end{table}

\begin{table}[h]
\centering
\caption{Area-normalized mIoU under corruptions on interior sensing data for DeepLabV3+, SegFormer, and Mask2Former. Values are normalized to the Real Baseline (100 = baseline). Bold marks the highest score per column within each model.}
\label{tab:robustness_norm_all_models}
\small
\setlength{\tabcolsep}{5pt}
\renewcommand{\arraystretch}{1.1}
\begin{tabular}{l c c c c c}
\toprule
\textbf{Model} & \textbf{Elastic} & \textbf{Low-pass} & \textbf{Swirl} & \textbf{Uniform} & \textbf{Mean} \\
\midrule
\multicolumn{6}{l}{\textbf{DeepLabV3+}} \\
Synthetic Baseline & 19.1 & 23.0 & 38.1 & 21.5 & 25.4 \\
TSA Only & 34.0 & 31.9 & 40.9 & 18.9 & 31.4 \\
VSD(1,1.0) & 41.7 & 18.1 & 31.7 & 24.7 & 29.0 \\
VSD(4,0.25)+ TSA [Gray] & 61.1 & 45.3 & 47.2 & 57.5 & 52.8 \\
VSD(4,0.75)+ TSA [Gray] & 60.9 & 51.0 & 44.0 & 57.0 & 53.2 \\
VSD(8,0.25)+ TSA [Gray] & 60.2 & 41.2 & 46.1 & 64.9 & 53.1 \\
VSD(8,0.75)+ TSA [Gray] & 60.6 & 55.7 & 42.8 & 70.4 & 57.4 \\
VSD(16,0.25)+ TSA [Gray] & 58.5 & 44.8 & 46.9 & 63.1 & 53.3 \\
VSD(16,0.75)+ TSA [Gray] & \textbf{60.8} & \textbf{57.3} & \textbf{46.5} & \textbf{70.8} & \textbf{58.8} \\
\midrule
\multicolumn{6}{l}{\textbf{SegFormer}} \\
Synthetic Baseline & 37.6 & 34.0 & 53.6 & 53.0 & 44.5 \\
TSA Only & 50.8 & 48.0 & 54.9 & 46.2 & 50.0 \\
VSD(1,1.0) & \textbf{74.5} & \textbf{66.7} & 58.7 & 76.8 & \textbf{69.2} \\
VSD(4,0.25)+ TSA [Gray] & 73.7 & 55.6 & \textbf{60.2} & 77.9 & 66.8 \\
VSD(4,0.75)+ TSA [Gray] & 72.3 & 54.9 & 55.2 & 73.8 & 64.1 \\
VSD(8,0.25)+ TSA [Gray] & 71.8 & 54.2 & 57.9 & 71.4 & 63.8 \\
VSD(8,0.75)+ TSA [Gray] & 72.7 & 52.7 & 56.8 & 75.1 & 64.3 \\
VSD(16,0.25)+ TSA [Gray] & 72.0 & 51.9 & 59.8 & 71.4 & 63.8 \\
VSD(16,0.75)+ TSA [Gray] & 69.0 & 52.4 & 54.8 & \textbf{77.9} & 61.0 \\
\midrule
\multicolumn{6}{l}{\textbf{Mask2Former}} \\
Synthetic Baseline & 47.9 & 53.4 & 60.0 & 68.3 & 57.4 \\
TSA Only & 61.9 & 53.2 & 59.3 & 52.3 & 56.7 \\
VSD(1,1.0) & \textbf{83.8} & \textbf{73.9} & \textbf{67.7} & 78.7 & \textbf{76.0} \\
VSD(4,0.25)+ TSA [Gray] & 79.1 & 63.7 & 63.8 & 77.6 & 71.1 \\
VSD(4,0.75)+ TSA [Gray] & 80.5 & 67.2 & 63.8 & 80.6 & 73.0 \\
VSD(8,0.25)+ TSA [Gray] & 77.9 & 61.8 & 62.7 & 75.1 & 69.4 \\
VSD(8,0.75)+ TSA [Gray] & 79.6 & 66.0 & 63.7 & \textbf{81.5} & 72.7 \\
VSD(16,0.25)+ TSA [Gray] & 75.0 & 59.7 & 63.3 & 73.0 & 67.7 \\
VSD(16,0.75)+ TSA [Gray] & 76.2 & 64.2 & 62.7 & 77.3 & 70.1 \\
\bottomrule
\end{tabular}
\end{table}

\begin{table}[h]
\centering
\caption{Area-normalized mIoU under corruptions for DeepLabV3+, SegFormer, and Mask2Former on RANUS NIR test images. Values are normalized to the Real Baseline (100 = baseline). Bold marks the highest score per column within each model.}
\label{tab:per_distortion_ranus_nir}
\small
\setlength{\tabcolsep}{5pt}
\renewcommand{\arraystretch}{1.1}
\begin{tabular}{l c c c c c}
\toprule
\textbf{Model} & \textbf{Elastic} & \textbf{Low-pass} & \textbf{Swirl} & \textbf{Uniform} & \textbf{Mean} \\
\midrule
\multicolumn{6}{l}{\textbf{DeepLabV3+}} \\
Synthetic Baseline & 23.5 & 49.8 & 27.6 & 22.9 & 31.0 \\
TSA Only & 38.2 & 72.0 & 60.9 & 39.1 & 52.5 \\
VSD(1,1.0) & 45.5 & 54.5 & 39.4 & 35.4 & 43.7 \\
VSD(4,0.25) + TSA & 51.0 & 67.1 & 67.6 & 70.8 & 64.1 \\
VSD(4,0.75) + TSA & 49.7 & 59.1 & 69.4 & \textbf{87.1} & 66.3 \\
VSD(8,0.25) + TSA & \textbf{51.1} & \textbf{78.3} & \textbf{70.6} & 72.4 & \textbf{68.1} \\
VSD(8,0.75) + TSA & 39.3 & 70.2 & 62.2 & 66.8 & 59.6 \\
VSD(16,0.25) + TSA & 44.7 & 53.2 & 68.4 & 82.0 & 62.1 \\
VSD(16,0.75) + TSA & 42.6 & 45.3 & 62.0 & 58.7 & 52.1 \\
\midrule
\multicolumn{6}{l}{\textbf{SegFormer}} \\
Synthetic Baseline & 41.7 & 27.7 & 39.9 & 30.8 & 35.0 \\
TSA Only & 53.8 & 39.2 & 57.2 & 55.7 & 51.5 \\
VSD(1,1.0) & 44.7 & 55.2 & 45.4 & 102.2 & 61.9 \\
VSD(4,0.25) + TSA & 57.8 & 50.1 & 52.8 & 103.5 & 66.1 \\
VSD(4,0.75) + TSA & \textbf{60.4} & 61.8 & 56.2 & \textbf{113.3} & \textbf{72.9} \\
VSD(8,0.25) + TSA & 57.5 & 54.1 & 57.0 & 97.2 & 66.4 \\
VSD(8,0.75) + TSA & 59.9 & \textbf{66.7} & 56.5 & 102.9 & 71.5 \\
VSD(16,0.25) + TSA & 58.3 & 55.3 & \textbf{59.5} & 89.9 & 65.8 \\
VSD(16,0.75) + TSA & 59.8 & 58.1 & 57.4 & 97.1 & 68.1 \\
\midrule
\multicolumn{6}{l}{\textbf{Mask2Former}} \\
Synthetic Baseline & 69.4 & 58.9 & 68.2 & 75.7 & 68.0 \\
TSA Only & 59.0 & 57.7 & 68.3 & 72.0 & 64.2 \\
VSD(1,1.0) & \textbf{93.8} & \textbf{78.2} & \textbf{77.6} & \textbf{86.6} & \textbf{84.0} \\
VSD(4,0.25) + TSA & 77.5 & 72.3 & 71.8 & 79.3 & 75.2 \\
VSD(4,0.75) + TSA & 82.0 & 77.1 & 74.2 & 84.2 & 79.4 \\
VSD(8,0.25) + TSA & 80.9 & 74.9 & 73.3 & 80.6 & 77.4 \\
VSD(8,0.75) + TSA & 84.3 & 75.3 & 75.7 & 86.0 & 80.3 \\
VSD(16,0.25) + TSA & 76.4 & 72.9 & 70.9 & 83.5 & 75.9 \\
VSD(16,0.75) + TSA & 77.8 & 68.2 & 71.8 & 81.9 & 74.9 \\
\bottomrule
\end{tabular}
\end{table}


\end{document}